\documentclass[pdflatex,sn-nature]{sn-jnl}
\usepackage{graphicx}%
\usepackage{multirow}%
\usepackage{amsmath,amssymb,amsfonts}%
\usepackage{amsthm}%
\usepackage{mathrsfs}%
\usepackage[title]{appendix}%
\usepackage{xcolor}%
\usepackage{textcomp}%
\usepackage{manyfoot}%
\usepackage{booktabs}%
\usepackage{algorithm}%
\usepackage{algorithmicx}%
\usepackage{algpseudocode}%
\usepackage{listings}%

\theoremstyle{thmstyleone}%

\theoremstyle{thmstyletwo}%

\theoremstyle{thmstylethree}%

\begin{document}

\title[Article Title]{Toward the Cognitive--Physical Limits of Embodied Intelligence through a World-Model-Centric Autonomous Racing Agent}

\author[1]{\fnm{Zitong} \sur{Shan}}
\equalcont{These authors contributed equally to this work.}

\author[1]{\fnm{Baichuan} \sur{Lou}}
\equalcont{These authors contributed equally to this work.}

\author[1]{\fnm{Yanxin} \sur{Zhou}}
\equalcont{These authors contributed equally to this work.}

\author[1]{\fnm{Shuge} \sur{Wu}}
\author[1]{\fnm{Xianqi} \sur{He}}
\author[1]{\fnm{Bolin} \sur{Zhao}}
\author[1]{\fnm{Sheng} \sur{Zhao}}
\author[1]{\fnm{Zhouheng} \sur{Li}}
\author[2]{\fnm{Chee Kiong} \sur{Ong}}
\author[1]{\fnm{King Ho Holden} \sur{Li}}
\author*[1]{\fnm{Chen} \sur{Lv}}\email{lyuchen@ntu.edu.sg}

\affil[1]{\orgdiv{School of Mechanical and Aerospace Engineering}, {\orgname{Nanyang Technological University}, \postcode{639798}, \country{Singapore}}}

\affil[2]{{\orgname{K2 Holding, L.L.C}, \city{Abu Dhabi}, \country{United Arab Emirates}}}

\abstract{Embodied artificial intelligence aims to develop agents that perceive, reason, and act through continuous interaction with the physical world. Despite rapid progress, most embodied systems are still evaluated within conservative safety margins or moderately complex interaction regimes, leaving their capability boundaries under extreme conditions insufficiently understood. For safe real-world deployment, it is essential to understand where the cognitive and physical limits of embodied agents lie, and how these limits interact. Autonomous racing provides a stringent testbed for this question by combining high-frequency localization and perception, adversarial interaction prediction and decision-making, near-saturated vehicle dynamics, and strict safety constraints. Existing autonomous racing systems have advanced high-speed performance, but rarely provide an explicit mechanism for jointly modeling and refining these coupled limits. Here we show that a world-model-centric autonomous racing agent provides a concrete step toward exploring the cognitive-physical limits of embodied intelligence. The framework learns predictive world models from near-limit successes and failures, capturing interaction evolution, ego dynamics, and feasible-motion boundaries. This couples world-state construction, future-aware reasoning, and near-limit control within a closed-loop refinement process. High quality training data were collected from real-vehicle autonomous racing, where the onboard system maintains robust localization and perception at speeds approaching 256.3 km/h with a peak lateral acceleration of 26.8 m/s\textsuperscript{2}. In full-scale autonomous racing setting, the well trained world-model-centric agent achieves an 88.3\% interaction success rate across various challenging simulated racing scenarios. Through closed-loop continual refinement of the world model and policy, the agent improves its utilization of cognitive-physical limits while recovering from failure modes and generalizing across varying conditions and unseen circuits. These results suggest a boundary-aware methodology in which the world model helps embodied agents represent, predict, and continually refine their capability boundaries for safer deployment.}


\maketitle

\section*{Introduction}

Embodied artificial intelligence (Embodied AI) has emerged as a central paradigm in robotics and artificial intelligence by emphasizing intelligence that arises through continuous interaction with the physical world \cite{pfeifer2006body, duan2022survey}. Unlike conventional AI systems that operate primarily in abstract or virtual domains, embodied agents must perceive, decide, and act under noisy sensing, nonlinear dynamics, and irreversible physical consequences \cite{spielberg2019neural}. Although substantial progress has been made across autonomous driving, aerial robotics, legged locomotion, and manipulation \cite{liu2025aligning}, most embodied systems are still developed and evaluated within conservative safety margins, moderately complex interactions, or well-modeled physical conditions \cite{yifan2025embodied}. As a result, their validity and capability boundaries under extreme conditions remain insufficiently understood \cite{amodei2016concrete}.

For safe real-world deployment, it is not sufficient to study only how embodied intelligence is generated, learned, or improved. It is equally important to understand where its limits lie, how these limits emerge from cognition and physics, and how they jointly constrain agent capability under demanding conditions. This challenge becomes most evident in extreme tasks, where agents operate persistently near the boundaries of feasible state and action spaces \cite{betz2022autonomous, kaufmann2023champion}. In such settings, small errors in perception, prediction, reasoning, or control can rapidly escalate into instability, failure, or collision \cite{ames2019control, hanover2024autonomous}.

In this work, the cognitive limit refers to the boundary of an agent's ability to construct, update, predict, and use task-relevant world understanding under severe time pressure, uncertainty, and dynamic interaction. For autonomous racing, this includes high-frequency localization and perception under high-speed motion, prediction of ego--opponent interaction evolution, and decision-making for safe, feasible, and competitive maneuvers. The physical limit refers to the boundary of feasible motion imposed by vehicle dynamics, actuator saturation, tire--road interaction, aerodynamic loading, friction, tire temperature, and stability constraints. These limits are coupled rather than independent: uncertainty in localization, perception, and interaction prediction constrains how safely the physical envelope can be exploited, while the evolving physical state determines which future actions remain predictable, feasible, and safe.

Autonomous racing provides a uniquely suitable testbed for investigating these coupled limits. While agile drones and legged robots have demonstrated remarkable near-limit physical performance \cite{sun2022aggressive, geles2024agileflight, hanover2024autonomous, hwangbo2019learning, wu2023learning, zhang2024learning}, and urban driving or long-horizon robotic manipulation emphasize complex reasoning under uncertainty \cite{huang2024uncertainty, shan2025deep}, relatively few embodied systems require sustained cognitive complexity and persistent operation near physical feasibility boundaries. Racing naturally combines high-speed motion, near-saturated tire--road dynamics, strict collision constraints, and adversarial multi-agent interaction among competing vehicles. It therefore requires an agent to operate near the edge of physical feasibility while simultaneously performing high-frequency localization and perception, interaction prediction, and rapid decisions such as overtaking, defending, and interaction-aware planning under uncertainty \cite{funke2012up}. Autonomous racing is thus not merely a high-performance control problem, but a platform for examining how cognitive reasoning and physical execution co-evolve under extreme conditions \cite{liniger2019noncooperative, wang2021game}.

Autonomous racing is distinguished by a mature and cross-scale research community \cite{betz2022autonomous,o2020f1tenth}. Scaled platforms offer low-cost, reproducible, and extensible testbeds for autonomous racing research \cite{o2020f1tenth, baumann2025enhancing}. These platforms are supported by open-source vehicle models, planning and control methods, and experimental tool chains \cite{baumann2025forzaeth,becker2022model,li6127037evo,hu2025fsdp}.  Recent work has further demonstrated the growing capability of autonomous racing systems, including trajectory-conditioned zero-shot racing and on-board reasoning with deployed large language models \cite{ghignone2024tc,baumann2025enhancing}. At the full-vehicle scale, studies on real-track racing have accumulated valuable experience in high-speed trajectory generation, near-limit maneuvering, and safety-constrained control \cite{heilmeier2020minimum,piccinini2025optimal,zarrouki2024safe}, providing a concrete basis for studying perception, planning, and control near physical feasibility boundaries \cite{betz2022autonomous}. The research ecosystem has given rise to a broad range of autonomous racing methods.

Existing autonomous racing approaches can be broadly grouped into optimization-based pipelines, learning-based policies, and hybrid frameworks \cite{kabzan2020amz, wischnewski2022indy, betz2023tum}. Optimization-based methods explicitly encode vehicle dynamics and feasibility constraints, allowing structured and interpretable operation near friction limits \cite{liniger2015optimization, scheffe2022sequential, alcala2020autonomous, wischnewski2022tube}; however, they remain difficult to extend to rapidly evolving multi-agent interactions involving opponent intent, behavioral uncertainty, and long-horizon coupling \cite{jung2021game,hu2025fsdp}. Learning-based methods, inspired by advances in game playing, simulated racing, and drone racing \cite{silver2016mastering, silver2018general, schrittwieser2020mastering, wurman2022outracing, kaufmann2023champion, song2023reaching,zhao2026vision}, can capture complex observation--action mappings for competitive driving \cite{wurman2022challenges, ji2018adaptive, fuchs2021super}, but often depend strongly on the training vehicle, track geometry, and interaction distribution \cite{pan2020imitation, sun2023benchmark}, with limited interpretability and reliability outside the original operating regime \cite{hamilton2022zero, prudencio2023survey, zare2024survey}. Hybrid methods improve robustness by combining model-based and data-driven components \cite{evans2023high}, for example by learning cost terms, vehicle dynamics, or residual corrections \cite{ivanov2020case, hermansdorfer2021e2evd, chen2023hybrid}.
Yet most efforts still optimize task performance within a given sensing--planning--control pipeline, rather than explicitly modeling how predictive reasoning, interaction understanding, and physical feasibility jointly influence the agent's behavior. As a result, autonomous racing lacks a systematic framework for studying embodied intelligence under coupled cognitive--physical limits.

This gap motivates the need for a unified predictive substrate that can connect cognitive reasoning with physical execution. From a system-level perspective, traditional autonomous racing methods often rely on instantaneous state estimation, reactive feature representations, or task-specific predictive modules rather than an integrated model of the world \cite{pfeifer2006body, betz2022autonomous, feng2025embodied}. For near-limit autonomous racing, such a model must predict ego dynamics near feasibility boundaries, capture structured interaction with opponents, represent uncertainty under tight safety margins, and refine decision-making and control capability over time. Recent advances in world models provide a promising foundation for this role \cite{ha2018recurrent, hafner2023mastering, li2025comprehensive}. By enabling structured simulation of future dynamics and interactions \cite{li2025comprehensive, feng2025embodied}, imagination-based reasoning, and counterfactual evaluation, world models offer a natural mechanism for linking predictive world understanding with decision-making and control \cite{hafner2019learning, schrittwieser2020mastering, ding2025understanding}.

\begin{figure*}[!htb]
\centering
\includegraphics[width=1.0\linewidth]{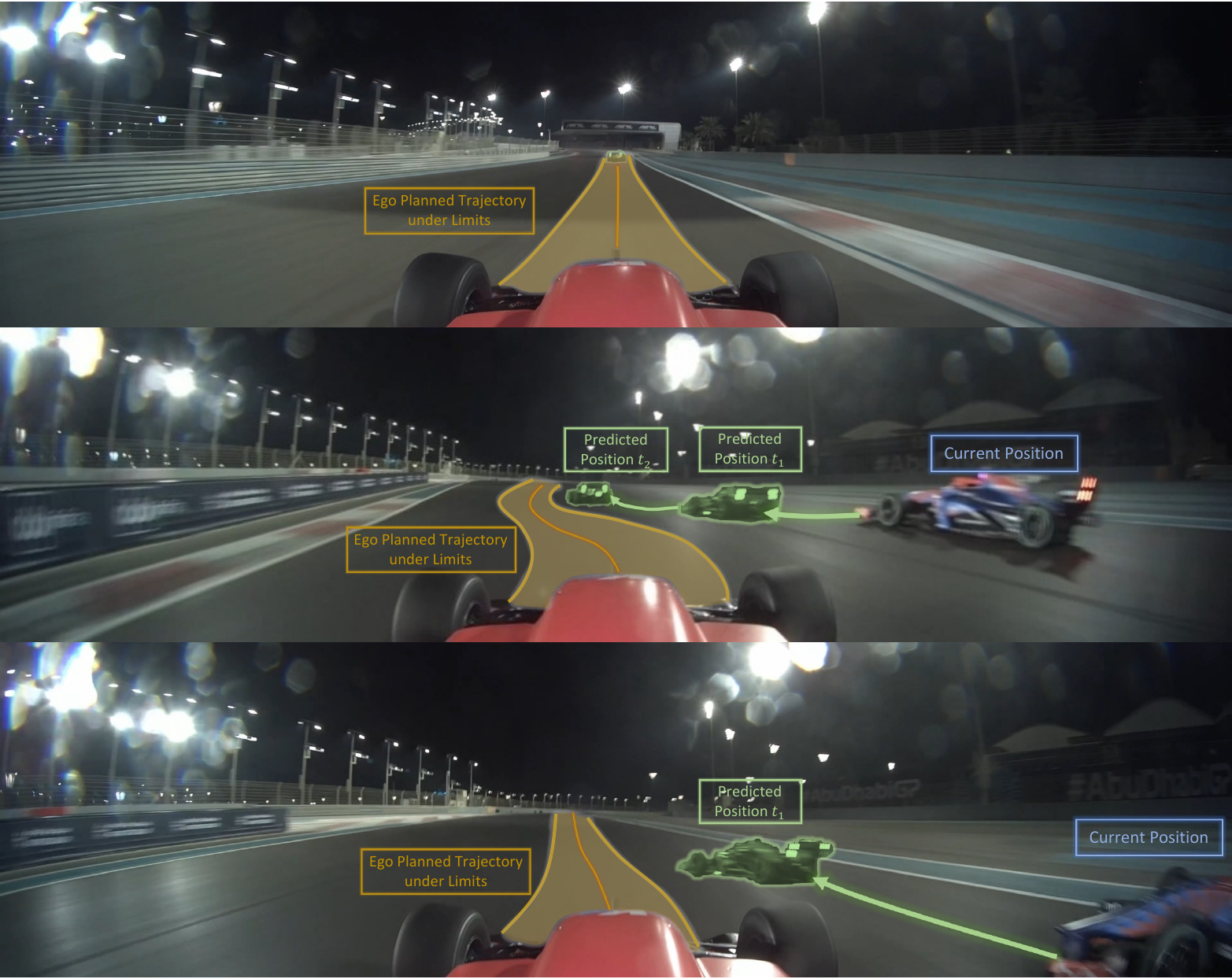}
\caption{\textbf{Autonomous racing under cognitive--physical limits.} Autonomous racing requires the agent to construct predictive world understanding, reason about the current and future positions of surrounding opponents, generate an ego trajectory that remains feasible under physical limits, and maneuver near the dynamics boundary.} 
\label{F0}
\end{figure*}

To this end, we take a step toward the cognitive-physical limits of embodied intelligence through a world-model-centric autonomous racing agent. The proposed framework places a predictive world model at the center of the agent, where it jointly models external interaction evolution, ego dynamics, and feasible motion boundaries. 
This supports high-frequency world-state construction, future-aware interaction reasoning, risk-sensitive decision-making, and limit-aware control across track geometries and operating conditions. 
Through continual refinement of both the world model and the policy, the agent improves decision-making and control capability for autonomous racing under joint cognitive-physical limits, as shown in Fig.~\ref{F0}. 
In this way, autonomous racing becomes not only an application domain, but also a concrete proving ground for studying how embodied agents can approach, exploit, and regulate coupled cognitive-physical limits.

\section*{Results}
\subsection*{World-Model-Centric Racing Agent Architecture}

Extreme autonomous racing makes the cognitive and physical limits of an embodied agent jointly observable. Cognitively, the agent must maintain robust localization and perception under high-speed and high-frequency operation, predict ego--opponent interaction evolution, and make safe and competitive decisions under severe time pressure. Physically, it must exploit the feasible dynamics envelope of the vehicle while respecting actuator constraints, tire--road interaction, and stability margins. The following results therefore evaluate not only racing performance, but also how the agent approaches, exploits, and regulates coupled cognitive-physical limits.

\begin{figure*}[!htb]
\centering
\includegraphics[width=1.0\linewidth]{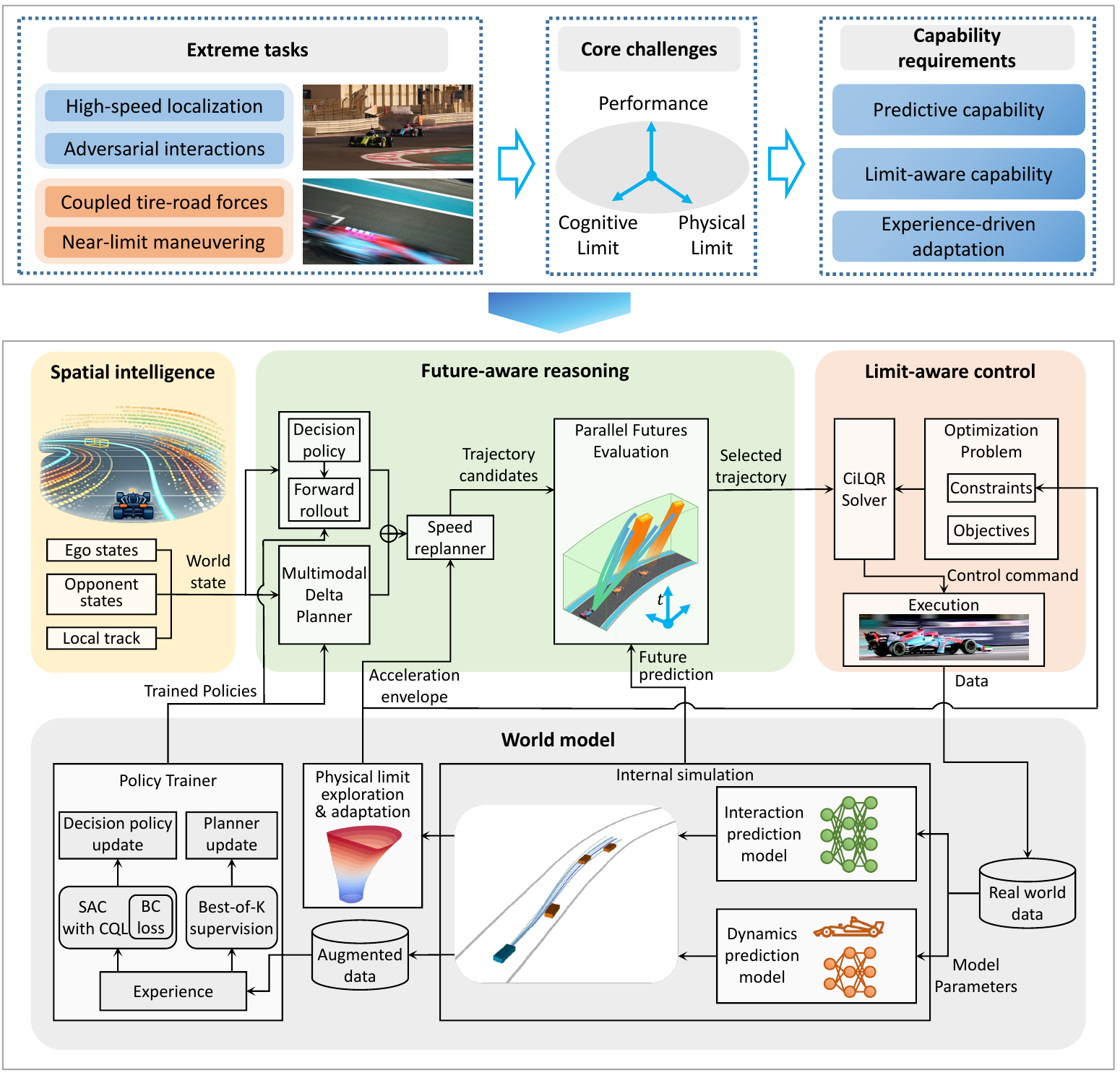}
\caption{\textbf{World-model-centric architecture for autonomous racing under coupled cognitive-physical limits.} Extreme racing tasks impose coupled cognitive and physical limits that require predictive capability, limit-aware control, and continual refinement beyond conventional paradigms. The proposed architecture is centered on a shared world model that connects spatial intelligence, future-aware reasoning, and limit-aware control. Real-vehicle racing data were collected for model training, providing physical grounding and interaction priors for world-model refinement. The the refined world model provides future predictions and feasible-motion priors for subsequent policy improvement.}
\label{F1}
\end{figure*}

Rather than treating localization, prediction, planning, and control as isolated modules, the proposed agent is organized around a shared world model that connects cognitive reasoning with physical control. As illustrated in Fig.~\ref{F1}, the architecture consists of three coupled components: spatial intelligence, future-aware reasoning, and limit-aware control. Spatial intelligence provides the observable basis of the agent's world-state representation through multi-modal sensor fusion for high-frequency localization and opponent perception. Future-aware reasoning uses world-model rollouts to evaluate possible interaction and dynamics futures. Limit-aware control translates the selected plan into executable commands while respecting the feasible motion boundary. Together, these components allow the agent to reason and act under simultaneous cognitive and physical constraints.

The world model serves as the central predictive substrate of the agent. It jointly models external interaction evolution, ego dynamics, and physical limits that constrain feasible motion. Candidate decisions are evaluated against predicted interaction outcomes and future physical feasibility, while physical-limit information feeds back into planning and control. At runtime, this forms a perception--prediction--decision--control flow: spatial intelligence constructs the current racing state; the world model predicts multi-agent interaction and ego-dynamics futures; future-aware reasoning selects a safe, feasible, and competitive trajectory; and limit-aware control executes it near the physical boundary.

Beyond online execution, the architecture supports continual refinement of both the world model and the policy. Real racing data update the model of ego dynamics, interaction evolution, and physical feasibility; the refined world model then provides structured predictions for improving future-aware reasoning and limit-aware control. This closed-loop refinement allows the agent to improve both cognitive capability, including interaction reasoning and decision-making, and physical capability, including dynamic-envelope utilization and stability near the limit.

Overall, the world-model-centric architecture provides a concrete mechanism for studying coupled cognitive-physical limits in autonomous racing. The subsequent results examine this mechanism through high-frequency localization and perception, predictive world modeling, adversarial interaction reasoning, limit-aware control, and progressive agent evolution across conditions.

\subsection*{Predictive World Model under Cognitive-Physical Limits}

Operating near the cognitive-physical limits requires the agent to both reconstruct the current racing state and predict its future evolution under near-limit conditions. This section evaluates these two connected layers: high-frequency localization and perception for current-state construction, and world modeling for predicting ego dynamics, physical feasibility, and interaction evolution. Together, they transform raw sensing into predictive world understanding for downstream reasoning and control.

\begin{figure*}[!htb]
\centering
\includegraphics[width=1.0\linewidth]{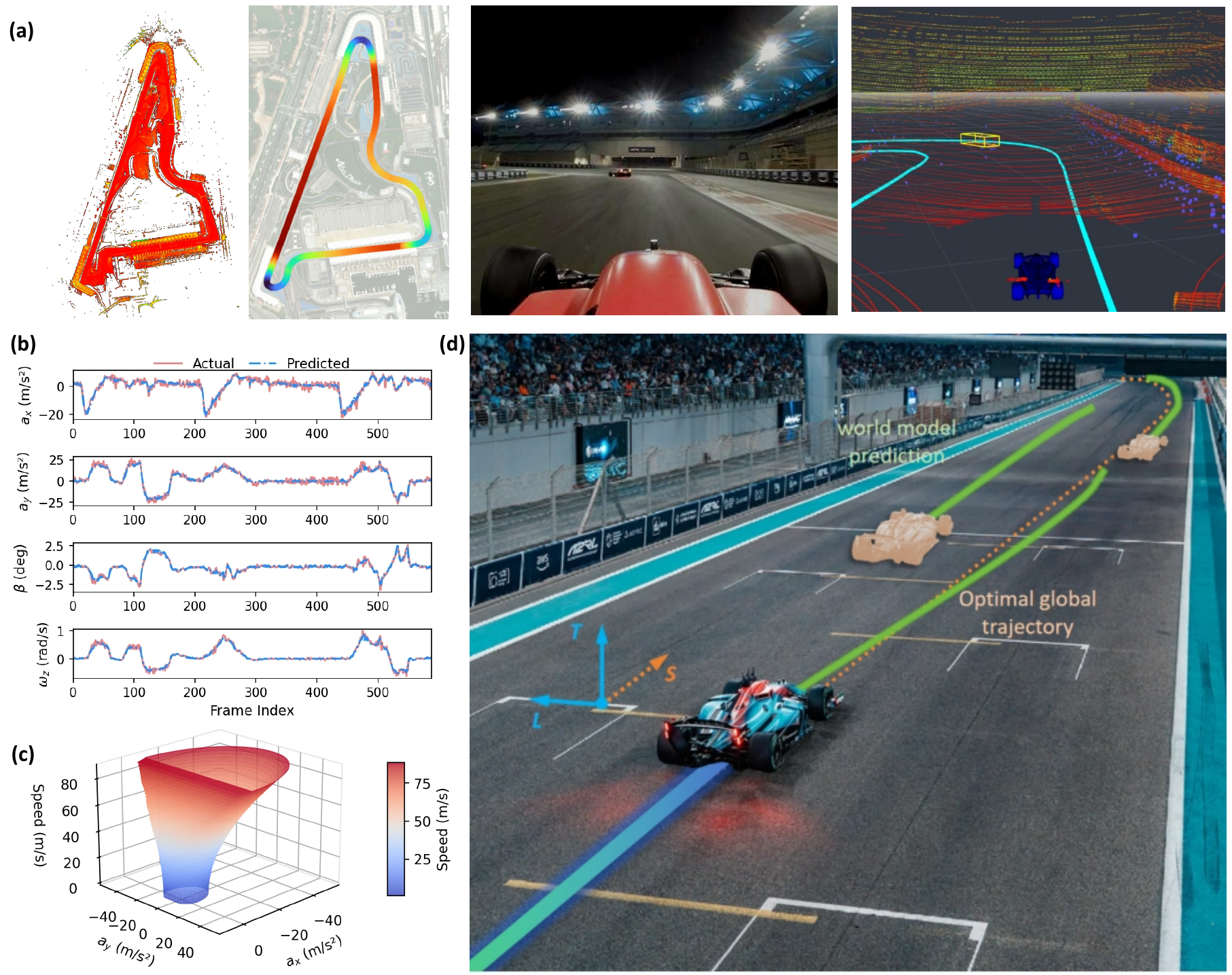}
\caption{\textbf{Predictive world understanding from high-frequency localization and world modeling.} High-frequency localization and opponent perception provide the current racing state, while the world model predicts ego dynamics, physical feasibility, and interaction evolution under near-limit racing conditions. (a) Localization and opponent perception in full-scale autonomous racing. (b) Ego dynamics prediction under aggressive maneuvering. (c) Physical motion boundary representation through feasible acceleration envelope evolution. (d) Interaction prediction under extreme conditions.}  
\label{F2}
\end{figure*}

High-frequency localization and perception provide the observable foundation of the current racing state. The localization and ego-state estimation module fuses LiDAR, RTK, IMU, and wheel-speed measurements to maintain a stable ego pose under aggressive maneuvering. As shown in Fig.~\ref{F2}(a), the point-cloud map and estimated motion trajectory remain consistently aligned during full-scale racing. At peak speeds approaching $256.3~\mathrm{km/h}$, the fused-state residuals remain bounded throughout the lap: the lateral position residual is generally within $0.3~\mathrm{m}$, the longitudinal position residual remains within about $2.0~\mathrm{m}$, and the longitudinal and lateral velocity residuals remain within about $1.8~\mathrm{m/s}$, with medians concentrated near $0.1~\mathrm{m/s}$. The yaw residual is bounded within about $0.06~\mathrm{rad}$ relative to dual-GPS heading and within about $0.01~\mathrm{rad}$ relative to LiDAR odometry. Opponent perception and tracking further extend the current racing state from ego localization to the interactive scene. Surrounding vehicles are detected and tracked under high-speed and close-proximity conditions by fusing LiDAR-based geometric measurements with radar-based velocity cues into unified object states through probabilistic state estimation. As shown in Fig.~\ref{F2}(a), the perceived opponent vehicles remain temporally coherent over full-lap runs, providing the opponent-state foundation required for interaction prediction and future-aware reasoning.

The world model then extends the current racing state into predictive understanding. It contains three complementary components: ego dynamics prediction, physical-limit representation, and interaction prediction. The ego dynamics model predicts future vehicle motion under varying control inputs using a differentiable dynamics backbone with a residual neural network model. As shown in Fig.~\ref{F2}(b), the model captures key near-limit motion states with RMSE values of $0.197~\mathrm{m/s}$ for longitudinal velocity, $0.099~\mathrm{m/s}$ for lateral velocity, $0.042~\mathrm{rad/s}$ for yaw rate, $1.514~\mathrm{m/s^2}$ for longitudinal acceleration, and $1.983~\mathrm{m/s^2}$ for lateral acceleration. These prediction errors quantify the ability of the world model to represent the ego vehicle's dynamics response under aggressive control inputs, providing the basis for evaluating future candidate maneuvers near the physical boundary.

The physical-limit representation quantifies the feasible motion boundary of the ego vehicle. As shown in Fig.~\ref{F2}(c), the feasible acceleration envelope varies with vehicle speed: achievable lateral and braking accelerations expand with increasing aerodynamic effects, while longitudinal acceleration capability decreases due to powertrain limitations. By integrating actuator capability, tire--road constraints, and speed-dependent vehicle dynamics into a coupled acceleration envelope, the world model provides a quantitative representation of physical feasibility. This envelope acts as a dynamic constraint for planning and control, allowing candidate actions to be evaluated not only by racing utility, but also by whether they remain within the current physical capability of the vehicle.

The interaction prediction component models the future evolution of multi-agent racing scenarios. An attention-based predictive interaction model encodes the relative spatial relations among interacting agents. By predicting opponent intentions and recursively rolling out the resulting state transitions, the interaction model anticipates interaction evolution beyond immediate observations. As illustrated in Fig.~\ref{F2}(d), the predictions are generated in a structured track-coordinate system, preserving the spatio-temporal consistency of the interactive scene. Over a $3~\mathrm{s}$ horizon, the model achieves a mean ADE of $1.58~\mathrm{m}$ and a median ADE of $0.68~\mathrm{m}$, together with a mean FDE of $2.55~\mathrm{m}$ and a median FDE of $0.78~\mathrm{m}$. This predictive capability allows downstream reasoning to evaluate candidate maneuvers against anticipated opponent motion rather than only the current observed state.

This two-layer characterization links the cognitive foundation of the agent to downstream decision-making and control. High-frequency localization and perception establish a reliable current racing state, while the world model predicts future ego dynamics, physical feasibility, and interaction evolution. The resulting predictive world understanding enables candidate maneuvers to be evaluated against both interaction uncertainty and physical feasibility, supporting future-aware decision-making and limit-aware control.

\subsection*{Future-Aware Reasoning for Adversarial Interaction}

Adversarial interaction in autonomous racing exposes the cognitive-physical coupling of the agent. A competitive maneuver is not safe or effective simply because it appears strategically advantageous at the current instant; it must also remain feasible under the future evolution of opponent motion and near-limit ego dynamics. Therefore, future-aware reasoning is formulated as a predictive decision process that evaluates candidate actions against both interaction uncertainty and physical feasibility.

\begin{figure*}[!htb]
\centering
\includegraphics[width=1.0\linewidth]{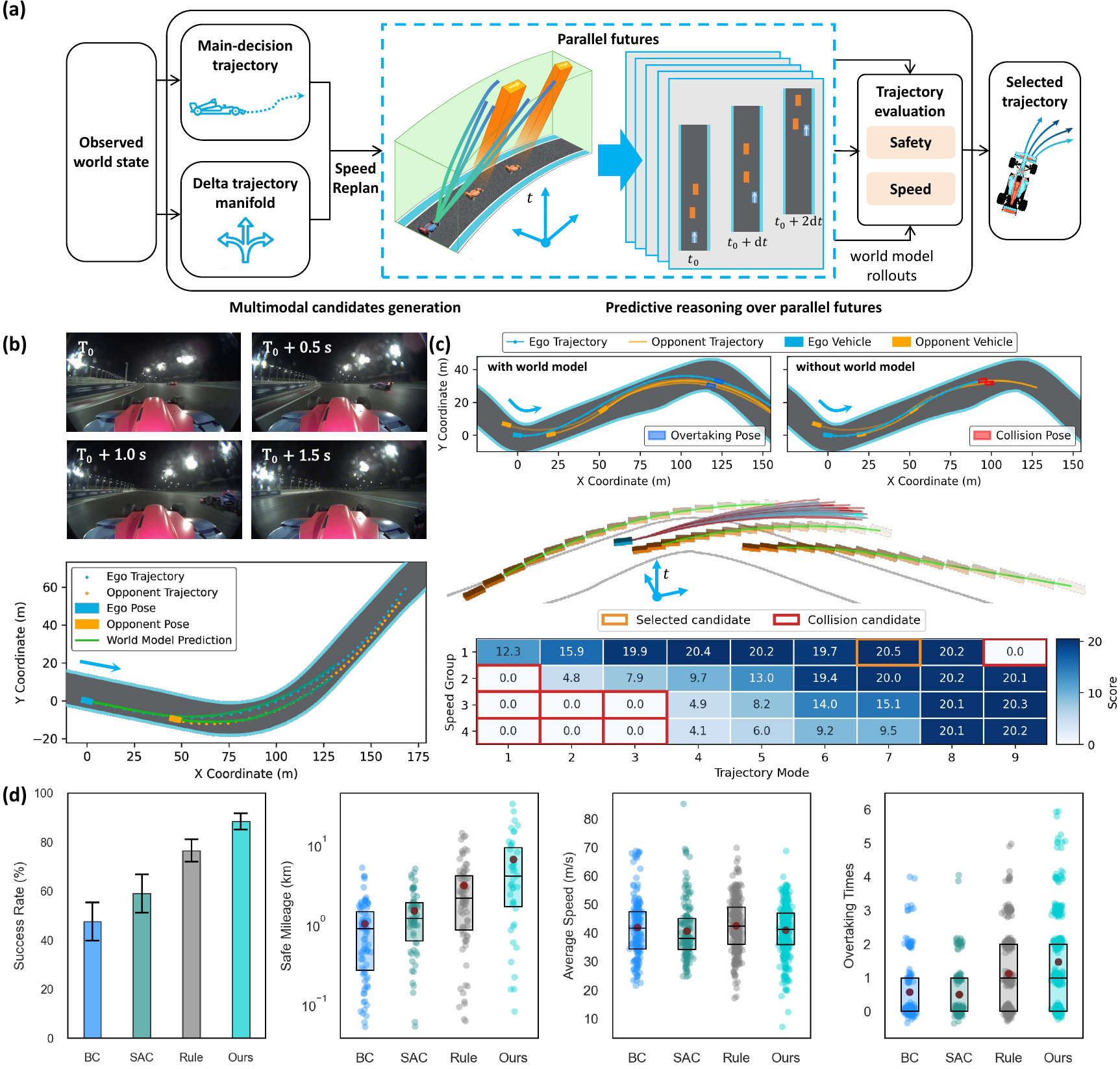}
\caption{\textbf{Future-aware reasoning for adversarial interaction under coupled cognitive-physical limits.} The world model supports multimodal trajectory generation, parallel future evaluation, and planning selection by jointly considering interaction safety, physical feasibility, and speed efficiency. (a) Future-aware reasoning process from observed world state to multimodal candidate generation and parallel future evaluation. (b) Real-world prediction validation during a wheel-to-wheel racing interaction. (c) Ablation of agent behavior with and without the world model, showing how prediction affects downstream physical execution. (d) Benchmark comparison of competitive interaction performance.}  
\label{F3}
\end{figure*}

As illustrated in Fig.~\ref{F3}(a), the proposed agent performs future-aware reasoning through parallel world-model rollouts. Based on the observed world state, the system generates a main-decision trajectory that represents the nominal racing intention. A delta-trajectory manifold is produced to encode multimodal deviations around this main trajectory. These delta trajectories are applied to the main-decision trajectory to form multimodal trajectories. To further diversify longitudinal behavior, the multimodal trajectories are combined with prescribed speed factors, yielding $36$ candidate trajectories for world-model evaluation. These candidate trajectories are propagated through the world model to produce parallel futures. Each future jointly contains predicted opponent responses, ego-motion evolution, and feasibility information under near-limit conditions. The resulting futures are then evaluated according to safety, physical feasibility, and speed efficiency, enabling the agent to select a maneuver that is not only competitive, but also executable by the downstream limit-aware control module.

The predictive basis of this reasoning process is validated through a real-world wheel-to-wheel racing interaction. As shown in Fig.~\ref{F3}(b), the onboard image sequence captures a short-horizon overtaking process, where the realized trajectories of the ego vehicle and the opponent are compared with the corresponding world-model prediction. The predicted interaction remains consistent with the observed trajectories, indicating that the world model captures the coupled short-horizon evolution of ego and opponent motion. This consistency is critical for decision-making near the cognitive-physical limit, because prediction errors can cause the agent to misjudge whether an overtaking maneuver will remain safe, feasible, and competitive in the next few seconds.

The functional role of the world model for real-time execution is demonstrated by ablation in Fig.~\ref{F3}(c). Under the same racing scenario, future-aware reasoning supported by the world model leads to a feasible overtaking maneuver, whereas removing the world model results in a collision. With the world model, the agent generates multimodal candidate maneuvers in future predicted interactions and evaluates the combined speed and safety scores of the candidates. Collision futures are identified before execution, while the selected candidate achieves the highest score among feasible maneuvers and results in a successful overtaking pose. Without the world model, the agent lacks reliable anticipation of future interaction and physical feasibility, leading to a decision that is plausible from the current observation but unsafe when executed. This comparison shows how world-model prediction directly affects downstream physical execution: insufficient anticipation of interaction evolution can turn a nominally reasonable decision into a physically unsafe maneuver.

Quantitative comparison across benchmark methods further demonstrates the effectiveness of the world-model-driven policy (Ours), as shown in Fig.~\ref{F3}(d). The proposed policy is evaluated in full-scale autonomous racing simulation for adversarial interaction against behavior cloning (BC), soft actor-critic (SAC), and a rule-based planner. The proposed method achieves the highest success rate, reaching $88.3\%$, corresponding to gains of $40.8\%$ over BC and $29.3\%$ over SAC. It also delivers the longest safe mileage, exceeding $6.4~\mathrm{km}$, while maintaining competitive average speed. Moreover, the higher overtaking frequency indicates that the performance gain is not obtained by conservative driving alone, but by more effective interaction-aware decision-making. World-model rollouts screen multimodal trajectory candidates in predicted interaction futures, filtering out maneuvers with safety risks before execution and thereby improving decision reliability. 

These metrics jointly reflect the cognitive side of the limit: the agent must predict interaction evolution accurately enough to distinguish safe, feasible, and competitive maneuvers under severe time pressure.
The results show that world-model rollouts link adversarial interaction reasoning with downstream limit-aware control. By evaluating candidate maneuvers against predicted opponent responses, safety risks, physical feasibility, and efficiency before execution, the agent selects safe actions that are both competitive and executable under joint cognitive-physical limits.

\subsection*{Limit-Aware Control near the Dynamics Boundary}

Limit-aware control tests whether maneuvers selected by future-aware reasoning can be executed safely near the dynamics boundary of the vehicle. In autonomous racing, this boundary is not a fixed speed or acceleration value, but a coupled feasible-motion envelope shaped by vehicle dynamics, actuator constraints, tire--road interaction, aerodynamic loading, road condition, tire temperature, and stability margins. The world model provides a predictive representation of this boundary through ego-dynamics prediction and physical-limit modeling, which regularizes candidate trajectories upstream and constrains the dynamics controller downstream.

\begin{figure*}[!htb]
\centering
\includegraphics[width=1.0\linewidth]{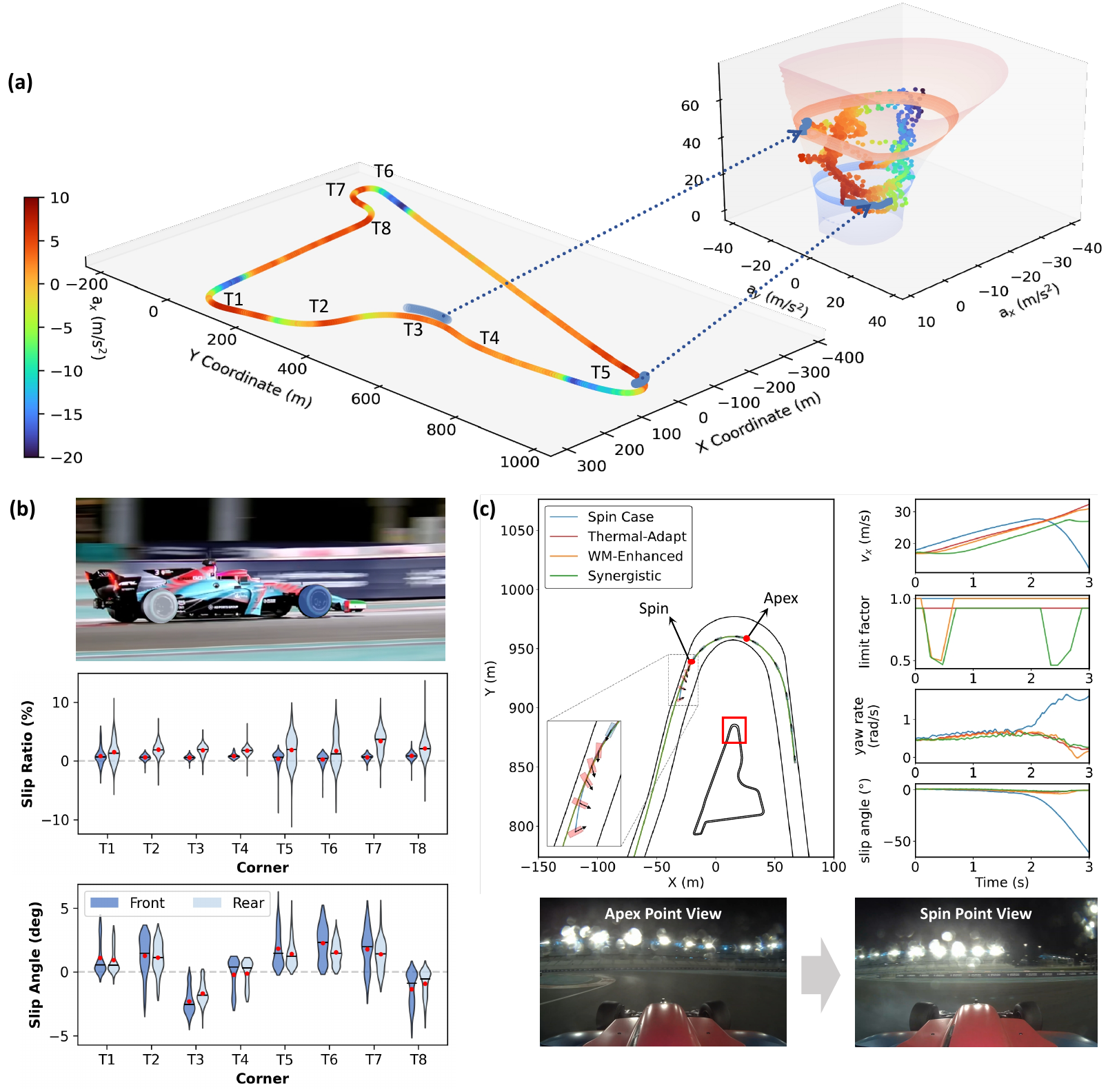}
\caption{\textbf{Limit-aware high-speed dynamics control performance.} World-model-informed physical-limit awareness constrains downstream control and enables aggressive high-speed operation near the feasible dynamics boundary while maintaining stability. (a) Full-lap trajectory and vehicle states in the $a_x$--$a_y$--$v_x$ space, showing operation near the feasible dynamics envelope. (b) Tire-level limit utilization by slip ratio and slip angle. (c) Spin suppression through adaptive degradation of the effective physical limit.} 
\label{F4}
\end{figure*}

Fig.~\ref{F4}(a) shows the full-lap racing trajectory and the corresponding vehicle states projected into the $a_x$--$a_y$--$v_x$ space. This projection provides a vehicle-level view of physical-limit utilization by comparing coupled longitudinal--lateral acceleration states with the feasible dynamics envelope informed by the world model. Along the track, aggressive braking before corners and strong acceleration at corner exits are reflected in the variation of $a_x$, showing active exploitation of the vehicle's coupled motion capability. The autonomous racing vehicle reaches a maximum speed of $71.2~\mathrm{m/s}$ and maintains an average speed of $47.6~\mathrm{m/s}$ over the full lap. The maximum longitudinal acceleration and deceleration reach approximately $1g$ and $3g$, respectively, while the peak lateral acceleration reaches $26~\mathrm{m/s^2}$. 
The vehicle states in Fig.~\ref{F4}(a) indicate that the vehicle motion is competitive while remaining within the feasible motion region.
Within the world model, physical-limit exploration yields a vehicle-level acceleration envelope that characterizes the feasible boundary of coupled longitudinal--lateral motion. 
By imposing this envelope as a systematic constraint of the racing agent, the vehicle motion is regulated to approach the physical boundary for performance exploitation while avoiding persistent excursions beyond the feasible region. 

The vehicle-level envelope is supported by tire-level force generation and closed-loop controllability. As shown in Fig.~\ref{F4}(b), tire utilization across the T1--T8 corners is characterized by slip ratio and slip angle distributions. The slip ratio peaks at approximately $\pm10\%$, while the slip angles of both axles remain within about $\pm5^\circ$, indicating operation close to the stable tire-force generation region without persistent saturation. The downstream controller also maintains accurate tracking and stability during high-speed near-limit racing: the mean absolute lateral tracking error is $0.11~\mathrm{m}$ and the mean absolute speed tracking error is $0.34~\mathrm{m/s}$, with peak values of $0.69~\mathrm{m}$ and $3.23~\mathrm{m/s}$, respectively. The mean absolute yaw-rate discrepancy is $0.073~\mathrm{rad/s}$, with a peak value of $0.525~\mathrm{rad/s}$. These peak errors correspond mainly to short-duration excursions during aggressive braking, cornering, or acceleration, rather than persistent tracking degradation or instability.
The physical-limit acceleration envelope moderates downstream tire-force demand and prevents persistent excursions beyond tire-force limits, thereby enabling safe and high-performance maneuvering near the boundary.

The role of world-model-informed physical-limit-aware adaptation is further highlighted by the failure and recovery scenario in Fig.~\ref{F4}(c). In real vehicle racing, when the safety-car phase ends, a sudden acceleration command near the apex of a high-curvature corner, combined with high lateral load and suboptimal tire temperature, causes abrupt traction loss and spin. To improve robustness near the physical boundary, three limit-degradation mechanisms are evaluated: a direct thermal-adaptive mechanism (Thermal-Adapt), a world-model-based prediction-deviation mechanism (WM-Enhanced), and a synergistic mechanism combining both sources of limit information. With the proposed approach, these mechanisms prevent the vehicle from spinning by reducing the effective physical boundary before instability develops. Thermal-Adapt and WM-Enhanced restrict the peak slip angle to $-2.34^\circ$ and $-4.01^\circ$, respectively, while the synergistic mechanism applies a more conservative fused boundary and further restricts the peak slip angle to $-1.43^\circ$, at the cost of reduced corner-exit performance. 
Physical-limit-aware adaptation therefore improves aggressive driving safety by adjusting the effective acceleration envelope according to thermal state and world-model prediction deviation. 
By contracting the admissible motion boundary before instability fully develops, it suppresses excessive acceleration demand and helps maintain stable vehicle motion during near-limit maneuvers.

Limit-aware control therefore forms the execution layer of the cognitive-physical framework. The world model provides predicted ego dynamics and feasible-motion boundaries, future-aware reasoning selects trajectories under safety, speed, and feasibility criteria, and the controller executes the selected maneuver while respecting tire, stability, and acceleration-envelope constraints. This closes the loop between upstream prediction, competitive decision-making, and near-limit physical execution.

\subsection*{Agent Evolution Through the World Model}

The world model enables the agent to refine its use of coupled cognitive-physical limits over time. Rather than improving racing performance through policy optimization alone, the framework updates predictive models of ego dynamics, interaction evolution, and physical feasibility, and uses them to improve both future-aware reasoning and limit-aware control. Agent evolution is therefore reflected not only in faster lap times, but also in expanded physical-limit utilization, safer competitive interaction, and adaptability across operating conditions.

\begin{figure*}[!htb]
\centering
\includegraphics[width=1.0\linewidth]{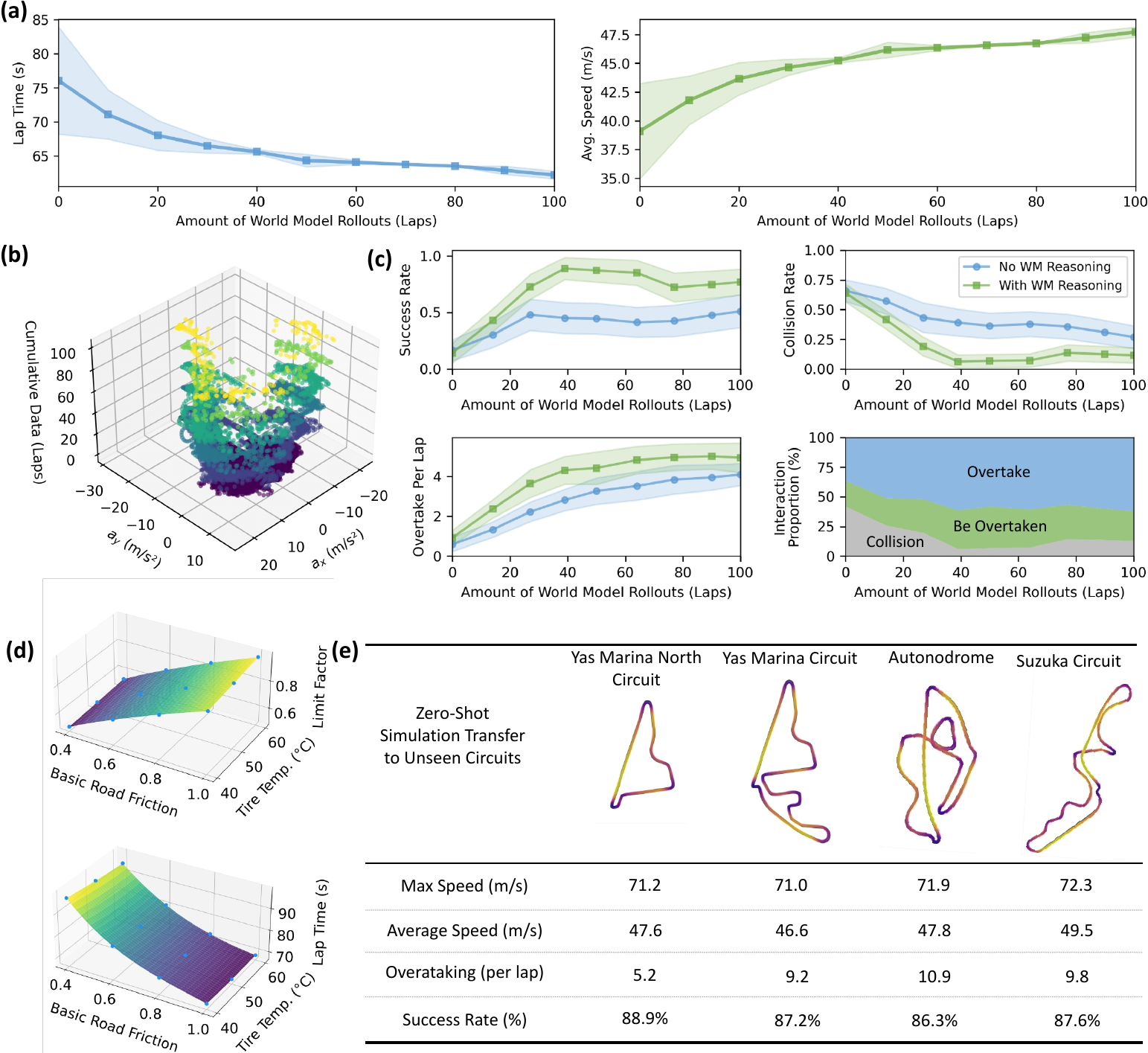}
\caption{\textbf{Agent evolution through the world model.} Closed-loop refinement improves racing performance, interaction capability, and robustness across conditions. 
(a) Improved racing performance through cumulative world-model data. 
(b) Progressive expansion of reachable states in the $a_x$--$a_y$--$v_x$ space during agent evolution. 
(c) Improved interaction performance with increasing world-model rollouts. 
(d) Adaptation under varying tire temperatures and road friction. 
(e) Zero-shot transfer to unseen circuits.}
\label{F5}
\end{figure*}

The physical side of world-model-driven agent evolution is reflected in the physical boundary expansion and lap-time reduction. Fig.~\ref{F5}(a) shows that lap time decreases while average speed increases as cumulative world-model data grows. Across the refinement process, the average lap time improves by $17.87\%$, while the average speed increases by $22.02\%$. This improvement is accompanied by expansion of the reachable acceleration states. As shown in Fig.~\ref{F5}(b), the distribution of vehicle states in the longitudinal--lateral acceleration space expands toward the feasible acceleration envelope, indicating that the agent exploits a larger portion of the vehicle's physical capability while maintaining feasible and stable motion.

The cognitive side of this evolution appears in improved interaction reasoning and decision-making. As shown in Fig.~\ref{F5}(c), reinforcement learning based on world-model rollouts generates interactive scenarios that are scarce in real racing, allowing the agent to improve competitive behavior under adversarial interaction. With accumulated world-model refinement, the interaction success rate increases from $42.6\%$ to $88.9\%$, reaching a peak of $91.7\%$; the average overtaking count per lap increases from $2.78$ to $4.80$; and the collision rate per interaction decreases from $41.9\%$ to $5.8\%$, with a minimum of $4.0\%$. Compared with the ablated variant without future-aware reasoning, the proposed method improves success rate by $34.7\%$, increases overtaking frequency by $23.4\%$, and reduces collision rate by $19.8\%$. These improvements indicate that the refined world model helps the agent distinguish actions that are safe, feasible, and competitive under interaction uncertainty.

Fig.~\ref{F5}(d) further shows how cognitive and physical limits are jointly regulated under changing operating conditions. The ego limit factor is adaptively adjusted under different tire temperatures and road friction levels, and the corresponding lap-time surface reflects how performance varies as the available physical boundary changes. Instead of treating the physical limit as fixed, the agent adapts the effective boundary to current conditions so that planning and control remain compatible with the vehicle's available capability.

The evolved agent also generalizes across unseen circuits in simulation, as shown in Fig.~\ref{F5}(e). After training on the Yas Marina North Circuit, the agent was transferred without additional training to the Yas Marina Full GP layout, Autonodrome, and Suzuka Circuit. These circuits differ in track length, curvature distribution, corner density, high-speed straight sections, and overtaking opportunities, thereby testing whether the refined agent preserves cognitive reasoning and physical-limit-aware execution under changed track geometry. Maximum speeds remain around $70~\mathrm{m/s}$ across all circuits, while overtaking frequency varies with track characteristics. The success rate in $20~\mathrm{s}$ random interactive scenarios remains close to $90\%$, indicating consistent interaction performance in zero-shot transfer. These results show that world-model refinement supports the co-evolution of cognitive reasoning and physical-limit utilization, enabling the agent to better predict interaction futures, select competitive yet feasible maneuvers, exploit the dynamics envelope, and adapt the effective physical boundary across conditions.

\section*{Discussion}
This work uses autonomous racing as an extreme testbed for exploring the cognitive-physical limits of embodied intelligence. The goal is not merely to develop a faster racing system, but to investigate how an embodied agent can construct, predict, exploit, and regulate its capability boundaries under simultaneous cognitive and physical constraints. In this setting, cognitive limits arise from high-frequency localization and perception, interaction prediction, and decision-making under severe time pressure, while physical limits arise from vehicle dynamics, actuator constraints, tire--road interaction, thermal variation, and stability margins. The results show that these limits are coupled: uncertainty in perception or interaction prediction affects how safely the physical envelope can be exploited, while the evolving physical state constrains which future actions remain feasible and safe.

A central implication is that world models can serve as a bridge between cognitive reasoning and physical control in extreme embodied tasks. In the proposed racing agent, the world model jointly predicts external interaction evolution, ego dynamics, and feasible motion boundaries. This allows candidate maneuvers to be evaluated before execution in terms of opponent response, safety risk, physical feasibility, and speed efficiency. As shown by the interaction-reasoning and dynamics-control results, this predictive structure transforms adversarial reasoning into executable near-limit maneuvers while constraining downstream control near the feasible dynamics envelope.

The results also suggest that embodied capability can evolve through closed-loop refinement of world modeling, decision-making, and physical-limit regulation. As the world model is refined, the agent improves lap time, average speed, acceleration-state coverage, interaction success, overtaking capability, collision avoidance, and adaptation to changing tire and road conditions. This improvement is therefore not a single-dimensional performance gain, but a coupled evolution of cognitive reasoning and physical-limit utilization.

Beyond autonomous racing, the present work suggests a boundary-aware methodology for embodied intelligent systems. For robots operating in the real world, safety and performance depend not only on selecting desirable actions, but also on whether those actions remain predictable, feasible, and safe under current sensing, interaction, and physical constraints. This perspective shifts the design objective from optimizing isolated modules toward modeling and regulating coupled system-level boundaries. High-speed mobile robots, agile aerial vehicles, legged robots on risky terrain, human--robot collaborative systems, and manipulation systems with rich contact dynamics all involve different forms of such boundary awareness: localization uncertainty can reduce control margins, inaccurate prediction of human or agent motion can make a feasible action unsafe, and changing contact, friction, or load conditions can invalidate a previously reasonable plan.

Several limitations remain. The current framework is developed and validated in autonomous racing, where task structure, vehicle dynamics, track geometry, and interaction patterns are more constrained than in many open-world robotic applications. Extending this approach will require richer world representations, broader uncertainty modeling, and stronger mechanisms for semantic ambiguity, human intent, deformable contact, and unmodeled environmental changes. Moreover, the results do not establish an absolute theoretical upper bound of embodied intelligence; rather, they operationalize cognitive and physical limits through observable indicators such as localization robustness, interaction prediction, decision safety, dynamic-envelope utilization, tire-level behavior, and stability margins. Overall, this work suggests that cognitive and physical limits should be treated as coupled system-level design objects rather than isolated module-level constraints, and that world-model-centric agents provide a concrete path toward representing, exploiting, and regulating these limits for safer real-world deployment.

\section*{Methods}

\subsection*{System Formulation under Cognitive-Physical Limits}

The proposed system is formulated as a world-model-centric architecture for autonomous racing near coupled cognitive-physical limits. The agent integrates current-state construction, predictive reasoning, and motion control through a shared world model. At runtime, the world model predicts both interaction evolution and ego dynamics, allowing candidate maneuvers to be evaluated against future safety, racing efficiency, and physical feasibility before execution. Over longer time scales, real racing data refine the world model, and the refined world model provides structured predictions for updating decision-making, planning, and limit-aware control.

At time step $t$, the agent receives surrounding observation $o_t$, ego vehicle state $x_t$, and local track information $\Gamma_t$. These inputs are integrated into a structured world-state representation
\begin{equation}
w_t = \Psi_w(x_t,o_t,\Gamma_t),
\end{equation}
where $\Psi_w(\cdot)$ denotes the state representation function.
Here, $w_t$ contains not only the basic ego, opponent, and track information, but also evaluation-related information that can be derived from the structured scene, such as collision status, boundary violation, and physical feasibility.
Using the world model $W$, the future evolution of the world state $\hat{w}_{t:t+H}$ is predicted over a finite horizon $H$.

Based on the current world state, the decision policy $\pi$ generates a main racing decision. Together with a transformer-based multimodal planner, this decision produces a set of candidate trajectories, 
\begin{equation}
\mathcal{T}_t = \left\{ \tau_{t}^k \right\}_{k=1}^{K},
\end{equation}
where $K$ is the number of candidate trajectories over the planning horizon $H$.
Each candidate trajectory is propagated through the world model to generate a corresponding future. The candidate set is then evaluated using a task-oriented score that jointly considers interaction safety, physical feasibility, and speed efficiency.  The optimal trajectory \( \tau^* \) is then selected from $\mathcal{T}_t$ based on their future-evaluation scores. This formulation allows the agent to choose maneuvers that are competitive in the racing task while remaining safe and physically executable.

Given the selected trajectory $\tau_t^\ast$, the motion-control module computes executable commands by solving a constrained optimization problem,
\begin{equation}
\label{Eq_control_opt}
u_t^\ast =
\arg\min_{u_{t:t+H_c-1}}
J(x_t,u_{t:t+H_c-1};\tau_t^\ast,\mathcal{E}_t),
\end{equation}
subject to
\begin{equation}
\label{Eq_control_con1}
x_{t+j+1} = f_{\mathrm{dyn}}(x_{t+j}, u_{t+j}),
\quad j=0,\dots,H_c-1,
\end{equation}
\begin{equation}
\label{Eq_control_con2}
h(x_{t+j}, u_{t+j}, \mathcal{E}_{t+j}) \le 0, \quad j = 0, \dots, H_c-1,
\end{equation}
where $x_t$ is the ego vehicle state, $u_t=[a_x,\delta]^\top$ contains longitudinal acceleration and steering commands, $f_{\mathrm{dyn}}(\cdot)$ is the control-oriented vehicle dynamics model, $\mathcal{E}_{t+j}$ denotes the feasible-motion envelope estimated from the world model, and $h(\cdot)$ enforces coupled longitudinal--lateral motion feasibility. The reference state $x_{t+j}^{\mathrm{ref}}$ is induced by the selected trajectory $\tau_t^\ast$ and enters the tracking objective in $J(\cdot)$. In this way, the downstream controller executes the selected maneuver while respecting the current physical capability of the ego vehicle.

Beyond online decision-making and control, the architecture supports closed-loop refinement of the world model and the policy. Real racing data $\mathcal{D}_{\mathrm{real}}$ are used to train the world model through supervised prediction of ego dynamics, interaction evolution, and physical feasibility. The refined world model then generates simulated rollouts $\mathcal{D}_{\mathrm{wm}}$ under controlled variations of interaction and near-limit operating conditions. The combined dataset $\mathcal{D}$ is used to update the decision and planning models and to refine physical-limit constraints. This closed-loop process links real racing data, world-model prediction, policy improvement, and limit-aware control, enabling progressive improvement under coupled cognitive and physical limits.

\subsection*{Predictive World Modeling for the Autonomous Racing Agent}

The world model is implemented as a structured predictive module of the autonomous racing agent operating near cognitive-physical limits. It supports online future-aware reasoning and limit-aware control, while also enabling offline policy refinement through structured internal inference. Starting from the world-state representation $w_t$, which integrates the ego vehicle state, surrounding opponents, and local track context, the world model is composed of three predictive components: an ego dynamics model, an interaction prediction model, and a physical-limit representation. These components jointly predict how the ego vehicle, the surrounding racing scene, and the feasible motion boundary evolve under near-limit conditions. 

The ego dynamics model provides short-horizon prediction of vehicle motion under aggressive control inputs. It adopts a physics-constrained residual-learning formulation, in which the next-step state is represented by a differentiable dynamics backbone with residual correction:
\begin{equation}
\hat{x}_{t+1}
=
f_{\mathrm{dyn}}
\!\left(x_t,u_t;\theta_\mathrm{dyn}\right)
+
f_{\mathrm{res}}
\!\left(x_{t-h:t},u_{t-h:t};\theta_\mathrm{res}\right),
\label{eq:wm_dyn}
\end{equation}
where $f_{\mathrm{dyn}}$ provides a structured state transition and $f_{\mathrm{res}}$ captures unmodeled nonlinear effects. The physics backbone preserves the main coupling among longitudinal motion, lateral motion, yaw dynamics, and wheel rotational behavior, while the dynamics-related parameters $\theta_\mathrm{dyn}$ are constrained within physically plausible ranges through bounded learnable parameters. The residual model, parameterized by $\theta_\mathrm{res}$, further refines the prediction using recent state and control histories. This formulation allows the model to capture transient nonlinearities that are difficult to analytically describe while retaining physical coherence under near-limit maneuvering.

The interaction prediction model captures the evolution of multi-agent racing scenarios. Given the structured world state, surrounding opponents are represented by their relative spatial relations to the ego vehicle and organized into position-aware slots. An attention-based interaction encoder, consisting of multilayer perceptrons followed by a masked attention-pooling layer, embeds this relational context and predicts a latent interaction intention for each agent. Racing-relevant behaviors, including attack, defense, hold, and yield, are implicitly represented in this latent intention space. At inference time, the world model is recursively rolled out to generate future interaction scenarios:
\begin{equation}
\label{Eq_wm_prediction}
\hat{w}_{t:t+H} = W^H(w_t, a_{t:t+H-1}),
\end{equation}
where $W^H(\cdot)$ denotes recursive world-model prediction over horizon $H$, $a_{t:t+H-1}$ denotes candidate action inputs or behavior hypotheses, and $\hat{w}_{t:t+H}$ denotes the predicted future world states. These predicted futures allow the agent to evaluate candidate trajectories against possible opponent responses rather than relying only on the current observation. 

The physical-limit representation describes the feasible motion boundary of the ego vehicle. In autonomous racing, this boundary is shaped by vehicle dynamics, actuator constraints, tire--road interaction, aerodynamic loading, road friction, tire temperature, and stability margins, and is manifested at the vehicle level as a coupled longitudinal--lateral acceleration envelope. The envelope is represented as 
\begin{equation}
\begin{cases}
\left(\dfrac{a_{x,k}}{a_{x,\max}^{\mathrm{acc}}(v_{x,k},\mu_k)}\right)^c
+
\left(\dfrac{|a_{y,k}|}{a_{y,\max}(v_{x,k},\mu_k)}\right)^c
\le 1,
& a_{x,k}\ge 0, \\[8pt]
\left(\dfrac{|a_{x,k}|}{a_{x,\max}^{\mathrm{dec}}(v_{x,k},\mu_k)}\right)^c
+
\left(\dfrac{|a_{y,k}|}{a_{y,\max}(v_{x,k},\mu_k)}\right)^c
\le 1,
& a_{x,k}<0.
\end{cases}
\label{eq_envelope_limit}
\end{equation}
Here $a_{x,k}$ and $a_{y,k}$ denote longitudinal and lateral accelerations, $v_{x,k}$ is the longitudinal speed, $\mu_k$ is the road-friction coefficient, and $c$ controls the shape of the coupled envelope. The acceleration limits $a_{x,\max}^{\mathrm{acc}}$, $a_{x,\max}^{\mathrm{dec}}$, and $a_{y,\max}$ describe speed- and friction-dependent acceleration, braking, and lateral capabilities.

A nominal envelope $\mathcal{E}_{0}$ is induced from the dynamics parameters $\theta_\mathrm{dyn}$ identified by the differentiable dynamics model, providing a structured prior of the vehicle's motion capability. Because the effective physical boundary varies with operating conditions and model mismatch, physical-limit exploration and online adaptation are introduced to scale the nominal envelope:
\begin{equation}
\mathcal{E}_{t}
=
\mathcal{E}(\rho_a(w_t),\rho_e;\mathcal{E}_{0}(\theta_\mathrm{dyn})),
\label{eq:adaptive_envelope}
\end{equation}
The exploration vector $\rho_e$ enlarges the envelope from a conservative initialization as near-limit racing data accumulate, while the adaptive factor $\rho_a(w_t)$ modulates the envelope online according to road friction, tire temperature, and world-model prediction error. This allows the effective envelope $\mathcal{E}_t$ to approach the available physical capability without exceeding the prescribed limit criterion.

Through these three components, the world model provides a predictive representation of both cognition-relevant and physics-relevant future evolution. Ego dynamics prediction estimates how the vehicle responds near the physical boundary; interaction prediction anticipates how opponents may react; and physical-limit representation determines which future motions remain feasible. This predictive representation supplies the basis for future-aware reasoning, limit-aware control, and closed-loop refinement of the agent.

\subsection*{Quantifying Cognitive and Physical Limits}

The cognitive and physical limits in this work are quantified through observable indicators rather than treated as absolute theoretical bounds. These indicators measure how the agent constructs the current racing state, predicts future evolution, selects competitive actions, and executes them near the feasible motion boundary. Detailed indicator definitions, normalization thresholds, and implementation settings are provided in Supplementary Methods Section S3.

The cognitive limit is evaluated along the perception--prediction--decision chain. Current-state construction is quantified by localization and ego-state residuals,
\begin{equation}
\mathbf{e}^{\mathrm{state}}_t =
\left[
|e_{x,t}|,\ |e_{y,t}|,\ |e_{v_x,t}|,\ |e_{v_y,t}|,\ |e_{\psi,t}|
\right],
\end{equation}
where $e_{x,t}$ and $e_{y,t}$ are longitudinal and lateral position residuals, $e_{v_x,t}$ and $e_{v_y,t}$ are velocity residuals, and $e_{\psi,t}$ is the heading residual. Opponent perception is further assessed by the temporal consistency of tracked opponent states. Predictive capability is evaluated by ego-dynamics prediction errors and interaction-prediction accuracy. For short-horizon ego--opponent interaction prediction, average displacement error and final displacement error are defined as
\begin{equation}
\mathrm{ADE} =
\frac{1}{T}
\sum_{i=1}^{T}
\left\|
\hat{o}_{t+i} - o_{t+i}
\right\|_2,
\end{equation}
\begin{equation}
\mathrm{FDE} =
\left\|
\hat{o}_{t+T} - o_{t+T}
\right\|_2,
\end{equation}
where $\hat{o}_{t+i}$ and $o_{t+i}$ denote the predicted and ground-truth opponent states at prediction step $i$. Closed-loop decision capability is measured by interaction success rate, collision rate, safe mileage, and overtaking frequency. The interaction success rate is defined as
\begin{equation}
R_{\mathrm{succ}} =
\frac{N_{\mathrm{succ}}}{N_{\mathrm{total}}},
\end{equation}
where $N_{\mathrm{succ}}$ the number of successfully completed interaction trials and $N_{\mathrm{total}}$ is the total number of interaction trials.
These metrics characterize whether the agent can distinguish maneuvers that are safe, feasible, and competitive under severe time pressure and interaction uncertainty.

The physical limit is evaluated by how closely the ego vehicle approaches and maintains feasible motion under near-limit execution. At the vehicle level, acceleration-envelope utilization is defined by the normalized envelope function
\begin{equation}
\eta_{\mathrm{env},t}
=
g_{\mathcal{E}}
\left(a_{x,t},a_{y,t},v_{x,t},\mu_t;\mathcal{E}_t\right),
\end{equation}
where $g_{\mathcal{E}}(\cdot)$ is induced by the coupled acceleration envelope in Eq.~(\ref{eq_envelope_limit}). A value $\eta_{\mathrm{env},t}\leq 1$ indicates that the executed longitudinal--lateral acceleration remains inside the feasible envelope, while values close to one indicate operation near the physical boundary.

To capture physical-limit phenomena across different layers, we further define a hybrid physical-limit utilization,
\begin{equation}
\eta_{\mathrm{lim},t}
=
\max
\left(
\eta_{\mathrm{tire},t},
\eta_{\mathrm{control},t},
\eta_{\mathrm{stability},t}
\right),
\end{equation}
where $\eta_{\mathrm{tire},t}$ is computed from tire slip ratio and slip angle, $\eta_{\mathrm{control},t}$ from lateral and speed tracking errors, and $\eta_{\mathrm{stability},t}$ from yaw-rate discrepancy. These terms respectively characterize tire--road force utilization, trajectory tracking capability, and maneuvering stability. 
The metric $\eta_{\mathrm{env},t}$ further represents $\eta_{\mathrm{lim},t}$ with the feasible envelope, thereby quantifying the proximity of the current state to the feasible motion boundary.
Their detailed normalized forms are provided in Supplementary Methods.

The cognitive and physical indicators are interpreted jointly. Localization and prediction errors quantify cognitive uncertainty; acceleration-envelope utilization, tire behavior, tracking errors, and yaw-rate discrepancy quantify physical-limit utilization; and closed-loop outcomes such as success rate, collision rate, safe mileage, and overtaking frequency reveal how the two sides interact during execution. A maneuver is effective only when it is strategically competitive, predicted to be safe, and executable within the current feasible-motion envelope.

\subsection*{World-Model-Driven Future-Aware Reasoning}

Future-aware reasoning integrates decision-making, multimodal trajectory generation, and predictive evaluation around the world model. The decision and planning models are trained offline with real and world-model-generated data, while online deployment uses world-model rollouts to evaluate candidate maneuvers before execution.

At runtime, the structured world state $w_t$ is input to a decision policy $\pi$, which outputs a compact racing intention containing longitudinal and lateral targets. This decision is converted into a main reference trajectory through a transition model:
\begin{equation}
\tau^{\mathrm{ref}}_t = f_{\mathrm{tran}}(x_t,\pi(w_t;\theta_\pi), H),
\end{equation}
where $\theta_\pi$ denotes the parameters of the decision policy, $f_{\mathrm{tran}}(\cdot)$ is the transition model, and $H$ is the planning horizon. The decision policy uses a structured interaction encoder that represents the ego vehicle, surrounding opponents, and relative racing context, and outputs longitudinal--lateral intentions for downstream trajectory generation.

To capture multiple possible interaction strategies, a multimodal delta planner generates trajectory deviations from recent world-state history. These deviations are added to the main reference trajectory and regularized by the current feasible-motion envelope $\mathcal{E}_t$:
\begin{equation}
\label{Eq_planning_out}
\mathcal{T}_t =
R\!\left(
\tau_t^{\mathrm{ref}} + \phi(w_{t-h:t}; \theta_{\phi}),
\mathcal{E}_t
\right),
\end{equation}
where $\phi(\cdot)$ is the delta trajectory planning network, $\theta_{\phi}$ denotes its parameters, $h$ is the history length, and $R(\cdot,\mathcal{E}_t)$ denotes physical-limit regularization. The planner encodes ego history, opponent history, and local track context with transformer-based modules, and outputs multimodal candidate trajectories representing different competitive behaviors. 
These multimodal trajectories are further combined with prescribed speed factors to diversify longitudinal behavior. The regularization step ensures that candidate trajectories are compatible with the current physical capability of the ego vehicle before predictive evaluation.

Each candidate trajectory is then evaluated through recursive world-model rollouts. Following Eq.~(\ref{Eq_wm_prediction}), the world model predicts the future world-state evolution induced by each candidate, yielding a set of parallel futures:
\begin{equation}
\mathcal{F}_t
=
\left\{
\left(\tau_t^k,\hat{w}_{t:t+H}^{\,k}\right)
\mid
\tau_t^k\in\mathcal{T}_t
\right\}.
\end{equation}
Each parallel future is scored by a task-oriented function that considers interaction safety, physical feasibility, and speed efficiency. Candidate trajectories that lead to predicted collision, boundary violation, or loss of control are rejected, and the selected trajectory is
\begin{equation}
\tau_t^{*}
=
\arg\max_{\tau_t^k\in\mathcal{T}_t}
S\!\left(\tau_t^k,\hat{w}_{t:t+H}^{\,k}\right).
\label{Eq_planning_selection}
\end{equation}

This procedure selects trajectories that are safe, feasible, and competitive before execution. By combining physical-limit regularization with world-model rollout evaluation, future-aware reasoning links adversarial interaction prediction to downstream limit-aware control.

\subsection*{Physical-Limit-Constrained Motion Control}

The motion-control module converts the trajectory selected by future-aware reasoning into executable commands under the current physical-limit envelope. Given the selected trajectory $\tau_t^\ast$ and the world-model-informed feasible-motion envelope $\mathcal{E}_t$, control is formulated as a constrained model predictive control problem. The controller tracks $\tau_t^\ast$ while enforcing vehicle dynamics, road-boundary constraints, actuator limits, and the coupled longitudinal--lateral acceleration constraint induced by $\mathcal{E}_t$. The detailed vehicle dynamics model, Jacobians, and barrier derivatives are provided in Supplementary Methods Section S6.

At each control step, the controller solves the constrained optimization problem described in Eq. (\ref{Eq_control_opt})--Eq. (\ref{Eq_control_con2}). 
Given the selected trajectory $\tau_t^\ast$, it optimizes a finite-horizon control sequence to track the induced reference states while satisfying the control-oriented vehicle dynamics and the feasible-motion constraints defined by the physical-limit envelope. The control input $u_t=[a_x,\delta]^\top$ includes the longitudinal acceleration and steering command, and the envelope $\mathcal{E}_{t+j}$ constrains the coupled longitudinal--lateral feasibility estimated from the world model. This formulation enables the selected maneuver to be executed as physically feasible control commands under the current capability of the ego vehicle.

The cost is decomposed into a tracking-and-smoothness term and a constraint-barrier term,
\begin{equation}
J =
J_{\mathrm{track}}(x_t,u_{t:t+H_c-1};\tau_t^\ast)
+
J_{\mathrm{barrier}}(x_t,u_{t:t+H_c-1};\mathcal{E}_t).
\end{equation}
The tracking term is
\begin{equation}
J_{\mathrm{track}} =
\sum_{j=0}^{H_c-1}
\left(
\|x_{t+j}-x_{t+j}^{\mathrm{ref}}\|_Q^2
+
\|u_{t+j}\|_R^2
+
\|\Delta u_{t+j}\|_P^2
\right),
\end{equation}
where $x_{t+j}^{\mathrm{ref}}$ is induced by $\tau_t^\ast$. Inequality constraints are imposed through exponential barriers,
\begin{equation}
J_{\mathrm{barrier}}
=
\sum_{k,i}
q_1^{(i)}
\exp\!\left(q_2^{(i)}c_i(x_k,u_{k-1};\mathcal{E}_k)\right),
\end{equation}
where $c_i\leq0$ indicates satisfaction of the corresponding constraint and $c_i>0$ indicates violation. The physical-limit barrier is induced by the coupled acceleration envelope in Eq.~(\ref{eq_envelope_limit}), so violations of the current feasible-motion boundary are penalized directly during optimization.

The constrained optimization is solved using a constrained iterative linear quadratic regulator (CiLQR). At each iteration, the dynamics and costs are locally approximated, a backward pass computes the feedforward correction and feedback gain, and a forward rollout tests the updated control sequence. The update takes the affine feedback form
\begin{equation}
u_k^{\mathrm{new}}
=
u_k+\alpha d_k+\gamma K_k(x_k^{\mathrm{new}}-x_k),
\end{equation}
where $d_k$ is the feedforward correction, $K_k$ is the feedback gain, $\alpha$ is the line-search parameter, and $\gamma$ scales the feedback term. This receding-horizon formulation allows the controller to regulate speed and steering before the vehicle violates the physical-limit envelope, thereby executing the selected maneuver under near-limit conditions.

\subsection*{Closed-Loop World-Model and Policy Refinement}

The world model and the policy are refined through a closed-loop process that links real racing data, predictive model learning, internal simulation, and policy update. 
During real-world racing, the onboard system collected high-quality dynamic data under real road, tire, vehicle, and interaction conditions, forming a dataset $\mathcal{D}_{\mathrm{real}}$.
These data are used in model training to align the world model with observed ego dynamics, opponent interactions, and physical-limit behavior.

The ego dynamics model and interaction model are updated through supervised prediction. The differentiable dynamics backbone and residual model are trained by minimizing the next-state prediction error:
\begin{equation}
(\theta_{\mathrm{dyn}}, \theta_{\mathrm{res}})
=
\arg\min_{\theta_{\mathrm{dyn}}, \theta_{\mathrm{res}}}
\sum_{\mathcal{D}_{\mathrm{real}}}
\left\|
x_{t+1}
-
f_{\mathrm{dyn}}
\!\left(x_t,u_t;\theta_\mathrm{dyn}\right)
-
f_{\mathrm{res}}
\!\left(x_{t-h:t},u_{t-h:t};\theta_\mathrm{res}\right)
\right\|^2 .
\label{eq:update_dyn_res}
\end{equation}
The interaction model is updated by supervised prediction of world-state evolution:
\begin{equation}
\theta_{\mathrm{int}}
=
\arg\min_{\theta_{\mathrm{int}}}
\sum_{\mathcal{D}_{\mathrm{real}}}
\left\|
w_{t+1}
-
W(w_t;\theta_{\mathrm{int}})
\right\|^2 ,
\label{eq:update_int}
\end{equation}
where $W(\cdot;\theta_{\mathrm{int}})$ denotes the learned interaction transition model. The physical-limit model is initialized conservatively and refined through world-model-based limit exploration. At each update step, a safe exploration policy proposes an envelope expansion based on local track context and limit-utilization information. The expansion is accepted only when the predicted rollout with the updated envelope satisfies the prescribed limit criterion, enabling progressive exploration of the feasible motion boundary without violating safety constraints.

The refined world model is then used to generate structured rollout data under controlled variations of interaction and near-limit operating conditions:
\begin{equation}
\mathcal{D}_{\mathrm{wm}}
=
\mathrm{Rollout}(W, \pi, \phi, \mathcal{E}) .
\end{equation}
Compared with real racing data alone, $\mathcal{D}_{\mathrm{wm}}$ can cover a broader range of adversarial interactions, corner cases, and near-limit scenarios. The real and world-model-generated datasets are combined as
\begin{equation}
\mathcal{D}
=
\mathcal{D}_{\mathrm{real}}
\cup
\mathcal{D}_{\mathrm{wm}},
\end{equation}
which is used to update the decision policy and multimodal planner.

The decision policy is updated using a soft actor--critic objective with conservative Q-learning and behavior-cloning regularization. Given state $s_t$, corresponding to the world state $w_t$, the policy samples a racing intention $a_t^\pi\sim\pi_{\theta_{\pi}}(\cdot|s_t)$ containing longitudinal and lateral decision components. The critic parameters are optimized by
\begin{equation}
\begin{aligned}
L_Q(\theta_{Q})
=
&\mathbb{E}_{(s_t,a_t,s_{t+1})\sim\mathcal{D}}
\left[
\left(Q_{\theta_Q}(s_t,a_t)-y_t\right)^2
\right] \\
&+
\lambda_{\mathrm{cql}}
\left\{
\mathbb{E}_{s_t\sim\mathcal D}
\left[Q_{\theta_Q}(s_t,a_t^\pi)\right]
-
\mathbb{E}_{(s_t,a_t)\sim\mathcal D}
\left[Q_{\theta_Q}(s_t,a_t)\right]
\right\},
\end{aligned}
\label{eq_cql_loss}
\end{equation}
and the policy is optimized by
\begin{equation}
\begin{aligned}
L_\pi(\theta_{\pi})
=
\mathbb{E}_{(s_t,a_t^{\mathrm{e}})\sim\mathcal{D}}
\Big[
&\alpha \log \pi_{\theta_{\pi}}(a_t^\pi|s_t)
-
\min(Q_1,Q_2) 
+
\lambda_{\mathrm{bc}}
\left\|
a_t^\pi-a_t^{\mathrm{e}}
\right\|^2
\Big],
\end{aligned}
\label{eq:update_dec}
\end{equation}
where $a_t^{\mathrm{e}}$ denotes the demonstration action in the dataset. The conservative term discourages overestimation on out-of-distribution actions, while the behavior-cloning term preserves useful behavior from real vehicle data and simulated data.

The multimodal planner is updated using best-of-$K$ supervision. Hybrid data are reorganized into pairs of world-state history and future trajectory. Given the historical segment $w_{t-h:t}$, the planner generates candidate trajectories as in Eq.~(\ref{Eq_planning_out}), and only the output mode closest to the ground-truth future trajectory is supervised:
\begin{equation}
\theta_\phi
=
\arg\min_{\theta_\phi}
\sum_{\mathcal{D}}
\left[
\min_{\tau_t^k \in \mathcal{T}_t}
\left\|
\tau_{t}^{\,\mathrm{gt}}-\tau_t^k
\right\|^2
\right],
\label{eq_planning_update}
\end{equation}
where $\tau_{t}^{\mathrm{gt}}$ denotes the ground-truth future trajectory corresponding to the segment $w_{t:t+H}$ extracted from $\mathcal{D}$. This training strategy allows different output heads to specialize in different behavioral modes.

This refinement process closes the loop between real-world deployment and world-model-based learning. Real racing data improve the fidelity of ego-dynamics, interaction, and physical-limit modeling; the refined world model generates structured rollouts for rare interactions and near-limit conditions; and the updated policy and planner collect more informative data in subsequent deployment. Through this cycle, the agent progressively improves future prediction, interaction reasoning, and physical-limit utilization under coupled cognitive-physical limits.

\subsection*{Experimental Platform and Validation}

A full-scale autonomous racing platform was used for collection of real-vehicle data collection with physics grounding. Real-vehicle deployment of the base software stack exposed the agent to extreme racing conditions and provided high-value data for world-model refinement. The racing task included high-speed operation, near-limit tire--road conditions, and adversarial multi-agent interactions, making it a stringent source of real-world data for studying embodied intelligence near cognitive and physical limits.

During real-vehicle experiments, state, control, and environment data were collected from the onboard system. These real-world racing data were used to train and refine the world model, including ego dynamics, interaction evolution, and physical-limit representation. Rather than treating real-world deployment as the final validation of the proposed method, the collected data were utilized and incorporated into a closed-loop data-driven refinement pipeline: competition deployment generated real racing data, the data updated the world model, and the refined world model supported subsequent policy improvement.

To evaluate the agent updated with the refined world model under controlled and repeatable interactive conditions, validation was conducted in the Autoverse simulator with virtual opponent vehicles. This setup provides high-fidelity vehicle dynamics simulation, allowing the updated agent to be tested in adversarial multi-vehicle scenarios with repeatable initial conditions. Interaction scenarios of $20~\mathrm{s}$ duration were constructed to reproduce representative competitive situations, with opponent behaviors generated by imitation learning from real racing interactions. 
In each trial, 2 to 4 opponent vehicles with racing behaviors were generated within a longitudinal range extending $80~\mathrm{m}$ ahead of and $50~\mathrm{m}$ behind the ego vehicle, with their lateral offsets sampled from $\mathcal{N}(0,2.5^2$).
A trial is considered successful if the ego vehicle reaches the end of the 20 s evaluation horizon without collision, track-boundary violation, or unrecoverable instability; otherwise, it is considered failed.

The proposed method was compared with three baselines: behavioral cloning (BC), soft actor--critic (SAC), and the state-machine-based rule planner used in the competition stack. BC and SAC serve as representative imitation-learning and model-free reinforcement-learning baselines, while the rule-based planner serves as the baseline corresponding to the real-vehicle competition stack.
All learning-based approaches were trained using the same training data.
Each method was evaluated in 360 trials, of which two were excluded due to simulation failures, leaving 358 valid trials.

Beyond statistical performance comparison, the validation further reproduced overtaking and spin scenarios observed in real-world racing. The overtaking case was used to assess whether the world model could predict short-horizon interaction evolution during wheel-to-wheel racing. The spin scenario was used to evaluate whether world-model-informed physical-limit-aware adaptation could correct a failure mode observed during the race by improving recovery and reducing instability. This validation pipeline links real-world deployment, data collection, world-model refinement, policy testing in simulation, and failure-case reproduction, thereby supporting continuous agent improvement under coupled cognitive-physical limits.

\bibliography{sn-bibliography}

\subsubsection*{Acknowledgements}
We would like to thank K2 Holding, Abu Dhabi, UAE, for its project support, and Alp Autonomy, Turkey, for its technical support. These contributions were instrumental in the implementation and experimental validation of this work. We also thank the members of the NTU AutoMan Research Lab and the former members of the Kinetiz autonomous racing team, particularly Binbin Hu, Wenhui Huang, Haohan Yang, Haochen Liu, Yiran Zhang, Tianchu Su, Andrea Piazzoni, Xiangkun He, Xiaoyu Mo, and Xinyu Zhou, for their valuable contributions to system development, vehicle integration, testing, and experimental deployment.

\subsubsection*{Funding}
This work was supported by K2 Holding L.LC, Abu Dhabi, UAE, under Research Collaboration Agreement (No. REQ0474834).

\subsubsection*{Competing interests}
The authors declare no competing interests.

\subsubsection*{Data availability} 
The datasets generated and/or analyzed during the current work are available from the corresponding author on reasonable request. 

\subsubsection*{Code availability} 
The source code developed in current work is available from the corresponding author on reasonable request.

\subsubsection*{Supplementary information} 
Supplementary information accompanies this paper and includes 
Supplementary Methods S1--S10 and Supplementary Videos 1--5.

\newpage

\begin{appendices}

\section{Supplementary Methods}

\subsection*{S1. Full System Architecture and Data Flow}

This section provides an implementation-level overview of the world-model-centric autonomous racing agent and its full-scale racing platform. 
The main manuscript defines the system-level formulation under coupled cognitive--physical limits; here we further introduce the full-scale racing platform, notation, online data flow, and closed-loop refinement pipeline used by the implemented system. 
The purpose of this section is to connect the physical platform with the detailed supplementary modules, including world-state construction, world modeling, future-aware reasoning, physical-limit-constrained control, policy refinement, and safety monitoring.

\subsubsection*{S1.1 Full-Scale Racing Platform and Hardware Domains}

\begin{figure}[htb]
    \centering
    \includegraphics[width=0.8\linewidth]{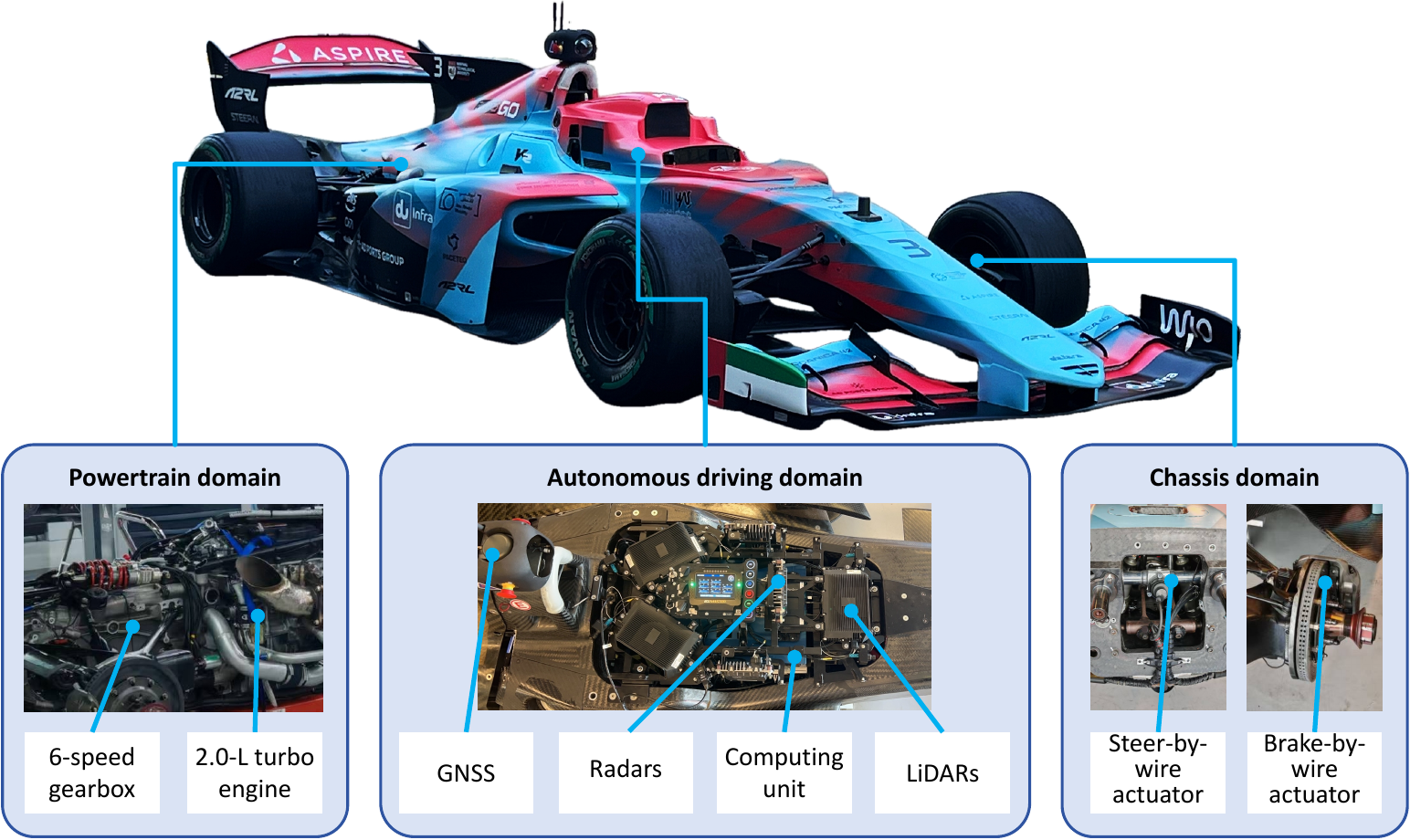}
    \caption{
    Full-scale autonomous racing platform and onboard hardware.
    }
    \label{SF1_vehicle}
\end{figure}

The proposed agent is implemented on a full-scale autonomous racing platform equipped with multi-modal sensing, onboard computation, actuation interfaces, and a real-time autonomy software stack, as shown in Fig.~\ref{SF1_vehicle}. 
The racing vehicle is organized into three hardware domains: the powertrain domain, the autonomous-driving domain, and the chassis domain. 
Together, these domains provide the physical basis for high-speed sensing, onboard computation, and executable actuation under near-limit racing conditions.

The autonomous-driving domain contains core sensing and computing units, including GNSS, 4D millimeter-wave radars, LiDARs, and an onboard computing unit. 
These components support ego-state estimation, opponent perception, world-state construction, and onboard decision-making and control computation. 
Ego-state estimation uses measurements from LiDAR, RTK GNSS, IMU, wheel-speed sensors.
Opponent perception uses LiDAR and 4D millimeter-wave radar to detect and track surrounding vehicles under high-speed and close-proximity racing conditions.
The powertrain domain supplies driving power through a 6-speed gearbox and a 2.0-L turbocharged racing engine. 
The chassis domain provides steer-by-wire and brake-by-wire actuation interfaces that realize the steering and braking commands on the vehicle.
This hardware organization connects the physical racing platform with the world-model-centric autonomy stack.

\subsubsection*{S1.2 World-State Representation and Notation}

At each time step $t$, the agent receives multi-modal sensory observations, the ego vehicle state estimate, and local track information. We denote the surrounding observations by $o_t$, the ego vehicle state by $x_t$, and the local track context by $\Gamma_t$. These inputs are integrated into a structured world-state representation $w_t$.

The ego vehicle state contains the pose, velocity, heading, and relevant dynamic variables required by prediction, planning, and control. The surrounding observation $o_t$ contains object-level opponent states obtained from LiDAR--radar perception and tracking. The local track context $\Gamma_t$ contains the track-coordinate representation, road boundaries, curvature, and local reference information. The resulting world state $w_t$ provides a common representation shared by the world model, decision policy, multimodal planner, and control module.
For consistency, we use the following notation throughout the Supplementary Methods:
\[
\begin{aligned}
x_t &: \text{ego vehicle state}, \\
o_t &: \text{surrounding object observations}, \\
\Gamma_t &: \text{local track context}, \\
w_t &: \text{structured world state}, \\
\mathcal{E}_t &: \text{world-model-informed feasible-motion envelope}, \\
\tau_t^k &: \text{the } k\text{-th candidate trajectory}, \\
\tau_t^\ast &: \text{selected trajectory}, \\
u_t &: \text{control input}.
\end{aligned}
\]
The feasible-motion envelope $\mathcal{E}_t$ represents the current physical-limit boundary of the ego vehicle and is used both in trajectory regularization and in downstream constrained control. Unless otherwise specified, the term ``agent'' refers to the complete autonomous racing system, while ``ego vehicle'' refers to the physical racing platform.

\subsubsection*{S1.3 Online Perception--Prediction--Decision--Control Loop}

The online system follows a perception--prediction--decision--control loop. First, high-frequency localization and opponent perception construct the current world state. The ego state is estimated from LiDAR, RTK GNSS, IMU, wheel-speed, and other onboard measurements, while opponent states are obtained by LiDAR--radar object detection and multi-object tracking. Details of the world-state construction process are provided in Supplementary Section~S2.

Given the current world state, the world model predicts the future evolution of ego dynamics, opponent interaction, and physical-limit availability over a short planning horizon. These predictions provide the basis for evaluating how different candidate maneuvers may evolve under dynamic multi-agent racing conditions. The world model consists of ego dynamics prediction, interaction prediction, and physical-limit representation, as described in Supplementary Section~S4.

The decision policy and multimodal planner then generate a set of candidate trajectories. These candidates correspond to different maneuver hypotheses, such as maintaining the current line, braking and following, or performing an overtaking maneuver. Candidate trajectories are regularized by the feasible-motion envelope to ensure compatibility with the physical capability of the ego vehicle. The world model evaluates the predicted future associated with each candidate trajectory according to interaction safety, physical feasibility, and speed efficiency. The candidate with the highest overall suitability is selected for execution. 
During onboard execution, the future-aware reasoning module operates at 10 Hz evaluating 36 candidate trajectories.
Details of candidate generation and future-aware scoring are provided in Supplementary Section~S5.

Finally, the motion-control module converts the selected trajectory into executable control commands. The controller tracks the selected trajectory while enforcing vehicle dynamics, road-boundary constraints, actuator limits, and coupled longitudinal--lateral physical-limit constraints. The control-oriented dynamics, Jacobians, barrier terms, and constrained CiLQR implementation are provided in Supplementary Section~S6.

The online loop can therefore be summarized as
\[
    (x_t,o_t,\Gamma_t)
    \rightarrow
    w_t
    \rightarrow
    \left\{\hat{w}_{t:t+H}^{\,k}\right\}_{k=1}^{K}
    \rightarrow
    \tau_t^\ast
    \rightarrow
    u_t^\ast .
\]
This flow links current-state construction, world-model prediction, future-aware decision-making, and physical-limit-constrained execution within a unified closed-loop racing agent.

\subsubsection*{S1.4 Closed-Loop Refinement Pipeline}

Beyond online execution, the system is refined through a closed-loop pipeline that links real racing data, world-model learning, internal simulation, and policy update. Real-vehicle deployment produces data containing ego dynamics, opponent interactions, physical-limit behavior, and track context under real operating conditions. These data provide the basis for improving the agent after each deployment phase.

The world model is trained and refined using real racing data. The ego dynamics component learns the vehicle response under high-speed and near-limit operation, the interaction component learns the temporal evolution of multi-agent racing scenarios. The refined world model is then used to generate additional rollout data covering interaction variations, rare competitive scenarios, and near-limit operating conditions that are difficult to collect exhaustively in real racing.

The real and world-model-generated data are combined to update the decision policy, multimodal planner, and physical-limit constraints. The decision policy is updated using conservative reinforcement learning with behavior-cloning regularization, while the planner is updated through best-of-$K$ trajectory supervision. The physical-limit representation is refined using validated near-limit data so that the agent can progressively improve its utilization of the available performance envelope. Details of this refinement process are provided in Supplementary Section~S7.

The refinement pipeline can be summarized as
\[
    \mathcal{D}_{\mathrm{real}}
    \rightarrow
    W
    \rightarrow
    \mathcal{D}_{\mathrm{wm}}
    \rightarrow
    \mathcal{D}
    \rightarrow
    (\pi,\phi,\mathcal{E})
    \rightarrow
    \text{next deployment}.
\]
This closed-loop process improves the fidelity of world-model prediction, expands the coverage of interaction and near-limit scenarios, and progressively refines the agent's decision-making and physical-limit utilization.

\subsection*{S2. High-Frequency World-State Construction}

This section describes how the current racing world state $w_t$ is constructed from high-frequency ego-state estimation, multi-modal sensing, and opponent perception. The resulting world state provides the observable foundation for world-model prediction, future-aware reasoning, and physical-limit-constrained control.

\subsubsection*{S2.1 Ego Vehicle State and Motion Model}

\begin{figure}[!htb]
    \centering
    \includegraphics[width=0.6\linewidth]{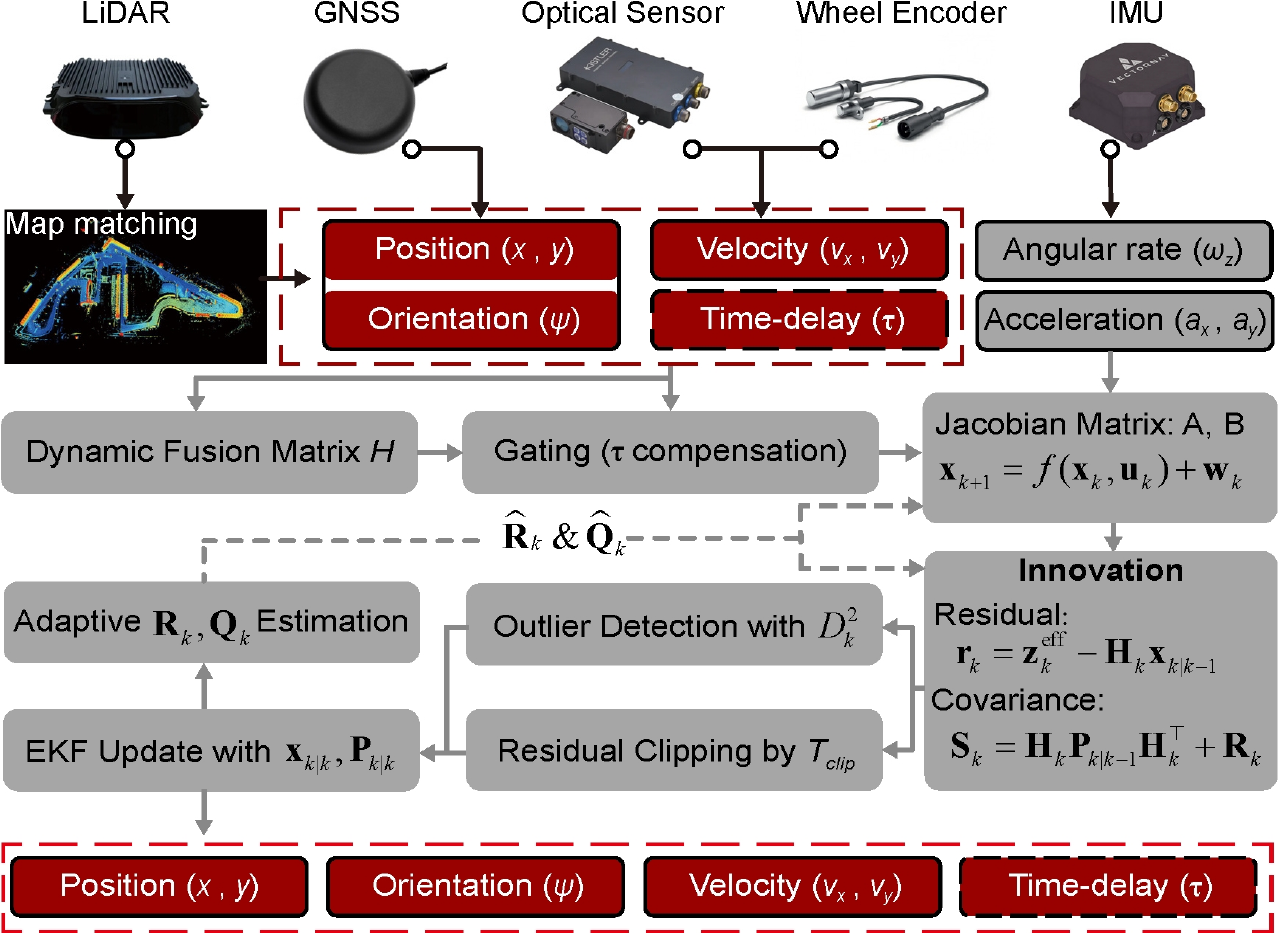}
    \caption{
    Multi-sensor ego-state estimation framework for high-frequency world-state construction.
    \label{SF2_se_framework}
    }
\end{figure}

The overall multi-sensor ego-state estimation framework is shown in Fig.~\ref{SF2_se_framework}, where LiDAR map matching, RTK GNSS, optical sensing, wheel encoders, and IMU measurements are fused with delay compensation, adaptive covariance estimation, innovation gating, and EKF update to estimate position, orientation, velocity, and sensor time delay.

The ego-state estimation module maintains a compact state of the ego vehicle in the global track frame,
\begin{equation}
    \mathbf{x}
    =
    \big(
    \mathbf{R},\,
    \mathbf{p},\,
    \mathbf{v},\,
    \mathbf{b}_g,\,
    \mathbf{b}_a,\,
    \mathbf{g}
    \big),
    \label{eq:supp_state_def}
\end{equation}
where $\mathbf{R}\in SO(3)$ is the body-to-track orientation, $\mathbf{p},\mathbf{v}\in\mathbb{R}^3$ are the position and velocity in the track frame, $\mathbf{b}_g,\mathbf{b}_a\in\mathbb{R}^3$ are gyroscope and accelerometer biases, and $\mathbf{g}\in\mathbb{R}^3$ is the gravity vector.

The IMU measures angular velocity and specific force in the body frame,
\begin{equation}
\begin{aligned}
    \boldsymbol{\omega}_m
    &=
    \boldsymbol{\omega}
    +
    \mathbf{b}_g
    +
    \mathbf{w}_g,
    \\
    \mathbf{a}_m
    &=
    \mathbf{a}
    +
    \mathbf{b}_a
    +
    \mathbf{w}_a,
\end{aligned}
\label{eq:supp_imu_model}
\end{equation}
where $\boldsymbol{\omega}$ and $\mathbf{a}$ denote the true angular velocity and specific force, and $\mathbf{w}_g$ and $\mathbf{w}_a$ are zero-mean white noises. The bias terms follow random walks driven by additional Gaussian noise. Gravity is assumed to be constant over the time scale of a racing lap.

Over one IMU sampling period $\Delta t$, the noise-free discrete-time propagation is
\begin{equation}
\begin{aligned}
    \mathbf{R}_{k+1}
    &=
    \mathbf{R}_{k}
    \exp\!\left(
    (\boldsymbol{\omega}_{m,k}-\mathbf{b}_{g,k})^\wedge
    \Delta t
    \right),
    \\
    \mathbf{v}_{k+1}
    &=
    \mathbf{v}_{k}
    +
    \left(
    \mathbf{R}_{k}
    (\mathbf{a}_{m,k}-\mathbf{b}_{a,k})
    +
    \mathbf{g}_{k}
    \right)
    \Delta t,
    \\
    \mathbf{p}_{k+1}
    &=
    \mathbf{p}_{k}
    +
    \mathbf{v}_{k}\Delta t
    +
    \frac{1}{2}
    \left(
    \mathbf{R}_{k}
    (\mathbf{a}_{m,k}-\mathbf{b}_{a,k})
    +
    \mathbf{g}_{k}
    \right)
    \Delta t^2,
    \\
    \mathbf{b}_{g,k+1}
    &=
    \mathbf{b}_{g,k},
    \qquad
    \mathbf{b}_{a,k+1}
    =
    \mathbf{b}_{a,k},
    \qquad
    \mathbf{g}_{k+1}
    =
    \mathbf{g}_{k}.
\end{aligned}
\label{eq:supp_disc_model}
\end{equation}
Here $(\cdot)^\wedge$ denotes the mapping from a rotation vector to a skew-symmetric matrix. Orientation is updated through right multiplication by $\exp(\cdot)$, while linear quantities are propagated in Euclidean space.

\subsubsection*{S2.2 Multi-Sensor Measurement Models}

All exteroceptive measurements are expressed as noisy observations of position, orientation, or velocity in the track frame. A generic measurement $\mathbf{z}_j$ is written as
\begin{equation}
    \mathbf{z}_j
    =
    h_j(\mathbf{x})
    +
    \mathbf{n}_j,
    \qquad
    \mathbf{n}_j
    \sim
    \mathcal{N}(\mathbf{0},\mathbf{R}_j),
    \label{eq:supp_meas_general}
\end{equation}
where $h_j(\cdot)$ selects and transforms the relevant components of $\mathbf{x}$, and $\mathbf{R}_j$ is the measurement covariance.

RTK GNSS and LiDAR map matching provide position observations,
\begin{equation}
    \mathbf{z}_p
    =
    \mathbf{p}
    +
    \mathbf{n}_p.
\end{equation}
Dual-antenna GNSS heading and LiDAR odometry provide orientation observations,
\begin{equation}
    \mathbf{z}_R
    \approx
    \mathbf{R}
    \exp(\mathbf{n}_R^\wedge).
\end{equation}
Wheel-speed encoders and the downward-looking optical-flow sensor provide body-frame velocity observations,
\begin{equation}
    \mathbf{z}_v
    =
    \mathbf{R}^{\top}
    \mathbf{v}
    +
    \mathbf{n}_v,
\end{equation}
which constrain longitudinal and lateral velocity relative to the road surface. The covariance of each modality is identified from calibration runs and short test laps.

\subsubsection*{S2.3 Delayed Measurements and Factorized Covariance}

High-speed autonomous racing requires delayed LiDAR and GNSS measurements to be inserted at their correct physical times. To support this, the estimator maintains a short history of recent motion states. The rapidly varying motion components are defined as
\begin{equation}
    \mathbf{x}^{\mathrm{mot}}
    =
    (
    \mathbf{R},
    \mathbf{p},
    \mathbf{v},
    \mathbf{b}_g
    ),
\end{equation}
while slowly varying parameters $(\mathbf{b}_a,\mathbf{g})$ are maintained outside the history.

At estimator time $k$, the short-horizon history is
\begin{equation}
    \mathcal{X}_k
    =
    \left\{
    \mathbf{x}^{\mathrm{mot}}_{k,1},
    \ldots,
    \mathbf{x}^{\mathrm{mot}}_{k,N}
    \right\},
    \qquad
    t_{k,1} > \cdots > t_{k,N},
    \label{eq:supp_hist_def}
\end{equation}
where $\mathbf{x}^{\mathrm{mot}}_{k,1}$ corresponds to the current time and the remaining nodes represent older states. The history length and capacity satisfy
\begin{equation}
    t_{k,1}-t_{k,N}
    \le
    T_{\mathrm{hist}},
    \qquad
    N
    \le
    N_{\max}.
\end{equation}
At each IMU step, a new state is propagated using Eq.~(\ref{eq:supp_disc_model}) and appended to the front of $\mathcal{X}_k$. The oldest node is removed once the time horizon or capacity is exceeded.

A delayed measurement $\mathbf{z}_j$ time-stamped at $t_j$ is associated with the closest history node,
\begin{equation}
    i^\star
    =
    \arg\min_{i\in\{1,\ldots,N\}}
    |t_{k,i}-t_j|,
    \label{eq:supp_assoc_idx}
\end{equation}
subject to
\begin{equation}
    |t_{k,i^\star}-t_j|
    <
    \tau_{\max},
\end{equation}
where $\tau_{\max}$ is the maximum allowed time offset. If this condition is not satisfied, the measurement is discarded. Multiple delayed measurements available at the same estimator time are collected into an ensemble and updated jointly after linearization around the current prediction.

\begin{figure}[!htb]
    \centering
    \includegraphics[width=1.0\linewidth]{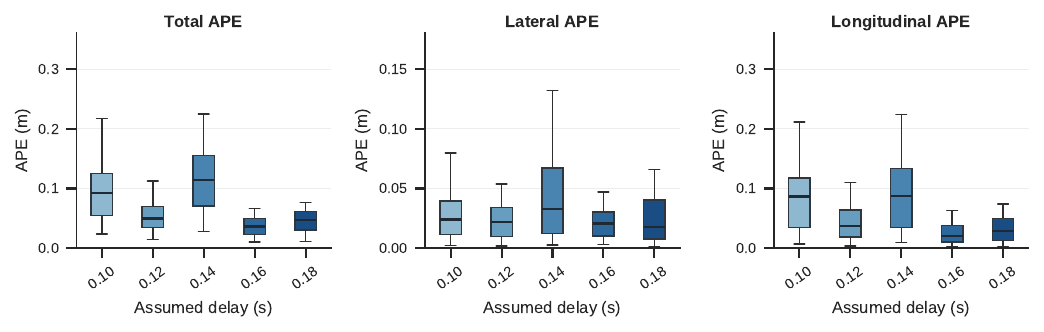}
    \caption{
    Delay sensitivity analysis of localization error.
    }
    \label{SF3_delay_sensitivity}
\end{figure}

The delay sensitivity analysis of localization error is shown in Fig.~\ref{SF3_delay_sensitivity}, where the APE distributions remain stable under different assumed measurement delays.

The covariance of the stack history $\mathbf{X}_k$ is stored in a factorized form,
\begin{equation}
    \mathbf{P}_k
    =
    \mathbf{S}_k^\top
    \mathbf{S}_k,
    \label{eq:supp_factorized_cov}
\end{equation}
where $\mathbf{S}_k$ is an upper-triangular square-root factor. Prediction and update operations are performed on $\mathbf{S}_k$ rather than on $\mathbf{P}_k$ directly, avoiding explicit matrix inversion and improving numerical conditioning as nodes are inserted and removed.

When process noise is injected or linear propagation is applied, covariance addition is implemented by stacking square-root factors and applying QR decomposition. Conceptually, if $\mathbf{S}_A$ and $\mathbf{S}_B$ are square-root factors of two covariance terms, a factor of their sum is obtained from the upper-triangular part of the QR factorization of
\begin{equation}
    \left[
    \mathbf{S}_A^\top
    \quad
    \mathbf{S}_B^\top
    \right]^\top .
\end{equation}
This operation is used when advancing the history with the IMU model and when compressing information from older nodes.

For measurement updates, the linearized ensemble model is
\begin{equation}
    \mathbf{Z}
    \approx
    \mathbf{H}\delta\mathbf{X}
    +
    \mathbf{N},
\end{equation}
where $\delta\mathbf{X}$ is the stacked perturbation of all history nodes, $\mathbf{H}$ is a block-sparse Jacobian, and $\mathbf{N}$ has block-diagonal covariance
\begin{equation}
    \mathbf{R}_{\mathrm{ens}}
    =
    \operatorname{diag}
    (
    \mathbf{R}_1,
    \ldots,
    \mathbf{R}_M
    ).
\end{equation}
The residuals and Jacobian are pre-whitened using a square-root factor of $\mathbf{R}_{\mathrm{ens}}$, stacked with $\mathbf{S}_k$, and factorized by QR decomposition to obtain the updated factor $\mathbf{S}_k^+$ and the corresponding state correction.

\subsubsection*{S2.4 Robust Measurement Update}

Occasional outliers can arise from GNSS multipath, partial LiDAR occlusion, sparse radar returns, or short-lived modeling errors. To reduce their effect, measurement innovations are weighted using a Huber-type robust loss. Let $\mathbf{r}_j$ denote the residual of measurement $j$ with nominal covariance $\mathbf{R}_j$. Its Mahalanobis norm is
\begin{equation}
    \|\mathbf{r}_j\|_{\mathbf{R}_j^{-1}}
    =
    \sqrt{
    \mathbf{r}_j^\top
    \mathbf{R}_j^{-1}
    \mathbf{r}_j
    }.
\end{equation}
The Huber cost is
\begin{equation}
    \rho_H(\mathbf{r}_j)
    =
    \begin{cases}
    \frac{1}{2}
    \mathbf{r}_j^\top
    \mathbf{R}_j^{-1}
    \mathbf{r}_j,
    &
    \|\mathbf{r}_j\|_{\mathbf{R}_j^{-1}}
    \le
    \delta,
    \\[3pt]
    \delta
    \|\mathbf{r}_j\|_{\mathbf{R}_j^{-1}}
    -
    \frac{1}{2}\delta^2,
    &
    \|\mathbf{r}_j\|_{\mathbf{R}_j^{-1}}
    >
    \delta,
    \end{cases}
    \label{eq:supp_huber_cost}
\end{equation}
where $\delta>0$ is the Huber threshold. In an iteratively reweighted form, this gives the scalar weight
\begin{equation}
    w(\mathbf{r}_j)
    =
    \begin{cases}
    1,
    &
    \|\mathbf{r}_j\|_{\mathbf{R}_j^{-1}}
    \le
    \delta,
    \\[4pt]
    \displaystyle
    \frac{\delta}
    {\|\mathbf{r}_j\|_{\mathbf{R}_j^{-1}}},
    &
    \|\mathbf{r}_j\|_{\mathbf{R}_j^{-1}}
    >
    \delta.
    \end{cases}
    \label{eq:supp_huber_weight}
\end{equation}
The ensemble covariance is then adjusted as
\begin{equation}
    \mathbf{R}_{\mathrm{ens}}
    =
    \operatorname{diag}
    \left(
    w(\mathbf{r}_1)\mathbf{R}_1,
    \ldots,
    w(\mathbf{r}_M)\mathbf{R}_M
    \right).
\end{equation}
In practice, a single reweighting step per update is sufficient to suppress outliers while preserving sensitivity to consistent trends in the measurements.

\begin{figure}[!htb]
    \centering
    \includegraphics[width=1.0\linewidth]{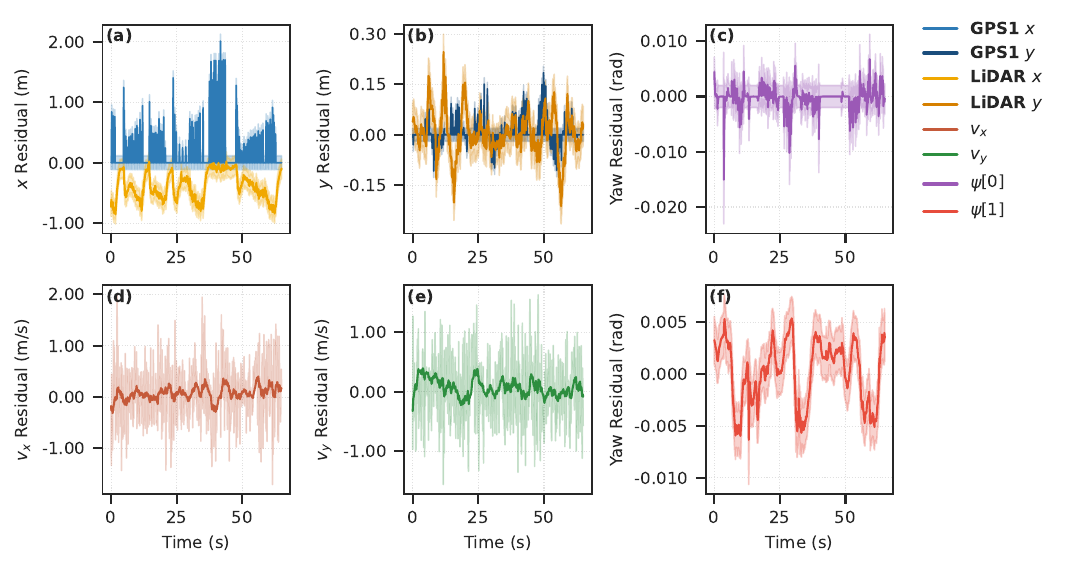}
    \caption{Residual profiles of multi-sensor ego-state localization.}
    \label{SF4_residual_profile}
\end{figure}
The resulting residual profiles for state estimation including the position, velocity, and yaw residuals over a full racing lap are shown in Fig.~\ref{SF4_residual_profile}, indicating bounded and temporally stable innovations across GNSS, LiDAR, velocity, and yaw observations.

\subsubsection*{S2.5 Opponent Perception and Multi-Object Tracking}

Opponent perception estimates the position, velocity, heading, and spatial extent of surrounding vehicles under high-speed and close-proximity racing conditions. LiDAR and 4D millimeter-wave radar are used as the primary exteroceptive sensors, and fusion is performed at the object-tracking level.

\begin{figure}[ht]
    \centering
    \includegraphics[width=0.8\linewidth]{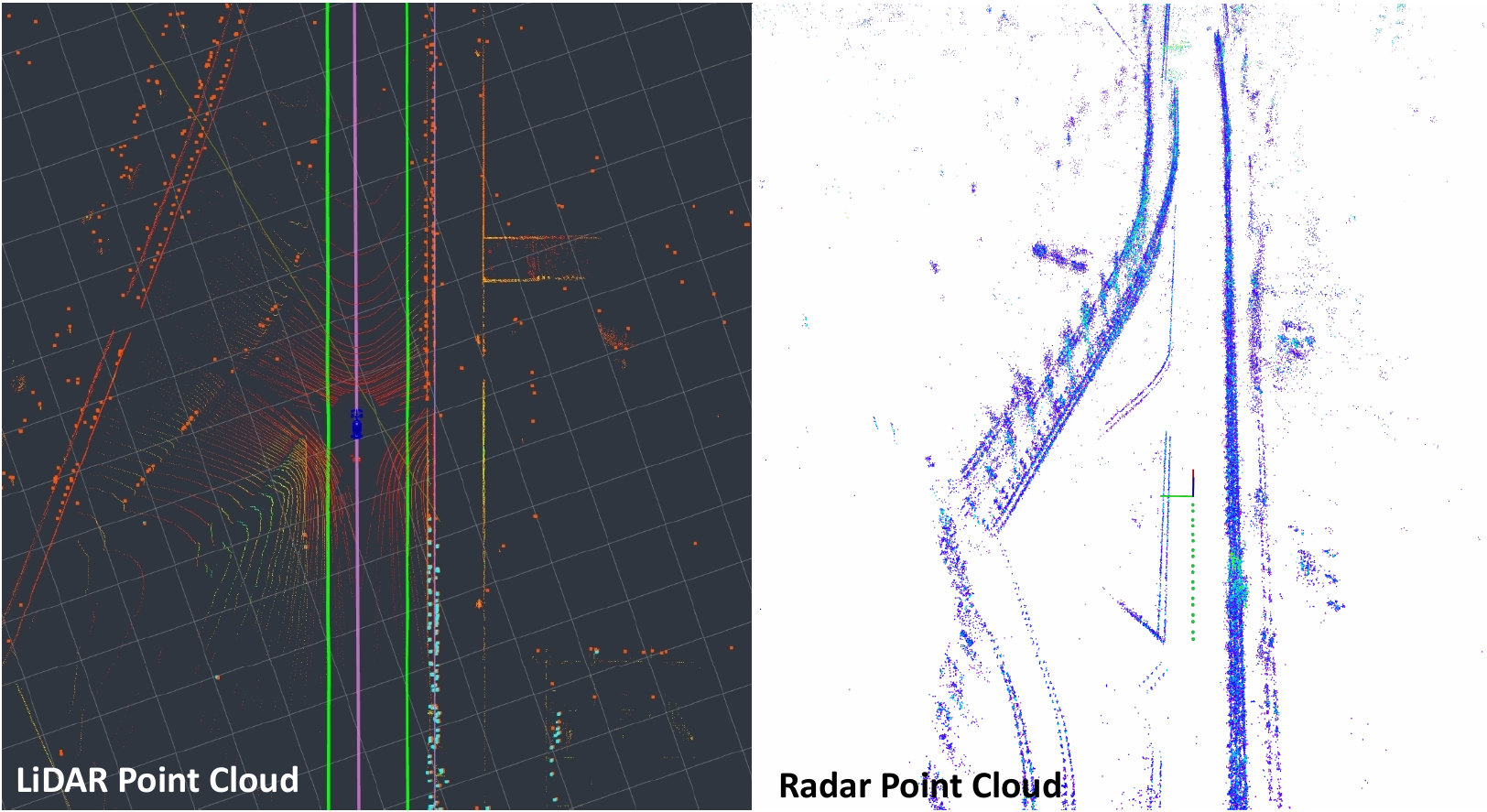}
    \caption{
    LiDAR and radar point clouds during driving.
    }
    \label{SF5_point_cloud_observations}
\end{figure}

The point cloud observations from LiDAR and radar are shown in Fig.~\ref{SF5_point_cloud_observations}.

\begin{figure}[ht]
    \centering
    \includegraphics[width=\linewidth]{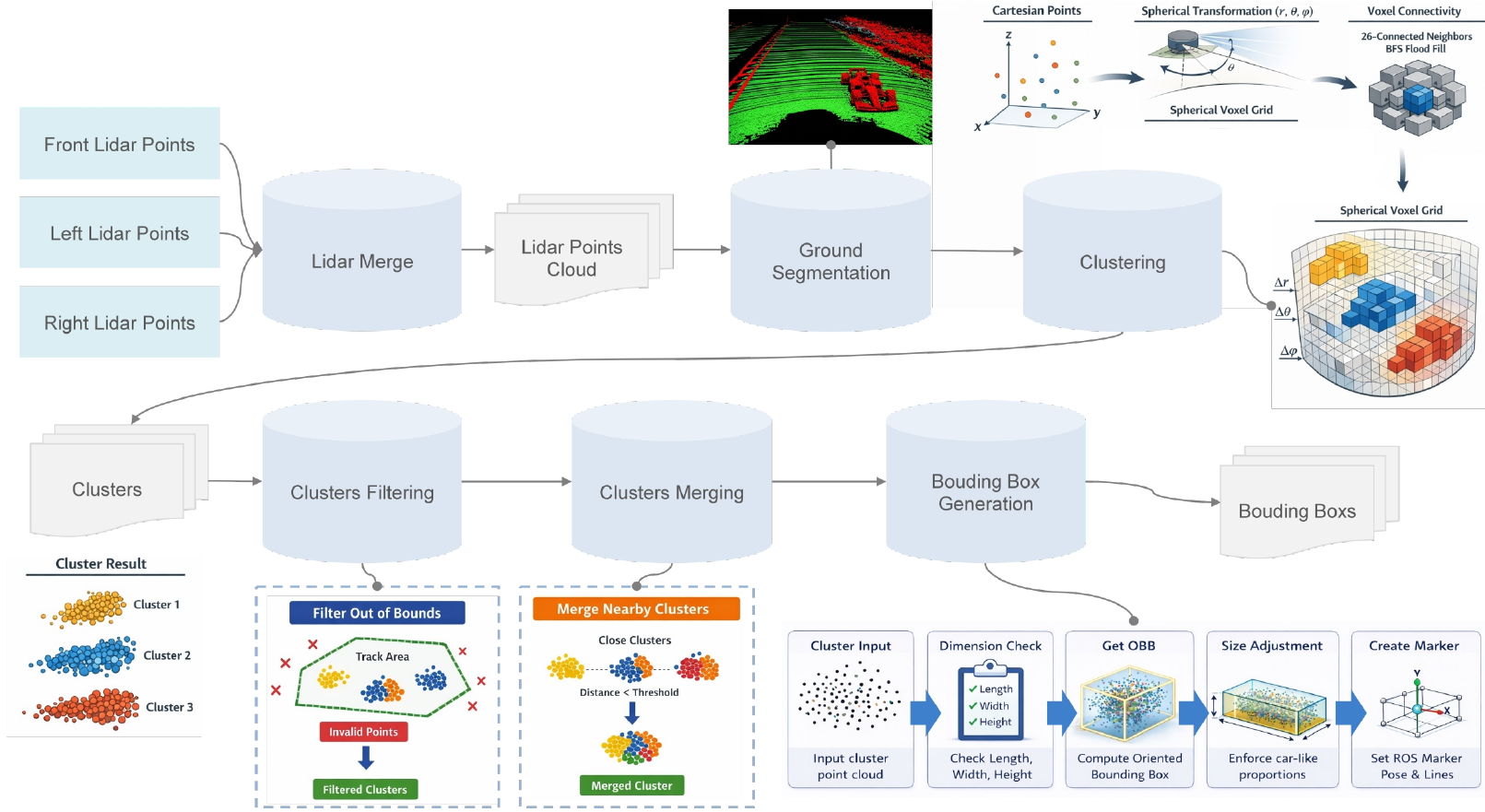}
    \caption{
    Point cloud processing pipeline for opponent perception.
    }
    \label{SF6_perception_pipline}
\end{figure}

The opponent perception pipeline is illustrated in Fig.~\ref{SF6_perception_pipline}.

Raw LiDAR and radar point clouds are first transformed into a common vehicle-centric coordinate frame. Ground points are removed using an elevation-consistency model defined over radial bins under bounded slope assumptions. The maximum allowable height increase between consecutive radial rings is
\begin{equation}
    \Delta z_{\max}
    =
    \min
    \left(
    \Delta r
    \tan(\mathrm{slope}_{\max}),
    h_{\mathrm{th}}-\varepsilon
    \right),
\end{equation}
where $\Delta r$ is the radial increment, $\mathrm{slope}_{\max}$ is the maximum admissible ground slope, $h_{\mathrm{th}}$ is the height threshold, and $\varepsilon$ is a safety margin. Local plane refinement is further applied in the near field.

After ground removal, non-ground points are discretized in spherical voxel space,
\begin{equation}
    (r,\mathrm{az},\mathrm{el})
    =
    \left(
    \sqrt{x^2+y^2+z^2},
    \operatorname{atan2}(y,x),
    \arctan
    \left(
    \frac{z}{\sqrt{x^2+y^2}}
    \right)
    \right),
\end{equation}
where $r$, $\mathrm{az}$, and $\mathrm{el}$ denote range, azimuth, and elevation. Connected components are extracted by neighborhood traversal in the voxel grid, and small clusters are removed. Spatially adjacent clusters are merged according to proximity and admissible vehicle dimensions. Hypotheses that violate geometric consistency are rejected.

Each retained cluster is represented by an oriented bounding box. Given a point set $\mathcal{P}=\{p_i\}_{i=1}^{N}$, the principal axes are obtained from the covariance matrix $C$ by eigen decomposition,
\begin{equation}
    C v_j
    =
    \lambda_j v_j,
    \qquad
    j=1,2,3,
\end{equation}
where $\lambda_j$ and $v_j$ denote eigenvalues and eigenvectors. The rotation matrix is
\begin{equation}
    R_{\mathrm{obb}}
    =
    [
    v_1
    \;\;
    v_2
    \;\;
    v_3
    ],
\end{equation}
and axis-aligned extents are computed in the principal frame and mapped back to obtain the oriented bounding-box pose and dimensions. Radar point clouds follow the same pipeline with modality-specific clustering thresholds to account for sparse and anisotropic returns.

The extracted bounding boxes are treated as object-level measurements and passed to an EKF-based multi-object tracking module. Each tracked opponent is represented by
\begin{equation}
    \mathbf{x}_o
    =
    (
    \mathbf{p}_o,
    \mathbf{v}_o,
    \psi_o
    ),
\end{equation}
where $\mathbf{p}_o$ and $\mathbf{v}_o$ denote object position and velocity in the vehicle-centric frame, and $\psi_o$ denotes object heading. Object hypotheses from LiDAR and radar are associated using spatial gating and track-consistency checks. Measurement updates are applied asynchronously as observations arrive, with modality-specific covariance settings retained in each measurement model. This tracking-level fusion allows LiDAR geometry and radar velocity cues to jointly support temporally coherent opponent states.

\subsubsection*{S2.6 Construction of the Racing World State}

The racing world state used by the world model and downstream decision-making is assembled from ego-state estimation, opponent tracking, and local track information. At time $t$, the structured world state is written as
\begin{equation}
    w_t
    =
    \Psi_w(x_t,o_t,\Gamma_t),
\end{equation}
where $x_t$ denotes the ego vehicle state, $o_t$ denotes the set of tracked opponent states, and $\Gamma_t$ denotes local track context.

For the ego vehicle, the state includes the pose, velocity, heading, yaw rate, and other dynamic quantities required by ego-dynamics prediction and motion control. For each opponent, the tracked object state includes position, velocity, heading, and bounding-box information. The local track context includes the track-coordinate representation, local curvature, reference line, and left--right road boundaries.

This world-state representation provides a common interface for the three downstream components. The world model uses $w_t$ to predict ego dynamics, interaction evolution, and physical feasibility. Future-aware reasoning uses $w_t$ and predicted futures to generate and evaluate candidate trajectories. Physical-limit-constrained control uses the selected trajectory together with the feasible-motion envelope $\mathcal{E}_t$ to compute executable control commands. In this way, high-frequency localization and opponent perception serve as the observable foundation for predictive world understanding under cognitive-physical limits.

\subsection*{S3. Quantification of Cognitive and Physical Limits}

The main manuscript evaluates cognitive and physical limits through observable indicators rather than claiming absolute theoretical upper bounds. This section provides the detailed definitions used to quantify these indicators. The cognitive side is evaluated along the perception--prediction--decision chain, while the physical side is evaluated through acceleration-envelope utilization, tire-force utilization, tracking capability, and maneuvering stability. These indicators are interpreted jointly to assess how the agent approaches and regulates coupled cognitive-physical limits during autonomous racing.

\subsubsection*{S3.1 Cognitive-Limit Indicators}

The cognitive limit refers to the boundary of the agent's ability to construct, update, predict, and use task-relevant world understanding under severe time pressure, uncertainty, and dynamic interaction. For autonomous racing, this includes high-frequency localization and perception under high-speed motion, prediction of ego--opponent interaction evolution, and decision-making for safe, feasible, and competitive maneuvers.

The cognitive limit is not determined by a single error source. Instead, uncertainty propagates along the perception--prediction--decision chain. Localization and ego-state residuals introduce uncertainty into the reconstructed racing world state. This uncertainty propagates into ego-dynamics prediction and interaction prediction, where it affects future trajectory rollouts. Prediction errors then influence trajectory ranking and action selection, and eventually appear as closed-loop outcomes such as unsafe overtaking, collision, or overly conservative behavior. The cognitive limit is approached when accumulated uncertainty prevents the agent from reliably distinguishing actions that are safe, feasible, and competitive.

Current-state construction is quantified by localization and ego-state residuals,
\begin{equation}
\mathbf{e}^{\mathrm{state}}_t
=
\left[
|e_{x,t}|,\,
|e_{y,t}|,\,
|e_{v_x,t}|,\,
|e_{v_y,t}|,\,
|e_{\psi,t}|
\right],
\label{eq:supp_cog_state_residual}
\end{equation}
where $e_{x,t}$ and $e_{y,t}$ denote longitudinal and lateral position residuals, $e_{v_x,t}$ and $e_{v_y,t}$ denote longitudinal and lateral velocity residuals, and $e_{\psi,t}$ denotes the heading residual. Opponent perception is assessed by the temporal consistency of tracked opponent states, including whether object identities, positions, velocities, and bounding boxes remain coherent under high-speed and close-proximity racing conditions.

Ego-dynamics prediction is evaluated by the prediction error between the predicted ego state and the corresponding measured or estimated state,
\begin{equation}
\mathbf{e}^{\mathrm{ego}}_{t+i}
=
\hat{x}_{t+i}
-
x_{t+i},
\qquad
i=1,\ldots,T.
\label{eq:supp_ego_prediction_error}
\end{equation}
For each ego-state channel, the root-mean-square error is computed as
\begin{equation}
\mathrm{RMSE}_{x_j}
=
\sqrt{
\frac{1}{N T}
\sum_{n=1}^{N}
\sum_{i=1}^{T}
\left(
\hat{x}^{(n)}_{j,t+i}
-
x^{(n)}_{j,t+i}
\right)^2
},
\label{eq:supp_ego_rmse}
\end{equation}
where $x_j$ denotes the $j$-th ego-state variable, $N$ is the number of evaluated sequences, and $T$ is the prediction horizon.

Future-interaction prediction is evaluated using average displacement error and final displacement error. For an opponent trajectory predicted over a horizon $T$,
\begin{equation}
\mathrm{ADE}
=
\frac{1}{T}
\sum_{i=1}^{T}
\left\|
\hat{o}_{t+i}
-
o_{t+i}
\right\|_2,
\label{eq:supp_ade}
\end{equation}
\begin{equation}
\mathrm{FDE}
=
\left\|
\hat{o}_{t+T}
-
o_{t+T}
\right\|_2,
\label{eq:supp_fde}
\end{equation}
where $\hat{o}_{t+i}$ and $o_{t+i}$ denote the predicted and ground-truth opponent states at prediction step $i$. In racing scenarios, these states are evaluated in a structured track-coordinate representation so that longitudinal progress, lateral offset, and interaction geometry are preserved.

Closed-loop decision capability is quantified by interaction success rate, collision rate, safe mileage, and overtaking frequency. The interaction success rate is
\begin{equation}
R_{\mathrm{succ}}
=
\frac{N_{\mathrm{succ}}}
{N_{\mathrm{total}}},
\label{eq:supp_success_rate}
\end{equation}
where $N_{\mathrm{succ}}$ denotes the number of successful interactions and $N_{\mathrm{total}}$ denotes the total number of interaction trials. The collision rate is
\begin{equation}
R_{\mathrm{col}}
=
\frac{N_{\mathrm{col}}}
{N_{\mathrm{total}}},
\label{eq:supp_collision_rate}
\end{equation}
where $N_{\mathrm{col}}$ denotes the number of interactions that lead to collision. The overtaking frequency is computed as
\begin{equation}
F_{\mathrm{over}}
=
\frac{N_{\mathrm{over}}}
{N_{\mathrm{lap}}},
\label{eq:supp_overtaking_frequency}
\end{equation}
where $N_{\mathrm{over}}$ is the number of completed overtaking maneuvers and $N_{\mathrm{lap}}$ is the number of evaluated laps. Safe mileage is the accumulated distance travelled without collision, boundary violation, or emergency termination.

These cognitive indicators are complementary. State residuals measure the reliability of current world-state construction; ego-dynamics and interaction-prediction errors measure the quality of future prediction; and closed-loop outcomes measure whether the agent can convert state estimation and prediction into safe, feasible, and competitive behavior.

\subsubsection*{S3.2 Physical-Limit Indicators}

The physical limit refers to the boundary of feasible motion imposed by vehicle dynamics, actuator constraints, tire--road interaction, aerodynamic loading, road friction, tire temperature, and stability constraints. For autonomous racing, this boundary is reflected not only by vehicle-level acceleration capability, but also by tire utilization, controller tracking capability, and maneuvering stability. Therefore, the physical limit is quantified using a set of layered indicators.

At the vehicle level, physical-limit utilization is evaluated by the executed acceleration state relative to the feasible-motion envelope $\mathcal{E}_t$. The coupled acceleration envelope is defined in the main manuscript as a speed- and friction-dependent boundary in the $(a_x,a_y)$ plane. We define the normalized envelope utilization as
\begin{equation}
\eta_{\mathrm{env},t}
=
g_{\mathcal{E}}
\left(
a_{x,t},
a_{y,t},
v_{x,t},
\mu_t;
\mathcal{E}_t
\right),
\label{eq:supp_eta_env}
\end{equation}
where $g_{\mathcal{E}}(\cdot)$ is induced by the coupled acceleration-envelope function. For the super-ellipse envelope used in this work, the utilization can be written as
\begin{equation}
\eta_{\mathrm{env},t}
=
\begin{cases}
\left(
\dfrac{a_{x,t}}
{a^{\mathrm{acc}}_{x,\max}(v_{x,t},\mu_t)}
\right)^c
+
\left(
\dfrac{|a_{y,t}|}
{a_{y,\max}(v_{x,t},\mu_t)}
\right)^c,
& a_{x,t}\ge 0,
\\[10pt]
\left(
\dfrac{|a_{x,t}|}
{a^{\mathrm{dec}}_{x,\max}(v_{x,t},\mu_t)}
\right)^c
+
\left(
\dfrac{|a_{y,t}|}
{a_{y,\max}(v_{x,t},\mu_t)}
\right)^c,
& a_{x,t}<0.
\end{cases}
\label{eq:supp_eta_env_explicit}
\end{equation}
A value $\eta_{\mathrm{env},t}\leq 1$ indicates that the executed acceleration remains inside the feasible envelope, while values close to one indicate operation near the vehicle-level physical boundary.

At the tire level, slip ratio and slip angle are used to quantify tire--road force utilization. The normalized tire-limit utilization is
\begin{equation}
\eta_{\mathrm{tire},t}
=
\min
\left(
\frac{1}{4}
\sum_{i\in\{\mathrm{fl,fr,rl,rr}\}}
\max
\left(
\frac{|s_{x,i,t}|}{s^{\mathrm{th}}_x},
\frac{|s_{y,i,t}|}{s^{\mathrm{th}}_y}
\right),
2
\right),
\label{eq:supp_tire_limit_utilization}
\end{equation}
where $s_{x,i,t}$ denotes the slip ratio of tire $i$, $s_{y,i,t}$ denotes the slip angle of tire $i$, and $s^{\mathrm{th}}_x$ and $s^{\mathrm{th}}_y$ are the corresponding limit thresholds. The value is clipped at $2$ instead of $1$ to retain information about the severity of limit violation for soft-constraint penalties.

At the control level, tracking capability is quantified by lateral and speed tracking errors,
\begin{equation}
\eta_{\mathrm{control},t}
=
\min
\left(
\max
\left(
\frac{|e^{\mathrm{trk}}_{y,t}|}
{e^{\mathrm{th}}_y},
\frac{|e^{\mathrm{trk}}_{v,t}|}
{e^{\mathrm{th}}_v}
\right),
2
\right),
\label{eq:supp_controller_limit_utilization}
\end{equation}
where $e^{\mathrm{trk}}_{y,t}$ denotes the lateral tracking error, $e^{\mathrm{trk}}_{v,t}$ denotes the speed tracking error, and $e^{\mathrm{th}}_y$ and $e^{\mathrm{th}}_v$ are the corresponding thresholds.

At the stability level, the discrepancy between actual and desired yaw rate is used to detect the onset of maneuvering instability. The desired yaw rate is approximated by
\begin{equation}
r^{\mathrm{des}}_t
=
\frac{v_{x,t}}{L}
\tan(\delta_{\mathrm{fb},t})
\cdot
\frac{1}{1+K v_{x,t}^2},
\label{eq:supp_desired_yaw_rate}
\end{equation}
where $L$ is the wheelbase, $\delta_{\mathrm{fb},t}$ is the steering feedback angle, and $K$ is the speed-dependent yaw-rate gain coefficient. The stability utilization is
\begin{equation}
\eta_{\mathrm{stability},t}
=
\min
\left(
\frac{|r_t-r^{\mathrm{des}}_t|}
{r^{\mathrm{th}}},
2
\right),
\label{eq:supp_stability_limit_utilization}
\end{equation}
where $r_t$ is the actual yaw rate and $r^{\mathrm{th}}$ is the stability threshold.

The hybrid physical-limit utilization combines the vehicle-level, tire-level, control-level, and stability-level indicators:
\begin{equation}
\eta_{\mathrm{lim},t}
=
\max
\left(
\eta_{\mathrm{tire},t},
\eta_{\mathrm{control},t},
\eta_{\mathrm{stability},t}
\right).
\label{eq:supp_hybrid_limit_utilization}
\end{equation}
This value is used to assess whether the ego vehicle remains within the prescribed physical-limit criterion during envelope exploration, planning, and control. An expanded physical-limit envelope is accepted only if the maximum utilization over the evaluated rollout remains below the prescribed threshold.
The metric $\eta_{\mathrm{env},t}$ serves as an envelope-based representation of $\eta_{\mathrm{lim},t}$ to quantify the proximity of the current state to the feasible motion boundary.

\subsubsection*{S3.3 Coupled Interpretation of Cognitive and Physical Indicators}

The cognitive and physical indicators are interpreted jointly because the two limits are coupled in closed-loop racing. A candidate maneuver may be strategically favorable according to interaction reasoning, but it is useful only if it remains executable within the current feasible-motion envelope. Conversely, a maneuver may be physically feasible in isolation but unsafe under localization uncertainty, opponent-tracking uncertainty, or interaction-prediction error.

In this work, localization residuals, state residuals, ego-dynamics prediction errors, and interaction-prediction ADE/FDE quantify cognitive uncertainty. Acceleration-envelope utilization, tire slip utilization, tracking errors, and yaw-rate discrepancy quantify physical-limit utilization. Closed-loop outcomes such as success rate, collision rate, safe mileage, and overtaking frequency reflect the consequence of their coupling during execution.

For a trajectory candidate $\tau_t^k$, the closed-loop interpretation can be summarized as
\begin{equation}
\tau_t^k
\;\text{is effective}
\quad
\Longleftrightarrow
\quad
\begin{cases}
\text{interaction prediction indicates acceptable risk},\\
\text{decision score indicates competitive racing utility},\\
\eta_{\mathrm{lim},t:t+H}^{k} \leq \eta_{\mathrm{th}},\\
\text{execution produces no collision or instability}.
\end{cases}
\label{eq:supp_effective_maneuver}
\end{equation}
where $\eta_{\mathrm{lim},t:t+H}^{k}$ denotes the maximum physical-limit utilization along the predicted or executed horizon of candidate $\tau_t^k$, and $\eta_{\mathrm{th}}$ is the prescribed physical-limit threshold.

This joint interpretation connects the supplementary metrics to the main results. The localization and prediction metrics support the assessment of current and future world understanding; the physical-limit metrics support the assessment of dynamic-envelope utilization and stability; and the interaction outcomes evaluate whether the agent can select and execute maneuvers that remain safe, feasible, and competitive under coupled cognitive-physical limits.

\subsection*{S4. Predictive World Modeling}

The world model provides the predictive representation used by future-aware reasoning, physical-limit-constrained control, and closed-loop refinement. Starting from the structured racing world state $w_t$, it predicts three coupled components: the future ego vehicle response, the evolution of ego--opponent interaction, and the feasible-motion boundary of the ego vehicle. This section provides the implementation details of these three components.

\subsubsection*{S4.1 Ego Dynamics Prediction}

The ego dynamics model predicts the short-horizon response of the ego vehicle under aggressive control inputs. It adopts a physics-constrained residual-learning structure, combining a differentiable vehicle-dynamics backbone with a learnable residual model:
\begin{equation}
\hat{x}_{t+1}
=
f_{\mathrm{dyn}}
\left(
x_t,u_t;\theta_{\mathrm{dyn}}
\right)
+
f_{\mathrm{res}}
\left(
x_{t-h:t},u_{t-h:t};\theta_{\mathrm{res}}
\right),
\label{eq:supp_wm_dyn}
\end{equation}
where $x_t$ is the ego vehicle state, $u_t$ is the control input, $f_{\mathrm{dyn}}(\cdot)$ is the differentiable dynamics backbone, and $f_{\mathrm{res}}(\cdot)$ is the residual model using recent state and control histories.

The differentiable backbone preserves the main physical coupling among longitudinal motion, lateral motion, yaw dynamics, and wheel rotational behavior. Its parameter vector $\theta_{\mathrm{dyn}}$ is constrained within physically plausible ranges through bounded learnable parameters. The residual model, parameterized by $\theta_{\mathrm{res}}$, compensates for effects that are difficult to capture analytically, including transient tire behavior, unmodeled aerodynamic variation, local surface variation, and actuator response mismatch.
Model training based on next-state prediction error is described in Supplementary Section~S7.

Prediction accuracy is evaluated using the channel-wise ego-dynamics prediction RMSE defined in Supplementary Section~S3.1. In the experiments, the evaluated channels include longitudinal velocity, lateral velocity, yaw rate, longitudinal acceleration, and lateral acceleration, corresponding to the near-limit motion states reported in the main manuscript.

\subsubsection*{S4.2 Interaction Prediction}

The interaction prediction model captures the future evolution of multi-agent racing scenarios. The input is the structured world state $w_t$, which contains ego vehicle state, tracked opponent states, and local track context. In a track-coordinate system, each surrounding opponent is encoded from its relative longitudinal and lateral position and speed with respect to the ego vehicle, together with a learned slot embedding indicating its relative bearing. 

The ego encoder and opponent encoders produce features that a masked attention-pooling layer aggregates into a scene representation. This representation is decoded into a latent interaction intention, expressed as a lateral reference offset and a target speed, that implicitly captures racing-relevant behaviors such as attack, defense, hold, and yield. A multilayer-perceptron transition model then predicts per-agent state increments and is recursively applied to roll out future interaction evolution.

At inference time, the interaction model is recursively rolled out as part of the world model:
\begin{equation}
\hat{w}_{t:t+H}
=
W^{H}
\left(
w_t,
a_{t:t+H-1}
\right),
\label{eq:supp_wm_rollout}
\end{equation}
where \(W^H(\cdot)\) denotes recursive world-model prediction over horizon \(H\), and \(a_{t:t+H-1}\) denotes candidate action inputs or behavior hypotheses. The predicted world-state sequence \(\hat{w}_{t:t+H}\) contains both ego-state evolution and opponent-state evolution. 
The interaction transition model is trained by supervised prediction of world-state evolution as described in Supplementary Section~S7.

For evaluation, predicted opponent trajectories are compared with ground-truth trajectories using the ADE and FDE metrics defined in Supplementary Section~S3.1. These metrics are computed in the structured track-coordinate representation and quantify whether the world model provides sufficiently accurate short-horizon interaction futures for trajectory evaluation.

\subsubsection*{S4.3 Physical-Limit Representation}

The physical-limit representation models the feasible-motion boundary of the ego vehicle. At the vehicle level, this boundary is represented as a coupled longitudinal--lateral acceleration envelope. For a state indexed by \(k\), the envelope is written as
\begin{equation}
\begin{cases}
\left(\dfrac{a_{x,k}}{a_{x,\max}^{\mathrm{acc}}(v_{x,k},\mu_k)}\right)^c
+
\left(\dfrac{|a_{y,k}|}{a_{y,\max}(v_{x,k},\mu_k)}\right)^c
\le 1,
& a_{x,k}\ge 0, \\[8pt]
\left(\dfrac{|a_{x,k}|}{a_{x,\max}^{\mathrm{dec}}(v_{x,k},\mu_k)}\right)^c
+
\left(\dfrac{|a_{y,k}|}{a_{y,\max}(v_{x,k},\mu_k)}\right)^c
\le 1,
& a_{x,k}<0,
\end{cases}
\label{eq:supp_envelope_limit}
\end{equation}
where \(a_{x,k}\) and \(a_{y,k}\) denote longitudinal and lateral accelerations, \(v_{x,k}\) is the longitudinal speed, \(\mu_k\) is the road-friction coefficient, and \(c\) controls the shape of the super-ellipse envelope. The functions \(a_{x,\max}^{\mathrm{acc}}(\cdot)\), \(a_{x,\max}^{\mathrm{dec}}(\cdot)\), and \(a_{y,\max}(\cdot)\) describe acceleration, braking, and lateral acceleration capabilities under the current speed and friction condition.

A nominal physical-limit envelope \(\mathcal{E}_0\) is generated from the identified dynamics parameters \(\theta_{\mathrm{dyn}}\). The envelope provides a structured prior of the vehicle's feasible motion capability before online adaptation. Boundary points are obtained by radial search in the \((a_x,a_y)\) space over the operating speed range, and the envelope parameters are calibrated from the resulting feasible boundary.

\begin{algorithm}[!t]
\caption{Nominal physical-limit envelope generation}
\label{alg_nominal_ggv}
\footnotesize
\begin{algorithmic}[1]
\State Initialize $\mathcal{P}_{\mathrm{lim}}\leftarrow\varnothing$
\For{each speed $v_x$}
    \State $a_{\mathrm{drag}}\leftarrow f_{\mathrm{drag}}(v_x)$
    \State $a_{x,\max}^{\mathrm{eng}}\leftarrow \mathrm{Engine}(v_x)$
    \For{each direction angle $\theta$}
        \State $R\leftarrow 0$
        \While{$R<R_{\max}$}
            \State $a_x\leftarrow R\cos\theta,\quad a_y\leftarrow R\sin\theta$
            \State $[F_x^{\mathrm{tire}},F_y^{\mathrm{tire}},F_z^{\mathrm{tire}}]\leftarrow \mathrm{VehicleDynamics}(a_x-a_{\mathrm{drag}},a_y;\theta_{\mathrm{dyn}})$
            \For{each wheel}
                \If{$\mathrm{TireLimit}(F_x^{\mathrm{tire}},F_y^{\mathrm{tire}},F_z^{\mathrm{tire}};\mu_{\mathrm{ref}})>0$}
                    \State $\mathcal{P}_{\mathrm{lim}}\leftarrow \mathcal{P}_{\mathrm{lim}}\cup P_{\mathrm{feasible}}$
                    \State Update $a_{x,\max}^{\mathrm{acc}}$, $a_{x,\max}^{\mathrm{dec}}$, and $a_{y,\max}$ at $v_x$ from $P_{\mathrm{feasible}}$
                    \State \textbf{break}
                \EndIf
            \EndFor
            \State $R\leftarrow R+\Delta R$
            \State $P_{\mathrm{feasible}}\leftarrow[\min(a_x,a_{x,\max}^{\mathrm{eng}}),a_y,v_x]$
        \EndWhile
    \EndFor
\EndFor
\For{each state $(a_x,a_y,v_x)\in\mathcal{P}_{\mathrm{lim}}$}
    \State $c\leftarrow c_{\min}$
    \While{$c\le c_{\max}$}
        \State $a_{x,\max}\leftarrow a_{x,\max}^{\mathrm{acc}}$ if $a_x\ge0$, otherwise $a_{x,\max}^{\mathrm{dec}}$
        \If{$\left[{a_x}/{a_{x,\max}(v_x)}\right]^c+\left[{|a_y|}/{a_{y,\max}(v_x)}\right]^c<1$}
            \State $c(v_x)\leftarrow c-\Delta c$
            \State \textbf{break}
        \EndIf
        \State $c\leftarrow c+\Delta c$
    \EndWhile
\EndFor
\State Return $\mathcal{E}_0$
\end{algorithmic}
\end{algorithm}

Starting from the nominal envelope \(\mathcal{E}_0\), the effective physical-limit boundary is refined through world-model-based exploration and online adaptation. The effective envelope is written as
\begin{equation}
\mathcal{E}_{t}
=
\mathcal{E}
\left(
\rho_a(w_t),
\rho_e;
\mathcal{E}_{0}(\theta_{\mathrm{dyn}})
\right),
\label{eq:supp_adaptive_envelope}
\end{equation}
where \(\rho_e\) is the physical-limit exploration factor and \(\rho_a(w_t)\) is the online adaptive limit factor.

Physical-limit exploration aims to expand the nominal envelope when doing so improves performance without violating the prescribed limit criterion. A safe reinforcement-learning policy proposes an envelope expansion action
\begin{equation}
a_k
=
\left[
\Delta \rho_{e,x}^{\mathrm{dec}},
\Delta \rho_{e,x}^{\mathrm{acc}},
\Delta \rho_{e,y}
\right]
\in
[0,\Delta\rho_{\max}]^3,
\label{eq:supp_exploration_action}
\end{equation}
where \(\rho_{e,x}^{\mathrm{acc}}\), \(\rho_{e,x}^{\mathrm{dec}}\), and \(\rho_{e,y}\) scale acceleration, braking, and lateral limits, respectively. The policy input is
\begin{equation}
s_k
=
[
\Gamma_{k-1},
\Gamma_k,
\eta_{\mathrm{lim},k-1}
],
\label{eq:supp_exploration_state}
\end{equation}
where \(\Gamma_{k-1}\) and \(\Gamma_k\) denote preceding and upcoming track segments, and \(\eta_{\mathrm{lim},k-1}\) is the limit utilization of the previous segment.

The exploration reward penalizes both excessive expansion and limit violation:
\begin{equation}
\bar{r}_k
=
r_k
-
\beta
\|a_k\|_2^2
-
\lambda
c_k,
\label{eq:supp_safe_rl_reward}
\end{equation}
where
\begin{equation}
r_k
=
-
\frac{
T_k
}{
T_k(\Gamma_{\mathcal{E}_0}^{*})
},
\qquad
c_k
=
\max(\eta_{\mathrm{lim},k}-1,0)^2 .
\label{eq:supp_safe_rl_terms}
\end{equation}
Here \(T_k\) is the lap-time-related segment objective, \(\Gamma_{\mathcal{E}_0}^{*}\) is the reference optimal trajectory under the nominal envelope, and \(c_k\) penalizes violation of the hybrid physical-limit utilization.

The proposed expansion is first applied as
\begin{equation}
\tilde{\rho}_{e,k+1}
=
\rho_{e,k}
+
a_k,
\end{equation}
\begin{equation}
\tilde{\mathcal{E}}_{k+1}
=
\tilde{\rho}_{e,k+1}
\odot
\mathcal{E}_0,
\end{equation}
where \(\odot\) denotes element-wise scaling of the nominal envelope parameters. The update is accepted only if the maximum physical-limit utilization satisfies the prescribed criterion:
\begin{equation}
\rho_{e,k+1}
=
\begin{cases}
\tilde{\rho}_{e,k+1},
&
\eta^{\max}_{\mathrm{lim}}(\tilde{\mathcal{E}}_{k+1})\le 1,
\\[4pt]
\rho_{e,k},
&
\text{otherwise}.
\end{cases}
\label{eq:supp_envelope_acceptance}
\end{equation}
This rule prevents unsafe over-expansion of the feasible-motion boundary.

During online execution, the envelope is further adjusted according to operating conditions. The adaptive factor is defined as
\begin{equation}
\rho_{a}(w_t)
=
\frac{\mu_t}{\mu_{\mathrm{ref}}}
\,
\psi
\left(
\frac{
T_{\mathrm{fl}}
+
T_{\mathrm{fr}}
+
T_{\mathrm{rl}}
+
T_{\mathrm{rr}}
}{4}
\right)
\exp
\left(
-\lambda_{\mathrm{wm}}
\varepsilon_{\mathrm{ego}}^{\mathrm{wm}}
\right),
\label{eq:supp_adaptive_factor}
\end{equation}
where \(\mu_t\) and \(\mu_{\mathrm{ref}}\) denote the estimated and reference friction coefficients, \(\psi(\cdot)\) is the tire-temperature attenuation function, and \(\varepsilon_{\mathrm{ego}}^{\mathrm{wm}}\) is the world-model prediction error for the ego state.

The implemented online envelope is
\begin{equation}
\mathcal{E}_{t}
=
\rho_a(w_t)
\rho_e
\odot
\mathcal{E}_0(\theta_{\mathrm{dyn}}).
\label{eq:supp_online_envelope}
\end{equation}
Thus, the physical-limit representation evolves through two mechanisms: offline exploration enlarges the nominal feasible-motion boundary when supported by world-model rollouts, while online adaptation contracts or scales the envelope according to friction, tire temperature, and prediction uncertainty.

\subsection*{S5. Future-Aware Reasoning and Trajectory Evaluation}

Future-aware reasoning transforms the current racing world state into a selected trajectory that is safe, feasible, and competitive before execution. The module combines a high-level decision policy, a multimodal delta planner, physical-limit-based speed replanning, and world-model-based predictive evaluation. The resulting trajectory is passed to the physical-limit-constrained controller described in Supplementary Section~S6.

\subsubsection*{S5.1 Decision Policy Network}

At each time step, the decision policy receives the structured world state $w_t$, including the ego vehicle state, tracked opponent states, and local track context. The policy outputs a compact racing command
\begin{equation}
    \xi_d
    =
    [v_d,l_d],
    \label{eq:supp_decision_command}
\end{equation}
where $v_d$ denotes the target speed and $l_d$ denotes the target lateral position or offset in the track-coordinate system. This command represents the main racing intention, such as accelerating, braking, holding line, defending, or preparing an overtaking maneuver.

The decision policy uses a structured interaction encoder. The ego vehicle is encoded using Frenet-frame states, while opponent vehicles are organized into semantic slots according to their relative longitudinal and lateral positions. Each opponent slot contains relative position, lateral offset, velocity, availability, and slot identity. Slot-wise interaction features are aggregated into a compact scene representation and combined with the ego feature. The policy head outputs the longitudinal--lateral racing command $\xi_d$. During training, stochastic action sampling is used for exploration; during deployment, deterministic inference is used for stable online decision-making.

Given the current ego state $x_t$ and the decision command $\xi_d$, a nominal reference trajectory is generated over a finite horizon $H$:
\begin{equation}
    \tau_t^{\mathrm{ref}}
    =
    f_{\mathrm{tran}}(x_t,\xi_d,H),
    \label{eq:supp_ref_traj}
\end{equation}
where $f_{\mathrm{tran}}(\cdot)$ denotes the transition model that converts the high-level racing command into a reference trajectory. This trajectory represents the main decision branch implied by the current racing intention.

\begin{figure}[!htb]
    \centering
    \includegraphics[width=1.0\linewidth]{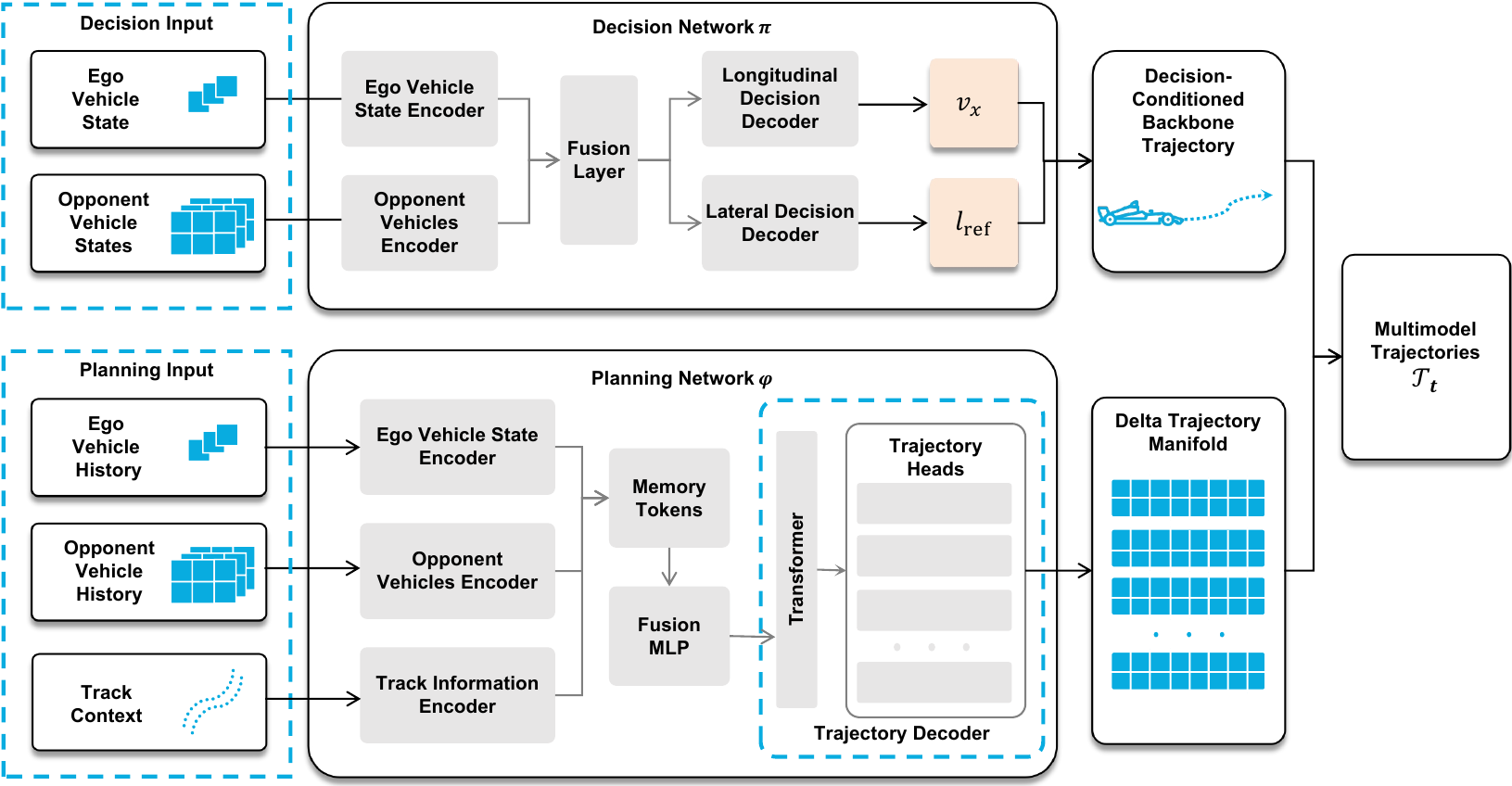}
    \caption{
    The architecture of the multimodal planning networks.
    }
    \label{SF7_planning_framework}
\end{figure}

The multimodal planning framework including the decision network and the planning network is shown in Fig.~\ref{SF7_planning_framework}. The decision network generates the decision-conditioned backbone trajectory, and the planning network produces the delta trajectory manifold for multimodal trajectory generation.

\subsubsection*{S5.2 Multimodal Delta Planner}

Competitive racing interactions are multimodal: the same observed world state may admit several plausible maneuvers, such as left-side overtaking, right-side overtaking, braking behind an opponent, or holding the current line. To represent this multimodality, a delta planner generates mode-specific trajectory deviations around the main reference trajectory.

Given recent world-state history $w_{t-h:t}$, the planner outputs $K$ candidate trajectory offsets,
\begin{equation}
    \Delta \tau_t^k
    =
    \phi^k(w_{t-h:t};\theta_{\phi}),
    \qquad
    k=1,\ldots,K,
    \label{eq:supp_delta_traj}
\end{equation}
where $\phi^k(\cdot)$ denotes the $k$-th output mode of the multimodal planner and $\theta_{\phi}$ denotes the planner parameters. Candidate trajectories are then formed as
\begin{equation}
    \tau_t^k
    =
    \tau_t^{\mathrm{ref}}
    +
    \Delta \tau_t^k.
    \label{eq:supp_candidate_traj}
\end{equation}
To diversify longitudinal behavior, a set of speed factors are prescribed.
The multimodal trajectories are scaled with speed factors, producing candidate trajectories for a subsequent world-model rollout evaluation.

\begin{figure}[!htb]
    \centering
    \includegraphics[width=0.6\linewidth]{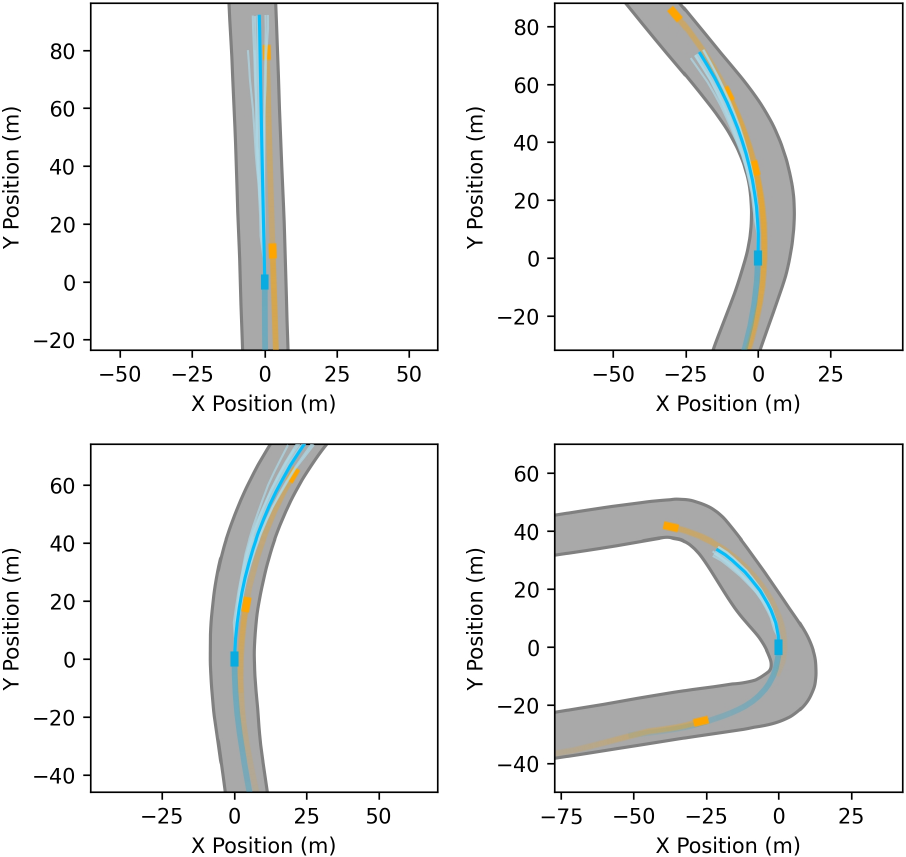}
    \caption{
    Planning results in diverse racing scenes.
    }
    \label{SF8_planning_trajectory}
\end{figure}

The planner encodes ego history, opponent history, and local track context using transformer-based modules. The resulting latent representation is decoded through multiple output heads, allowing different modes to specialize in different racing behaviors. This multimodal structure prevents the reasoning process from committing prematurely to a single maneuver before future interaction outcomes are evaluated.
Planning results in diverse racing scenes are shown in Fig.~\ref{SF8_planning_trajectory}. The deeper blue center trajectory is the decision-conditioned backbone trajectory. The light blue trajectories are the multimodal trajectories.

\subsubsection*{S5.3 Speed Replanning under Physical-Limit Constraints}

The prescribed speed factors diversify the longitudinal profiles of the multimodal trajectories, while the following speed-replanning step enforces compatibility with the current physical-limit envelope.

The candidate trajectories are further regularized by the current feasible-motion envelope $\mathcal{E}_t$. This step converts each candidate into an executable trajectory with a speed profile compatible with the physical capability of the ego vehicle:
\begin{equation}
    \mathcal{T}_t
    =
    \left\{
    R(\tau_t^k,\mathcal{E}_t)
    \right\}_{k=1}^{K},
    \label{eq:supp_regularized_candidate_set}
\end{equation}
where $R(\tau_t^k,\mathcal{E}_t)$ denotes speed replanning under the adaptive physical-limit envelope. The replanning process respects the coupled longitudinal--lateral acceleration constraint induced by $\mathcal{E}_t$.

\begin{algorithm}[htb]
\caption{Speed replanning under the feasible-motion envelope}
\label{alg:speed_planning}
\footnotesize
\begin{algorithmic}[1]
\Require Candidate trajectory $\tau_t^k$, current longitudinal speed $v_x^0$, feasible-motion envelope $\mathcal{E}_t$
\State Extract arc-length increments $\Delta s_{t+h}^k$ and curvature $\kappa_{t+h}^k$ from $\tau_t^k$
\State $v_{t+h}^k \leftarrow \min\!\left(\sqrt{a_{y,\max}(v_x^0,\mathcal{E}_t)/\max(|\kappa_{t+h}^k|,\epsilon)},\, v_{\max}\right)$
\State Set $v_t^k \leftarrow v_x^0$
\For{$h=0$ to $H-1$}
    \State $a_{y,t+h} \leftarrow (v_{t+h}^k)^2 \kappa_{t+h}^k$
    \State $a_{x,t+h}^{+} \leftarrow \mathcal{E}_t(a_{y,t+h},v_{t+h}^k)$
    \State $v_{t+h+1}^k \leftarrow \min\!\left(v_{t+h+1}^k,\sqrt{\max((v_{t+h}^k)^2+2a_{x,t+h}^{+}\Delta s_{t+h}^k,0)}\right)$
\EndFor
\For{$h=H$ down to $1$}
    \State $a_{y,t+h} \leftarrow (v_{t+h}^k)^2 \kappa_{t+h}^k$
    \State $a_{x,t+h}^{-} \leftarrow \mathcal{E}_t(a_{y,t+h},v_{t+h}^k)$
    \State $v_{t+h-1}^k \leftarrow \min\!\left(v_{t+h-1}^k,\sqrt{\max((v_{t+h}^k)^2+2a_{x,t+h}^{-}\Delta s_{t+h}^k,0)}\right)$
\EndFor
\State $a_{x,t+h}^k \leftarrow \big((v_{t+h+1}^k)^2-(v_{t+h}^k)^2\big)/(2\Delta s_{t+h}^k+\epsilon)$
\State Update $\tau_t^k$ with $\{v_{t+h}^k,a_{x,t+h}^k\}$
\State Densify and resample $\tau_t^k$ at fixed interval $\Delta t$
\end{algorithmic}
\end{algorithm}

This replanning step provides a direct connection between physical-limit representation and future-aware reasoning. A candidate trajectory is not evaluated solely by its geometric path or racing intention; it is first converted into a physically compatible trajectory under the current envelope $\mathcal{E}_t$.

\subsubsection*{S5.4 Parallel World-Model Rollouts}

After candidate generation and physical-limit regularization, each candidate trajectory is evaluated through recursive world-model rollouts. Starting from the current world state $w_t$, the world model predicts the future evolution induced by the candidate:
\begin{equation}
    \hat{w}_{t:t+H}^{\,k}
    =
    W^H(w_t,a_{t:t+H-1}^{\,k}),
    \label{eq:supp_parallel_rollout}
\end{equation}
where $a_{t:t+H-1}^{\,k}$ denotes the behavior associated with candidate trajectory $\tau_t^k$. The resulting set of parallel futures is
\begin{equation}
    \mathcal{F}_t
    =
    \left\{
    \left(
    \tau_t^k,
    \hat{w}_{t:t+H}^{\,k}
    \right)
    \mid
    \tau_t^k\in\mathcal{T}_t
    \right\}.
    \label{eq:supp_parallel_futures}
\end{equation}

Each predicted future contains ego-motion evolution, opponent-motion evolution, and feasibility information under the current physical-limit representation. This allows the agent to assess whether a maneuver that appears competitive at the current instant remains safe and executable over the prediction horizon.

\subsubsection*{S5.5 Safety, Feasibility, and Speed Scoring}

Each candidate future is evaluated by a task-oriented score that combines safety, feasibility, and speed efficiency. The total score is written as
\begin{equation}
    S^k
    =
    S_{\mathrm{safety}}^k
    +
    S_{\mathrm{speed}}^k
    -
    S_{\mathrm{violation}}^k,
    \label{eq:supp_score_total}
\end{equation}
where $S_{\mathrm{safety}}^k$ penalizes predicted collision or boundary risk, $S_{\mathrm{speed}}^k$ rewards racing progress and speed efficiency, and $S_{\mathrm{violation}}^k$ penalizes predicted physical-limit violation or loss of control.

The track-boundary risk is defined as
\begin{equation}
\rho_{\mathrm{track}}^k
=
\max_{h=1,\ldots,H}
\max
\left[
\left(l_{t+h}^k-l_L(s_{t+h}^k)\right)_+,
\left(l_R(s_{t+h}^k)-l_{t+h}^k\right)_+
\right],
\label{eq:supp_track_risk}
\end{equation}
where $(x)_+=\max(0,x)$, $l_L(\cdot)$ and $l_R(\cdot)$ are the safe left and right track boundaries, and $(s,l)$ denotes the track-coordinate position.

The opponent-collision risk is
\begin{equation}
\rho_{\mathrm{opponent}}^k
=
\max_{j\in\mathcal{V}_t,\;h\in\{1,\ldots,H\}}
\left[
\left(
d_s^{\mathrm{th}}
-
\left|s_{t+h}^k-\hat{s}_{t+h}^{j}\right|
\right)_+
\left(
d_l^{\mathrm{th}}
-
\left|l_{t+h}^k-\hat{l}_{t+h}^{j}\right|
\right)_+
\right],
\label{eq:supp_opponent_risk}
\end{equation}
where $\mathcal{V}_t$ is the set of valid opponent vehicles, $(\hat{s}_{t+h}^{j},\hat{l}_{t+h}^{j})$ is the predicted track-coordinate position of opponent $j$, and $d_s^{\mathrm{th}}$ and $d_l^{\mathrm{th}}$ are longitudinal and lateral safety thresholds.

The safety score is
\begin{equation}
S_{\mathrm{safety}}^{k}
=
\exp
\left(
-
\rho_{\mathrm{opponent}}^k
-
\rho_{\mathrm{track}}^k
-
\lambda_r
\max_{h=1,\ldots,H}|r_{t+h}^k|
\right),
\label{eq:supp_safety_score}
\end{equation}
where $r_{t+h}^k$ is the yaw rate along candidate trajectory $k$, and $\lambda_r$ weights yaw-rate-related instability risk.

Physical feasibility is evaluated using the hybrid physical-limit utilization defined in Supplementary Section~S3:
\begin{equation}
\eta_{\mathrm{lim},t:t+H}^{k}
=
\max_{h=1,\ldots,H}
\eta_{\mathrm{lim},t+h}^{k}.
\label{eq:supp_future_limit_utilization}
\end{equation}
The physical-limit violation penalty is
\begin{equation}
S_{\mathrm{violation}}^k
=
\lambda_{\eta}
\left(
\max
\left(
\eta_{\mathrm{lim},t:t+H}^{k}
-
\eta_{\mathrm{th}},
0
\right)
\right)^2,
\label{eq:supp_limit_violation_penalty}
\end{equation}
where $\eta_{\mathrm{th}}$ is the prescribed limit threshold and $\lambda_{\eta}$ is a penalty weight.

The speed score is defined by the terminal speed ratio,
\begin{equation}
S_{\mathrm{speed}}^k
=
\frac{
v_{x,t+H}^k
}{
v_x^{\mathrm{ref}}
(s_{t+H}^k;\mathcal{E}_0)
},
\label{eq:supp_score_speed}
\end{equation}
where $v_x^{\mathrm{ref}}(s_{t+H}^k;\mathcal{E}_0)$ is the reference speed at terminal progress $s_{t+H}^k$ under the nominal envelope.

The final selected trajectory is
\begin{equation}
    \tau_t^\ast
    =
    \arg\max_{\tau_t^k\in\mathcal{T}_t}
    S^k.
    \label{eq:supp_selected_candidate}
\end{equation}
This selection rule favors trajectories that are interaction-safe, physically feasible, and competitive in speed before they are committed to the physical vehicle.

\begin{figure}[!htb]
    \centering
    \includegraphics[width=0.65\linewidth]{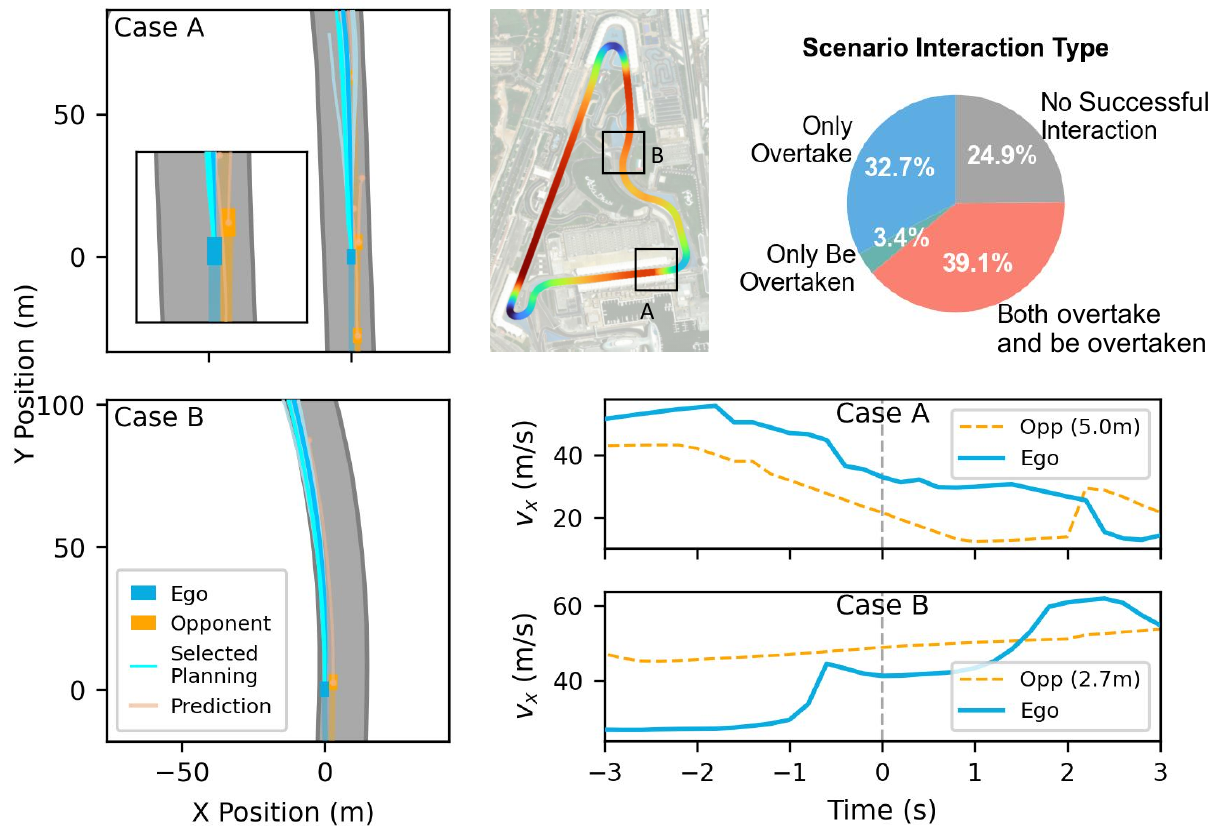}
    \caption{
    Interaction cases and scenario interaction statistics.
    }
    \label{SF9_interaction}
\end{figure}

The resulting interaction cases and the interaction statistics of the scenarios are shown in Fig.~\ref{SF9_interaction}.

\subsection*{S6. Physical-Limit-Constrained Motion Control}

\begin{figure}[!htb]
    \centering
    \includegraphics[width=0.8\linewidth]{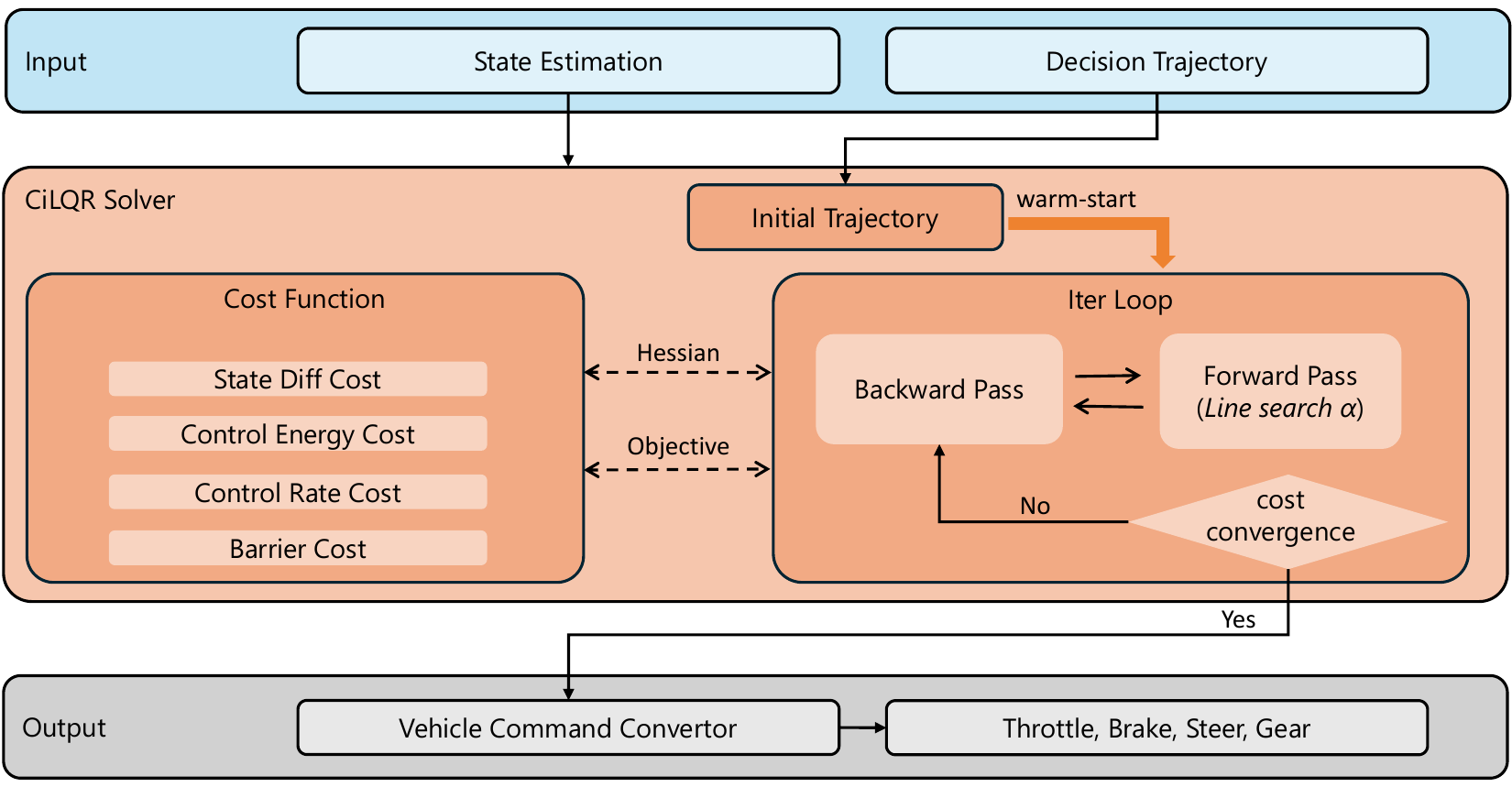}
    \caption{
    Physical-limit-constrained CiLQR control framework.
    }
    \label{SF10_control_framework}
\end{figure}

The physical-limit-constrained controller converts the trajectory selected by future-aware reasoning into executable control commands. Given the selected trajectory $\tau_t^\ast$ and the world-model-informed feasible-motion envelope $\mathcal{E}_t$, the controller tracks the reference trajectory while enforcing vehicle dynamics, road-boundary constraints, actuator limits, and coupled longitudinal--lateral physical-limit constraints. This section provides the vehicle dynamics, linearization, barrier derivatives, and constrained iLQR implementation used by the controller.
The architecture of the motion controller is illustrated in Fig.~\ref{SF10_control_framework}.

\subsubsection*{S6.1 Control-Oriented Vehicle Dynamics}

The control-oriented vehicle model uses the state
\begin{equation}
x =
[
X,\;
Y,\;
\psi,\;
v_x,\;
v_y,\;
r
]^\top,
\label{eq:supp_control_state}
\end{equation}
where $(X,Y)$ denotes the position in the global frame, $\psi$ is the heading angle, $v_x$ and $v_y$ are the longitudinal and lateral velocities in the body frame, and $r=\dot{\psi}$ is the yaw rate. The control input is
\begin{equation}
u =
[
a_x,\;
\delta
]^\top,
\label{eq:supp_control_input}
\end{equation}
where $a_x$ is the commanded longitudinal acceleration and $\delta$ is the front-wheel steering angle.

The continuous-time vehicle dynamics are
\begin{equation}
\left\{
\begin{aligned}
    \dot X &= v_x \cos\psi - v_y \sin\psi, 
    &\qquad \dot v_x &= a_x + v_y r, \\
    \dot Y &= v_x \sin\psi + v_y \cos\psi, 
    &\qquad \dot v_y &= \frac{F_{yf}\cos\delta+F_{yr}}{m} - v_x r, \\
    \dot\psi &= r, 
    &\qquad \dot r &= \frac{l_fF_{yf}\cos\delta-l_rF_{yr}}{I_z}.
\end{aligned}
\right.
\label{eq:supp_dyn_system}
\end{equation}
Here $m$ is the vehicle mass, $I_z$ is the yaw moment of inertia, $l_f$ and $l_r$ are the distances from the center of gravity to the front and rear axles, and $F_{yf}$ and $F_{yr}$ are the front and rear lateral tire forces.

The discrete-time transition is obtained by Euler integration:
\begin{equation}
x_{k+1}
=
x_k
+
f_{\mathrm{dyn}}(x_k,u_k;\theta_{\mathrm{dyn}})\Delta t,
\label{eq:supp_euler}
\end{equation}
where $f_{\mathrm{dyn}}(\cdot)$ denotes the control-oriented dynamics model and $\theta_{\mathrm{dyn}}$ contains the physical parameters calibrated from the dynamics component of the world model.

\subsubsection*{S6.2 Dynamics Linearization and Jacobians}

The iLQR backward pass requires local linearization of the discrete dynamics. Let $f_c(x,u)$ denote the continuous-time dynamics in Eq.~(\ref{eq:supp_dyn_system}). The state Jacobian is
\begin{equation}
\frac{\partial f_c}{\partial x}
=
\begin{bmatrix}
0 & 0 & f_{1,3} & \cos\psi & -\sin\psi & 0 \\
0 & 0 & f_{2,3} & \sin\psi & \cos\psi & 0 \\
0 & 0 & 0 & 0 & 0 & 1 \\
0 & 0 & 0 & 0 & r & v_y \\
0 & 0 & 0 & f_{5,4} & f_{5,5} & f_{5,6} \\
0 & 0 & 0 & f_{6,4} & f_{6,5} & f_{6,6}
\end{bmatrix},
\label{eq:supp_A_jacobian}
\end{equation}
where
\begin{equation}
f_{1,3}
=
-v_x\sin\psi-v_y\cos\psi,
\qquad
f_{2,3}
=
v_x\cos\psi-v_y\sin\psi.
\label{eq:supp_A_pos}
\end{equation}

The lateral and yaw dynamics terms are
\begin{align}
f_{5,4}
&=
\frac{\cos\delta}{m}
\frac{\partial F_{yf}}{\partial v_x}
+
\frac{1}{m}
\frac{\partial F_{yr}}{\partial v_x}
-
r,
\\
f_{5,5}
&=
\frac{\cos\delta}{m}
\frac{\partial F_{yf}}{\partial v_y}
+
\frac{1}{m}
\frac{\partial F_{yr}}{\partial v_y},
\\
f_{5,6}
&=
\frac{\cos\delta}{m}
\frac{\partial F_{yf}}{\partial r}
+
\frac{1}{m}
\frac{\partial F_{yr}}{\partial r}
-
v_x,
\\
f_{6,4}
&=
\frac{l_f\cos\delta}{I_z}
\frac{\partial F_{yf}}{\partial v_x}
-
\frac{l_r}{I_z}
\frac{\partial F_{yr}}{\partial v_x},
\\
f_{6,5}
&=
\frac{l_f\cos\delta}{I_z}
\frac{\partial F_{yf}}{\partial v_y}
-
\frac{l_r}{I_z}
\frac{\partial F_{yr}}{\partial v_y},
\\
f_{6,6}
&=
\frac{l_f\cos\delta}{I_z}
\frac{\partial F_{yf}}{\partial r}
-
\frac{l_r}{I_z}
\frac{\partial F_{yr}}{\partial r}.
\end{align}

The control Jacobian is
\begin{equation}
\frac{\partial f_c}{\partial u}
=
\begin{bmatrix}
0 & 0 & 0 & 1 & 0 & 0 \\
0 & 0 & 0 & 0 & g_{5,2} & g_{6,2}
\end{bmatrix}^{\mathsf{T}},
\label{eq:supp_B_jacobian}
\end{equation}
where
\begin{align}
g_{5,2}
&=
\frac{1}{m}
\left(
\cos\delta
\frac{\partial F_{yf}}{\partial \delta}
-
\sin\delta F_{yf}
\right),
\\
g_{6,2}
&=
\frac{l_f}{I_z}
\left(
\cos\delta
\frac{\partial F_{yf}}{\partial \delta}
-
\sin\delta F_{yf}
\right).
\end{align}
For the linear tire-force approximation used in the controller,
\begin{equation}
\frac{\partial F_{yf}}{\partial \delta}
=
-C_f,
\end{equation}
so that
\begin{align}
g_{5,2}
&=
\frac{1}{m}
\left(
-C_f\cos\delta-\sin\delta F_{yf}
\right),
\\
g_{6,2}
&=
\frac{l_f}{I_z}
\left(
-C_f\cos\delta-\sin\delta F_{yf}
\right).
\end{align}

The discrete-time Jacobians used by iLQR are obtained from the Euler-discretized model:
\begin{equation}
A_k
=
I
+
\Delta t
\frac{\partial f_c}{\partial x}
\bigg|_{x_k,u_k},
\qquad
B_k
=
\Delta t
\frac{\partial f_c}{\partial u}
\bigg|_{x_k,u_k}.
\label{eq:supp_discrete_jacobians}
\end{equation}

\subsubsection*{S6.3 Tracking, Smoothness, and Constraint Costs}

Given the selected trajectory $\tau_t^\ast$, the controller solves the finite-horizon problem
\begin{equation}
u_t^\ast
=
\arg\min_{u_{t:t+H_c-1}}
J(x_t,u_{t:t+H_c-1};\tau_t^\ast,\mathcal{E}_t),
\label{eq:supp_cilqr_opt}
\end{equation}
subject to vehicle dynamics and inequality constraints. The cost is decomposed as
\begin{equation}
J
=
J_{\mathrm{track}}
+
J_{\mathrm{barrier}}.
\label{eq:supp_control_cost_decomp}
\end{equation}

The tracking and smoothness cost is
\begin{equation}
J_{\mathrm{track}}
=
\sum_{j=0}^{H_c-1}
\left(
\|x_{t+j}-x_{t+j}^{\mathrm{ref}}\|_{Q}^{2}
+
\|u_{t+j}\|_{R}^{2}
+
\|\Delta u_{t+j}\|_{P}^{2}
\right),
\label{eq:supp_tracking_cost}
\end{equation}
where $x_{t+j}^{\mathrm{ref}}$ is induced by $\tau_t^\ast$, and $Q$, $R$, and $P$ are weighting matrices for tracking error, control effort, and control smoothness.

Inequality constraints are imposed through exponential barrier functions:
\begin{equation}
J_{\mathrm{barrier}}
=
\sum_{k,i}
q_1^{(i)}
\exp
\left(
q_2^{(i)}c_i(x_k,u_{k-1};\mathcal{E}_k)
\right),
\label{eq:supp_barrier_cost}
\end{equation}
where $c_i\leq0$ indicates satisfaction of the corresponding constraint and $c_i>0$ indicates violation. The constraints include road boundaries, velocity limits, lateral velocity limits, yaw-rate limits, input limits, and the physical-limit envelope.

\subsubsection*{S6.4 Physical-Limit Barrier Terms}

The physical-limit constraint is induced by the coupled acceleration envelope $\mathcal{E}_k$. In the simplified quadratic form used for barrier differentiation, the constraint is expressed as
\begin{equation}
c_{\mathrm{lim}}
=
\frac{a_x^2}{a_{x,\max}^2}
+
\frac{a_y^2}{a_{y,\max}^2}
-
1,
\label{eq:supp_limit_constraint_quad}
\end{equation}
where $a_{x,\max}$ and $a_{y,\max}$ are selected from the current feasible-motion envelope according to speed, road condition, and acceleration or braking regime. A constraint violation corresponds to $c_{\mathrm{lim}}>0$.

The exponential barrier for the physical-limit constraint is
\begin{equation}
B(c_{\mathrm{lim}})
=
q_1^{\mathrm{lim}}
\exp
\left(
q_2^{\mathrm{lim}}c_{\mathrm{lim}}
\right).
\label{eq:supp_limit_barrier}
\end{equation}
For a generic variable vector $z$, its gradient and Gauss--Newton Hessian approximation are
\begin{equation}
\nabla_z B
=
q_2^{\mathrm{lim}}
B(c_{\mathrm{lim}})
\nabla_z c_{\mathrm{lim}},
\label{eq:supp_barrier_grad}
\end{equation}
\begin{equation}
\nabla_z^2 B
\approx
(q_2^{\mathrm{lim}})^2
B(c_{\mathrm{lim}})
(\nabla_z c_{\mathrm{lim}})
(\nabla_z c_{\mathrm{lim}})^\top.
\label{eq:supp_barrier_hess}
\end{equation}

The corresponding control gradient and Hessian contributions are
\begin{align}
l_{u}^{\mathrm{lim}}
&=
q_2^{\mathrm{lim}}
B(c_{\mathrm{lim}})
\begin{bmatrix}
2a_x/a_{x,\max}^2\\
0
\end{bmatrix},
\label{eq:supp_lu_limit}
\\
l_{uu}^{\mathrm{lim}}
&=
(q_2^{\mathrm{lim}})^2
B(c_{\mathrm{lim}})
\begin{bmatrix}
4a_x^2/a_{x,\max}^4 & 0\\
0 & 0
\end{bmatrix}.
\label{eq:supp_luu_limit}
\end{align}

Using the approximation \(a_y\approx v_x r\), the corresponding state gradient contribution is
\begin{equation}
l_x^{\mathrm{lim}}
=
q_2^{\mathrm{lim}}
B(c_{\mathrm{lim}})
\begin{bmatrix}
0\\
0\\
0\\
2v_x r^2/a_{y,\max}^2\\
0\\
2v_x^2 r/a_{y,\max}^2
\end{bmatrix},
\label{eq:supp_lx_limit}
\end{equation}
and the Gauss--Newton state Hessian contribution is
\begin{equation}
l_{xx}^{\mathrm{lim}}
=
(q_2^{\mathrm{lim}})^2
B(c_{\mathrm{lim}})
\frac{\partial c_{\mathrm{lim}}}{\partial x}
\left(
\frac{\partial c_{\mathrm{lim}}}{\partial x}
\right)^\top.
\label{eq:supp_lxx_limit}
\end{equation}
These derivatives are accumulated with the nominal stage-cost derivatives and other box-constraint barrier terms before the iLQR backward pass constructs the local quadratic model.

\subsubsection*{S6.5 Box Constraints and Numerical Safeguards}

The box constraints are summarized in Table~\ref{tab:supp_box_constraints}. Each constraint $c_i$ is defined such that $c_i>0$ indicates violation.

\begin{table}[htb]
    \centering
    \caption{Box constraints and non-zero derivatives used by the constrained controller.}
    \label{tab:supp_box_constraints}
    \begin{tabular}{@{}lcl@{}}
        \toprule
        Constraint & $c_i$ & Non-zero derivative \\
        \midrule
        Acceleration upper & $a_x-a_{\max}$ & $\partial c_i/\partial a_x = +1$ \\
        Acceleration lower & $a_{\min}-a_x$ & $\partial c_i/\partial a_x = -1$ \\
        Steering upper & $\delta-\delta_{\max}$ & $\partial c_i/\partial \delta = +1$ \\
        Steering lower & $-\delta_{\max}-\delta$ & $\partial c_i/\partial \delta = -1$ \\
        Velocity upper & $v-v_{\max}$ & $\partial c_i/\partial v = +1$ \\
        Velocity lower & $v_{\min}-v$ & $\partial c_i/\partial v = -1$ \\
        Lateral upper & $d_k-d_{\max}$ & $\partial c_i/\partial p = \hat{n}$ \\
        Lateral lower & $d_{\min}-d_k$ & $\partial c_i/\partial p = -\hat{n}$ \\
        \bottomrule
    \end{tabular}
\end{table}

Here $d_k$ denotes the signed lateral distance to the reference line, $\hat{n}$ is the unit normal at the nearest reference point, and $d_{\max}$ and $d_{\min}$ are derived from road width and vehicle width. Each barrier is parameterized by a pair $(q_1^{(i)},q_2^{(i)})$.

Several numerical safeguards are applied. First, for $|v_x|<v_{\mathrm{thr}}$, the physical-limit constraint is set to \(c_{\mathrm{lim}}=-1\) and its derivatives are set to zero. Second, the acceleration envelope switches between traction and braking bounds according to the sign of $a_x$. Third, barrier values are monitored to avoid numerical overflow in the exponential function. Fourth, the final control sequence is hard-clamped to box limits after the forward rollout.

\subsubsection*{S6.6 CiLQR Solver}

The constrained iLQR solver minimizes the cost in Eq.~(\ref{eq:supp_control_cost_decomp}) over a finite horizon by iteratively linearizing the dynamics and quadratizing the cost. Starting from the terminal time step, the backward pass computes
\begin{equation}
Q_x
=
l_x
+
A^\top V_x,
\end{equation}
\begin{equation}
Q_u
=
l_u
+
B^\top V_x,
\end{equation}
\begin{equation}
Q_{xx}
=
l_{xx}
+
A^\top V_{xx}A,
\end{equation}
\begin{equation}
Q_{uu}
=
l_{uu}
+
B^\top V_{xx}B,
\end{equation}
\begin{equation}
Q_{ux}
=
l_{ux}
+
B^\top V_{xx}A.
\end{equation}

The feedforward correction and feedback gain are
\begin{equation}
d_k
=
-
(Q_{uu}+\lambda I)^{-1}Q_u,
\label{eq:supp_feedforward}
\end{equation}
\begin{equation}
K_k
=
-
(Q_{uu}+\lambda I)^{-1}Q_{ux},
\label{eq:supp_feedback_gain}
\end{equation}
where $\lambda>0$ is a Tikhonov regularization parameter.

The forward rollout applies the affine feedback update
\begin{equation}
u_k^{\mathrm{new}}
=
u_k
+
\alpha d_k
+
\gamma K_k
(x_k^{\mathrm{new}}-x_k),
\label{eq:supp_cilqr_update}
\end{equation}
where $\alpha\in(0,1]$ is a line-search parameter and $\gamma\in[0,1]$ scales the feedback term. If the rollout reduces the objective, the update is accepted and the regularization is relaxed. Otherwise, the regularization is increased to produce a more conservative update.

Algorithm~\ref{alg:supp_cilqr_ggv} summarizes the constrained CiLQR implementation.

\begin{algorithm}[htb]
\caption{Constrained CiLQR with physical-limit envelope}
\label{alg:supp_cilqr_ggv}
\footnotesize
\begin{algorithmic}[1]
\Require Initial state $x_0$, selected reference trajectory $\tau_t^\ast$, feasible-motion envelope $\mathcal{E}_t$
\State Initialize controls $u_{0:N-1}$ using warm start or reference-based initialization
\State Roll out dynamics to obtain $x_{0:N}$
\State $J \leftarrow \textsc{TotalCost}(x_{0:N},u_{0:N-1};\tau_t^\ast,\mathcal{E}_t)$
\State $\lambda \leftarrow \lambda_{\mathrm{init}}$
\For{$\mathrm{iter}=1$ to $\mathrm{max\_iter}$}
    \For{$k=N$ down to $1$}
        \State Linearize dynamics to obtain $A_k,B_k$
        \State Compute stage-cost and barrier derivatives
        \State Form $Q_x,Q_u,Q_{xx},Q_{uu},Q_{ux}$
        \State Compute $d_k,K_k$
    \EndFor
    \For{$\alpha \in \{1.0,0.5,0.25,\ldots\}$}
        \State Roll out updated controls using affine feedback
        \State $J^+ \leftarrow \textsc{TotalCost}(x^+,u^+;\tau_t^\ast,\mathcal{E}_t)$
        \If{$J^+<J$}
            \State $x,u,J \leftarrow x^+,u^+,J^+$
            \State $\lambda \leftarrow \lambda\cdot\lambda_{\mathrm{decay}}$
            \State \textbf{break}
        \EndIf
    \EndFor
    \If{no line-search step reduces cost}
        \State $\lambda \leftarrow \lambda\cdot\lambda_{\mathrm{amp}}$
        \If{$\lambda>\lambda_{\max}$}
            \State \textbf{break}
        \EndIf
    \EndIf
    \If{relative cost improvement $<\epsilon$}
        \State \textbf{break}
    \EndIf
\EndFor
\State Hard-clamp final controls to actuator box limits
\State Return Optimized control sequence $u_{0:N-1}$ and predicted state sequence $x_{0:N}$
\end{algorithmic}
\end{algorithm}

The physical-limit terms in Eqs.~(\ref{eq:supp_lu_limit})--(\ref{eq:supp_lxx_limit}) are evaluated at each time step and accumulated into the stage-cost derivatives during the backward pass. This allows the controller to anticipate violations of the acceleration envelope before they appear in the executed vehicle response.

\begin{figure}[htb]
    \centering
    \includegraphics[width=\linewidth]{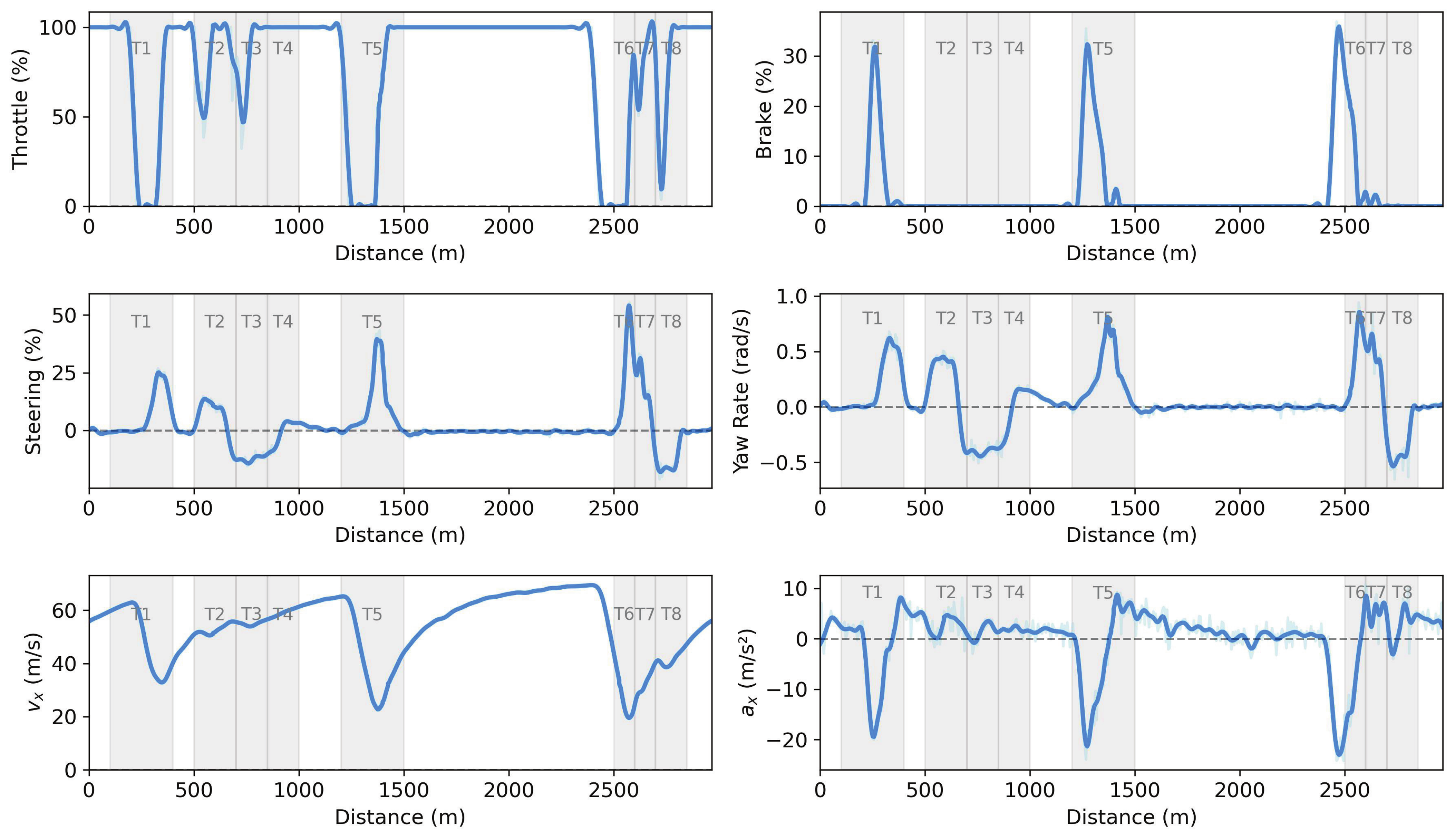}
    \caption{
    Representative full-lap control response.
    }
    \label{SF11_control_result}
\end{figure}

The resulting closed-loop control profiles are illustrated in Fig.~\ref{SF11_control_result}, showing how the controller coordinates actuation and vehicle response along the racing lap.

\begin{figure}[htb]
    \centering
    \includegraphics[width=0.8\linewidth]{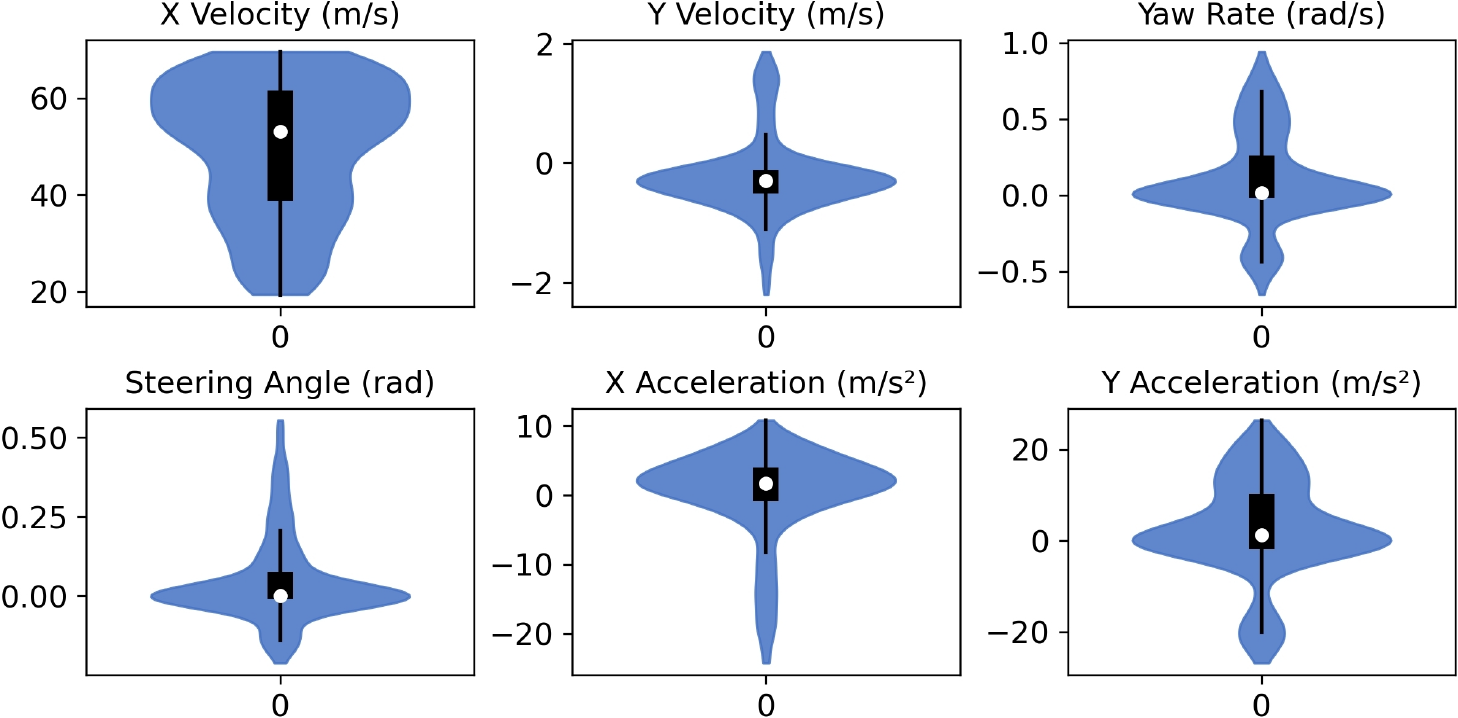}
    \caption{
    Full-lap distributions of closed-loop vehicle states and control variables.
    }
    \label{SF13_state_distribution}
\end{figure}

The closed-loop state and control variable distribution are further summarized in Fig.~\ref{SF13_state_distribution}.

\subsection*{S7. Closed-Loop World-Model and Policy Refinement}

The autonomous racing agent is refined through a closed-loop pipeline that connects real-world deployment, world-model learning, internal simulation, and policy update. Rather than updating the decision policy alone, the framework refines the predictive models of ego dynamics, interaction evolution, and physical feasibility, and then uses the refined world model to generate structured rollout data for improving future-aware reasoning and planning.

\subsubsection*{S7.1 Real Racing Dataset}

During full-scale autonomous racing, the agent collects data under real road, tire, vehicle, and interaction conditions. The resulting real-racing dataset is denoted as
\begin{equation}
\mathcal{D}_{\mathrm{real}}
=
\left\{
\left(
w_{t-h:t},\,
a_{t},\,
w_{t+1:t+H} ;\Gamma
\right)
\right\}_{t=1}^{N}.
\label{eq:supp_real_dataset}
\end{equation}
where $w_{t-h:t}$ denotes the historical structured racing world states, $a_t$ denotes the executed agent action, $w_{t+1:t+H}$ denotes the future world-state sequence, and $\Gamma$ denotes the global track context. 
Here, $a_t$ is a composite action variable that contains the reasoning output, the motion-control command $u_t$, and the actuator-level command recorded by the onboard system. 
The global track reference $\Gamma$ provides track-level information, including the racing line, track-coordinate map, road boundaries, and curvature profile, while the local track context used at each time step is included in the structured world state $w_t$.

These data provide the empirical basis for aligning the world model with the deployed system. In particular, $\mathcal{D}_{\mathrm{real}}$ is used to update ego dynamics prediction, interaction prediction, and the physical-limit representation. The physical-limit envelope is initialized conservatively and then refined through the exploration and adaptation process described in Supplementary Section~S4.

\subsubsection*{S7.2 World-Model Training}

The ego dynamics model is trained using supervised next-state prediction. The differentiable dynamics backbone and residual model are updated by minimizing
\begin{equation}
(\theta_{\mathrm{dyn}},\theta_{\mathrm{res}})
=
\arg\min_{\theta_{\mathrm{dyn}},\theta_{\mathrm{res}}}
\sum_{\mathcal{D}_{\mathrm{real}}}
\left\|
x_{t+1}
-
f_{\mathrm{dyn}}
\left(
x_t,u_t;\theta_{\mathrm{dyn}}
\right)
-
f_{\mathrm{res}}
\left(
x_{t-h:t},u_{t-h:t};\theta_{\mathrm{res}}
\right)
\right\|^2 .
\label{eq:supp_s7_update_dyn}
\end{equation}
Here $f_{\mathrm{dyn}}(\cdot)$ represents the differentiable dynamics backbone and $f_{\mathrm{res}}(\cdot)$ represents the residual model.

The interaction model is updated by supervised prediction of world-state evolution,
\begin{equation}
\theta_{\mathrm{int}}
=
\arg\min_{\theta_{\mathrm{int}}}
\sum_{\mathcal{D}_{\mathrm{real}}}
\left\|
w_{t+1}
-
W(w_t;\theta_{\mathrm{int}})
\right\|^2 ,
\label{eq:supp_s7_update_int}
\end{equation}
where $W(\cdot;\theta_{\mathrm{int}})$ denotes the learned interaction transition model. This update aligns the world model with observed ego--opponent interaction patterns in real racing.

The physical-limit representation is refined using the same real racing data and world-model rollouts. The nominal envelope $\mathcal{E}_0$ is identified from the dynamics model, while the effective envelope $\mathcal{E}_t$ is adjusted through the exploration and adaptation factors described in Supplementary Section~S4. This ensures that the physical-limit representation used by planning and control remains consistent with observed tire, road, and vehicle behavior.

\subsubsection*{S7.3 World-Model Rollout Data Generation}

After training, the refined world model is used as a structured predictor for generating additional interaction and near-limit data. Given the current world model $W$, decision policy $\pi$, multimodal planner $\phi$, and feasible-motion envelope $\mathcal{E}$, world-model-generated data are produced by
\begin{equation}
\mathcal{D}_{\mathrm{wm}}
=
\mathrm{Rollout}
\left(
W,\pi,\phi,\mathcal{E}
\right).
\label{eq:supp_wm_rollout_dataset}
\end{equation}
Each rollout contains predicted world-state transitions, candidate decisions, planned trajectories, opponent responses, and physical-limit indicators.

\begin{figure}[htb]
    \centering
    \includegraphics[width=1.0\linewidth]{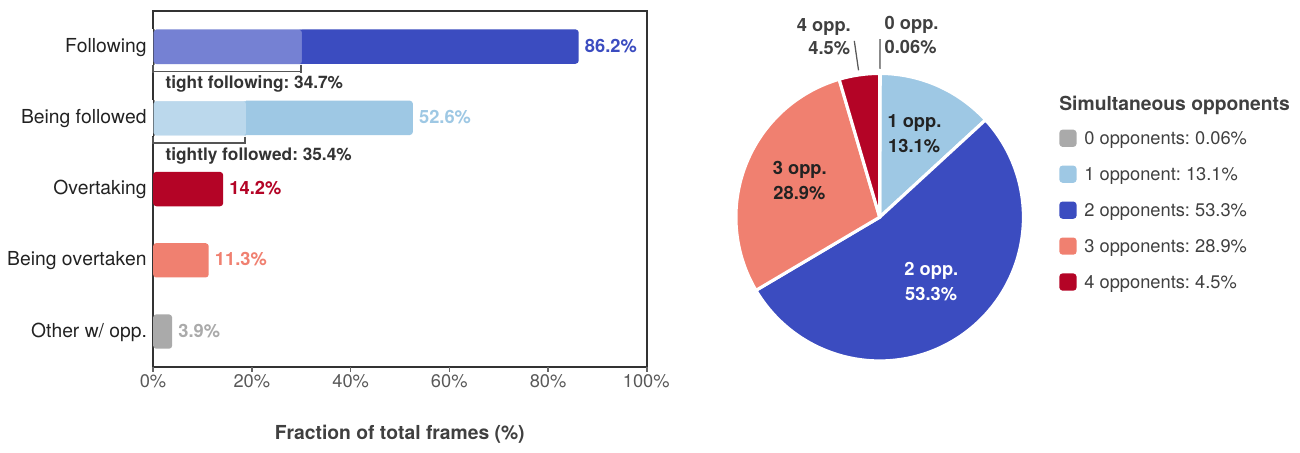}
    \caption{
    World model rollout interaction distribution.
    }
    \label{SF14_wm_interaction_distribution}
\end{figure}

Compared with real racing data alone, $\mathcal{D}_{\mathrm{wm}}$ can be generated under controlled variations of initial condition, opponent behavior, track segment, and physical-limit condition. This enables the training set to include rare adversarial interactions, near-limit maneuvers, and corner cases that are difficult or unsafe to collect exhaustively in real-world racing.
The world model rollout interaction distribution is summarized in Fig.~\ref{SF14_wm_interaction_distribution}.

The real racing dataset and the world-model-generated dataset are combined as
\begin{equation}
\mathcal{D}
=
\mathcal{D}_{\mathrm{real}}
\cup
\mathcal{D}_{\mathrm{wm}},
\label{eq:supp_hybrid_dataset_s7}
\end{equation}
where $\mathcal{D}$ is the hybrid dataset used for decision policy and planner update.

\subsubsection*{S7.4 Decision Policy Update}

The decision policy is updated using a soft actor--critic objective with conservative Q-learning and behavior-cloning regularization. Let $s_t$ denote the policy state corresponding to the racing world state $w_t$. The policy samples a racing intention
\begin{equation}
a_t^\pi
\sim
\pi_{\theta_\pi}(\cdot|s_t),
\label{eq:supp_policy_sample}
\end{equation}
where $a_t^\pi$ contains longitudinal and lateral decision components.

The critic parameters are optimized by
\begin{equation}
\begin{aligned}
L_Q(\theta_Q)
=
&
\mathbb{E}_{(s_t,a_t,s_{t+1})\sim\mathcal{D}}
\left[
\left(
Q_{\theta_Q}(s_t,a_t)-y_t
\right)^2
\right]
\\
&
+
\lambda_{\mathrm{cql}}
\left\{
\mathbb{E}_{s_t\sim\mathcal{D}}
\left[
Q_{\theta_Q}(s_t,a_t^\pi)
\right]
-
\mathbb{E}_{(s_t,a_t)\sim\mathcal{D}}
\left[
Q_{\theta_Q}(s_t,a_t)
\right]
\right\}.
\end{aligned}
\label{eq:supp_cql_loss}
\end{equation}
The first term is the temporal-difference critic loss, and the conservative term discourages overestimation for actions outside the data distribution.

The policy network is optimized by
\begin{equation}
\begin{aligned}
L_\pi(\theta_\pi)
=
\mathbb{E}_{(s_t,a_t^{\mathrm{e}})\sim\mathcal{D}}
\Big[
&
\alpha
\log
\pi_{\theta_\pi}(a_t^\pi|s_t)
-
\min(Q_1,Q_2)
\\
&
+
\lambda_{\mathrm{bc}}
\left\|
a_t^\pi
-
a_t^{\mathrm{e}}
\right\|^2
\Big],
\end{aligned}
\label{eq:supp_policy_loss}
\end{equation}
where $a_t^{\mathrm{e}}$ denotes the demonstration or reference action in the dataset. The entropy term encourages exploration during training, the critic term improves competitive decision quality, and the behavior-cloning term preserves useful behavior observed in real and world-model-generated data.

\subsubsection*{S7.5 Multimodal Planner Update}

The multimodal planner is updated using best-of-$K$ supervision. The hybrid dataset is reorganized into pairs of recent world-state history and future trajectory:
\begin{equation}
\left(
w_{t-h:t},
\tau_t^{\mathrm{gt}}
\right)
\in
\mathcal{D},
\end{equation}
where $w_{t-h:t}$ is the recent history segment and $\tau_t^{\mathrm{gt}}$ is the ground-truth or rollout-generated future trajectory.

Given the history $w_{t-h:t}$, the planner generates a set of candidate trajectories $\mathcal{T}_t$ as described in Supplementary Section~S5. Only the output mode closest to the target future trajectory is supervised:
\begin{equation}
\theta_\phi
=
\arg\min_{\theta_\phi}
\sum_{\mathcal{D}}
\left[
\min_{\tau_t^k\in\mathcal{T}_t}
\left\|
\tau_t^{\mathrm{gt}}
-
\tau_t^k
\right\|^2
\right].
\label{eq:supp_planner_update}
\end{equation}
This best-of-$K$ objective allows different output heads to specialize in different behavioral modes, such as overtaking, defending, yielding, braking, or holding the current line.

\subsubsection*{S7.6 Closed-Loop Refinement Algorithm}

Algorithm~\ref{alg:supp_closed_loop_refinement} summarizes the closed-loop refinement pipeline.

\begin{algorithm}[htb]
\caption{Closed-loop world-model and policy refinement}
\label{alg:supp_closed_loop_refinement}
\footnotesize
\begin{algorithmic}[1]
\Require Initial world model $W$, decision policy $\pi$, multimodal planner $\phi$, feasible-motion envelope $\mathcal{E}$
\For{each refinement round}
    \State Deploy the agent in real racing or high-fidelity validation scenarios
    \State Collect real racing data $\mathcal{D}_{\mathrm{real}}$
    \State Update ego dynamics model using Eq.~(\ref{eq:supp_s7_update_dyn})
    \State Update interaction model using Eq.~(\ref{eq:supp_s7_update_int})
    \State $\mathcal{D}_{\mathrm{wm}}=\mathrm{Rollout}(W,\pi,\phi,\mathcal{E})$
    \State $\mathcal{D}=\mathcal{D}_{\mathrm{real}}\cup\mathcal{D}_{\mathrm{wm}}$
    \State Update decision policy $\pi$ using Eqs.~(\ref{eq:supp_cql_loss})--(\ref{eq:supp_policy_loss})
    \State Update multimodal planner $\phi$ using Eq.~(\ref{eq:supp_planner_update})
    \State Refine feasible-motion envelope $\mathcal{E}$ as Section~S4.4
\EndFor
\State Return refined world model $W$, policy $\pi$, planner $\phi$, and feasible-motion envelope $\mathcal{E}$
\end{algorithmic}
\end{algorithm}

This loop links real-world deployment, world-model prediction, policy learning, planner update, and physical-limit regulation. Real racing data improve the fidelity of ego dynamics, interaction prediction, and physical-limit representation. The refined world model then generates structured rollouts for rare interactions and near-limit scenarios. The updated agent can subsequently collect more informative data in the next deployment round, enabling progressive improvement of cognitive reasoning and physical-limit utilization.

\subsection*{S8. Experimental Protocol, Simulation Setup, and Baselines}

This section describes the experimental protocol, simulation environment, interaction scenario construction, baseline methods, and evaluation metrics used in the experiments. 
The validation combines real-vehicle deployment with simulation testing. 
Real-vehicle deployment provides high-value data under actual racing conditions, while simulation enables repeatable evaluation of adversarial interaction, ablation studies, and failure-case reproduction.

\subsubsection*{S8.1 Real-Vehicle Deployment}

Real-vehicle experiments were conducted on the full-scale autonomous racing platform introduced in Supplementary Section~S1.1 and Fig.~\ref{SF1_vehicle}. 
During real-vehicle deployment, the onboard system operated localization, opponent perception, behavior decision and motion control modules and recorded synchronized state, control, perception, and environment data.
The real-vehicle deployment exposed the onboard system to high-speed motion, near-limit tire--road interaction, and close-proximity racing conditions. It therefore provided the real-world basis for physical grounding, world-model construction, and validation in simulation.

\begin{figure}[!htb]
    \centering
    \includegraphics[width=0.8\linewidth]{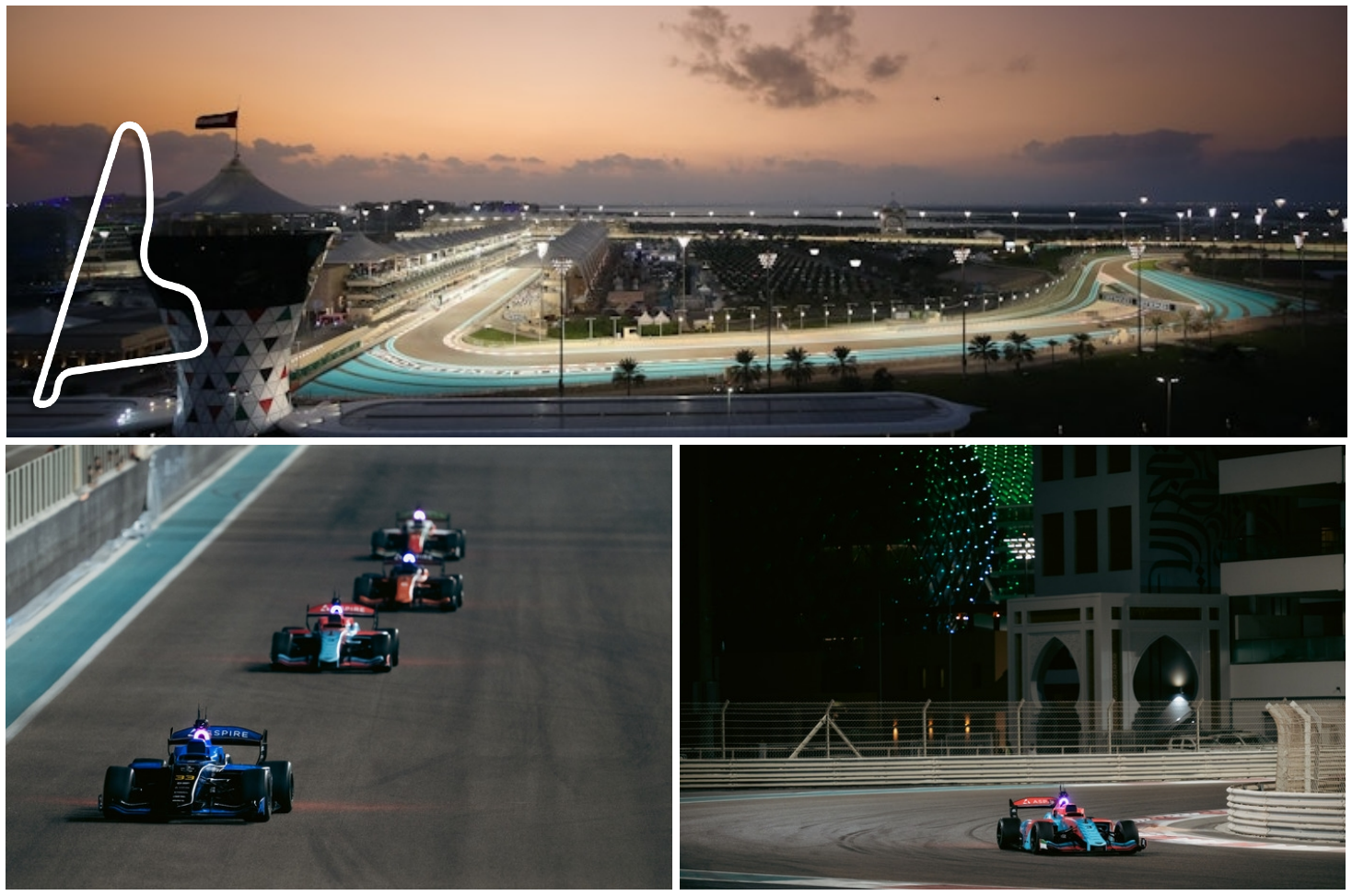}
    \caption{
    Racetrack environment for real-vehicle deployment.
    }
    \label{SF15_race_track}
\end{figure}

The real-vehicle deployment was carried out in the context of the Abu Dhabi Autonomous Racing League (A2RL). A2RL is a full-scale autonomous racing competition designed to evaluate autonomous-driving software under high-speed and competitive racing conditions. 
The real-vehicle experiment was conducted on the North Circuit of the Yas Marina F1 circuit in Abu Dhabi, which provides a full-scale racing environment with high-speed straights and consecutive cornering segments. 
The track has a total length of $3002~\mathrm{m}$ and contains $8$ corners.
The racetrack environment for real-vehicle deployment is shown in Fig.~\ref{SF15_race_track}.

The real-vehicle deployment exposed the onboard system to high-speed motion, near-limit tire--road interaction, and close-proximity racing conditions. It therefore provided the real-world basis for physical grounding, world-model construction, and simulation-based validation.

\subsubsection*{S8.2 Real-Racing Data Collection}

Real-racing data were collected during full-scale autonomous racing deployment under high-speed operation, near-limit tire--road interaction, and adversarial multi-agent interaction. 
Each record follows the transition structure defined in Supplementary Section~S7 and contains the historical world states, the executed action, and the future world-state sequence, conditioned on the global track context.
The onboard system recorded multi-source data streams at different sampling frequencies. These data streams were time-synchronized and resampled to $10~\mathrm{Hz}$ to form aligned real-racing samples.

The real-racing dataset is used in three ways. First, it provides ground-truth or reference signals for evaluating localization, ego-state estimation, opponent tracking, and prediction accuracy. Second, it is used to train and refine the ego dynamics model, interaction model, and physical-limit representation. Third, it provides real-world cases, including overtaking and spin events, that are reproduced in simulation for controlled analysis of prediction, decision-making, and failure recovery.

\subsubsection*{S8.3 Autoverse Simulation Setup}

Controlled validation was conducted in the Autoverse simulator with virtual opponents. 
Autoverse provides high-fidelity vehicle dynamics simulation, while the virtual opponent vehicles enable multi-vehicle interaction scenarios. 
This setup enables systematic evaluation of the refined agent under controlled initial conditions, track configurations, opponent behaviors, and physical-limit conditions.

\begin{figure}[!htb]
    \centering
    \includegraphics[width=0.8\linewidth]{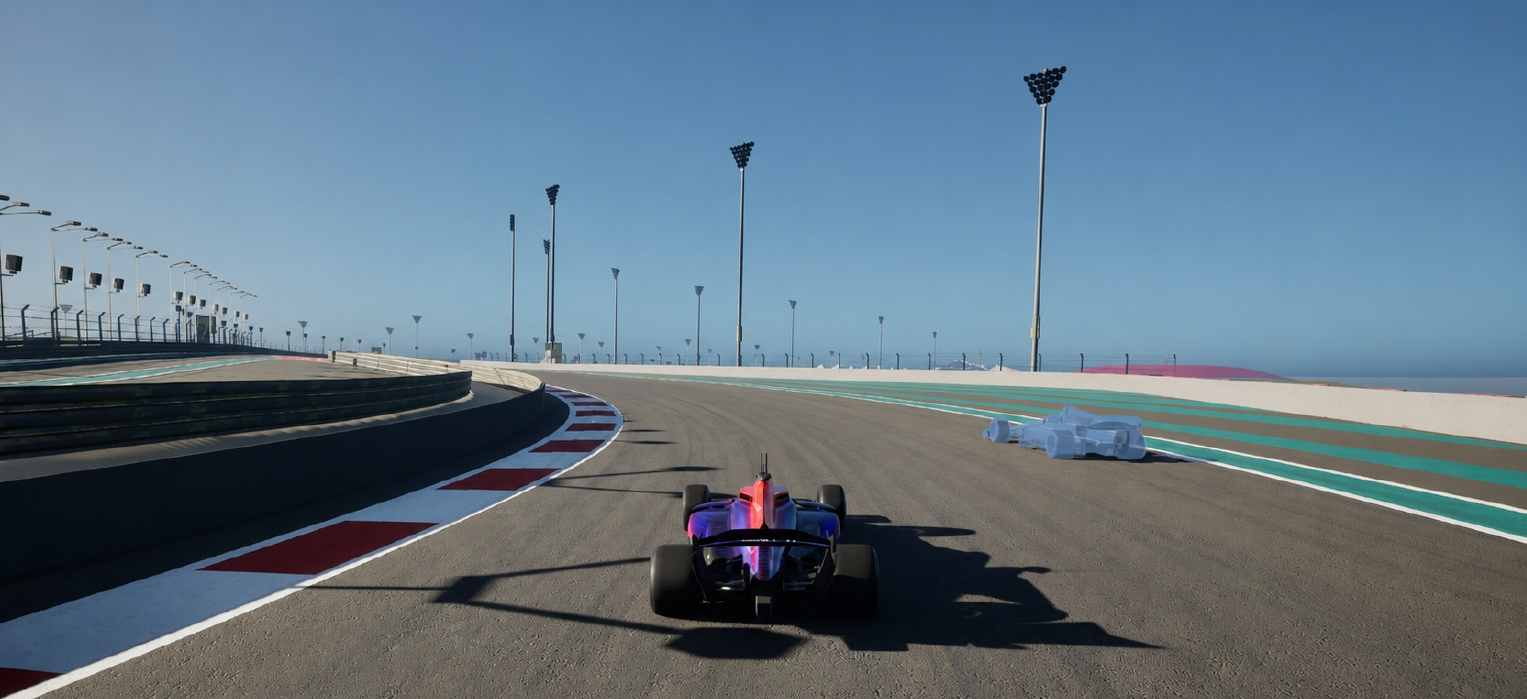}
    \caption{
    Illustration of the environment of Autoverse simulator with virtual opponents.
    }
    \label{SF16_simulator_env}
\end{figure}

The ego vehicle is running in the Autoverse simulator.
The baseline simulation experiments are conducted on the Yas Marina North Circuit, consistent with the real-vehicle deployment setting. 
Additional cross-circuit generalization experiments are performed on different track layouts, with detailed racetrack information described in Section~S10.4.
Virtual opponents are implemented as vectorized interactive agents responding to the ego vehicle in real time. 
Fig.~\ref{SF16_simulator_env} provides a conceptual visualization of the environment provided by the Autoverse simulator with virtual opponents.

Adversarial interaction scenarios are constructed to evaluate interactive performance under physical constraints. Each interaction scenario is initialized with the opponent vehicles placed around the ego vehicle. The default interaction duration is $20$ s.

Opponent vehicles are initialized by sampling the relative longitudinal distance, lateral position, and initial speed with respect to the ego vehicle. 
After scenario initialization, the opponent behaviors are generated by policies trained from interaction demonstrations using imitation learning rather than fixed scripted rules. 
Each opponent policy takes the local track context, ego--opponent relative states, and recent motion history as inputs, and produces interactive driving actions in real time.
The resulting virtual opponents reproduce representative racing behaviors such as holding line, defending, yielding, and competing during overtaking. 
Scenario variations are introduced by sampling the local track segment, ego--opponent initial relative position, speed difference, and opponent behavior condition.

For each method, scenarios are generated using the same scenario generation setting to ensure the repeatable and fair comparison. 
A trial is terminated when any of the following conditions occurs:
\begin{itemize}
    \item the evaluation horizon is reached;
    \item collision occurs;
    \item the ego vehicle violates track boundaries;
    \item the ego vehicle enters an unrecoverable unstable state.
\end{itemize}

In the simulation environment, the ego agent uses the same decision, planning, world-model rollout, and physical-limit-constrained control interfaces as in the real-vehicle stack. 
The simulation environment provides ego and opponent states, track information, collision status, boundary violation status, and dynamic-response variables. 
This makes it possible to evaluate both cognitive-side performance, such as interaction prediction and decision success, and physical-side performance, such as acceleration-envelope utilization, tire slip, tracking performance, and maneuvering stability.

Simulation testing enables controlled comparison between the proposed method and baselines, ablation of world-model-based future-aware reasoning, and reproduction of representative real-racing cases for failure analysis and recovery validation.

\subsubsection*{S8.4 Baseline Methods}

The proposed world-model-driven method is compared with three baselines: behavioral cloning, soft actor--critic, and a rule-based planner.

\paragraph{Behavioral cloning.}
The behavioral cloning baseline learns a direct mapping from world-state representation to racing decisions using supervised learning from demonstration data. Given a policy state $s_t$ corresponding to the racing world state $w_t$, the BC policy is trained by minimizing
\begin{equation}
L_{\mathrm{BC}}
=
\mathbb{E}_{(s_t,a_t^{\mathrm{e}})\sim\mathcal{D}}
\left[
\left\|
\pi_{\theta}(s_t)
-
a_t^{\mathrm{e}}
\right\|^2
\right],
\end{equation}
where $a_t^{\mathrm{e}}$ denotes the demonstration action. This baseline evaluates whether supervised imitation alone can provide sufficient interaction reasoning and generalization under adversarial racing conditions.

\paragraph{Soft actor--critic.}
The SAC baseline trains a model-free reinforcement-learning policy using the same state and action interface as the proposed decision policy. It optimizes racing reward through interaction with the simulator but does not use world-model-based parallel future evaluation for online candidate scoring. This baseline evaluates the performance of a policy-centric reinforcement-learning approach without explicit predictive world-model reasoning during deployment.

\paragraph{Rule-based planner.}
The rule-based planner corresponds to the state-machine-based competition stack. It selects racing behaviors according to manually designed logic, such as following, overtaking, defending, and speed adjustment based on relative opponent position and track context. This baseline provides a real-deployment reference for comparing the refined world-model-driven policy against the original competition stack.

\subsubsection*{S8.5 Evaluation Metrics}

For the benchmark comparison, all baselines are assessed using the same interaction and safety metrics. 
The downstream controller and actuation interfaces are kept consistent when applicable. 
To evaluate whether each method can complete competitive interaction scenarios safely and efficiently, the success rate, safe mileage, average speed, and overtaking frequency are adopted as evaluation metrics.

Beyond the benchmark comparison, additional metrics are used to evaluate agent performance and evolution in physical terms. 
Tire slips, tracking errors, and maneuvering stability are adopted to assess limit-aware control near the dynamics boundary. The  average speed, lap time, and reachable acceleration states are used to characterize the evolution of the agent across refinement stages.

\subsubsection*{S8.6 Real-World Case Reproduction}

Two representative real-racing cases are reproduced in simulation for controlled validation: a wheel-to-wheel overtaking case and a spin case. The details are introduced in Supplementary Section~S10.2 and S10.3.

These case reproductions connect simulation evaluation with real-racing events. They allow the validation pipeline to test not only average performance, but also whether the world-model-centric agent can predict, diagnose, and correct the critical interaction and physical-limit failure modes.

\subsection*{S9. Safety Monitoring and Intervention Logic}

The safety monitoring layer supervises the localization and ego-state estimation subsystem during real-vehicle deployment. Its purpose is to detect sensor dropout, inconsistent velocity measurements, and abnormal estimation residuals before they propagate to world-state construction, future-aware reasoning, and downstream control. The safety layer does not replace the world model or controller; instead, it provides an independent intervention mechanism for real-time deployment under high-speed racing conditions.

\begin{figure}[!htb]
    \centering
    \includegraphics[width=12cm]{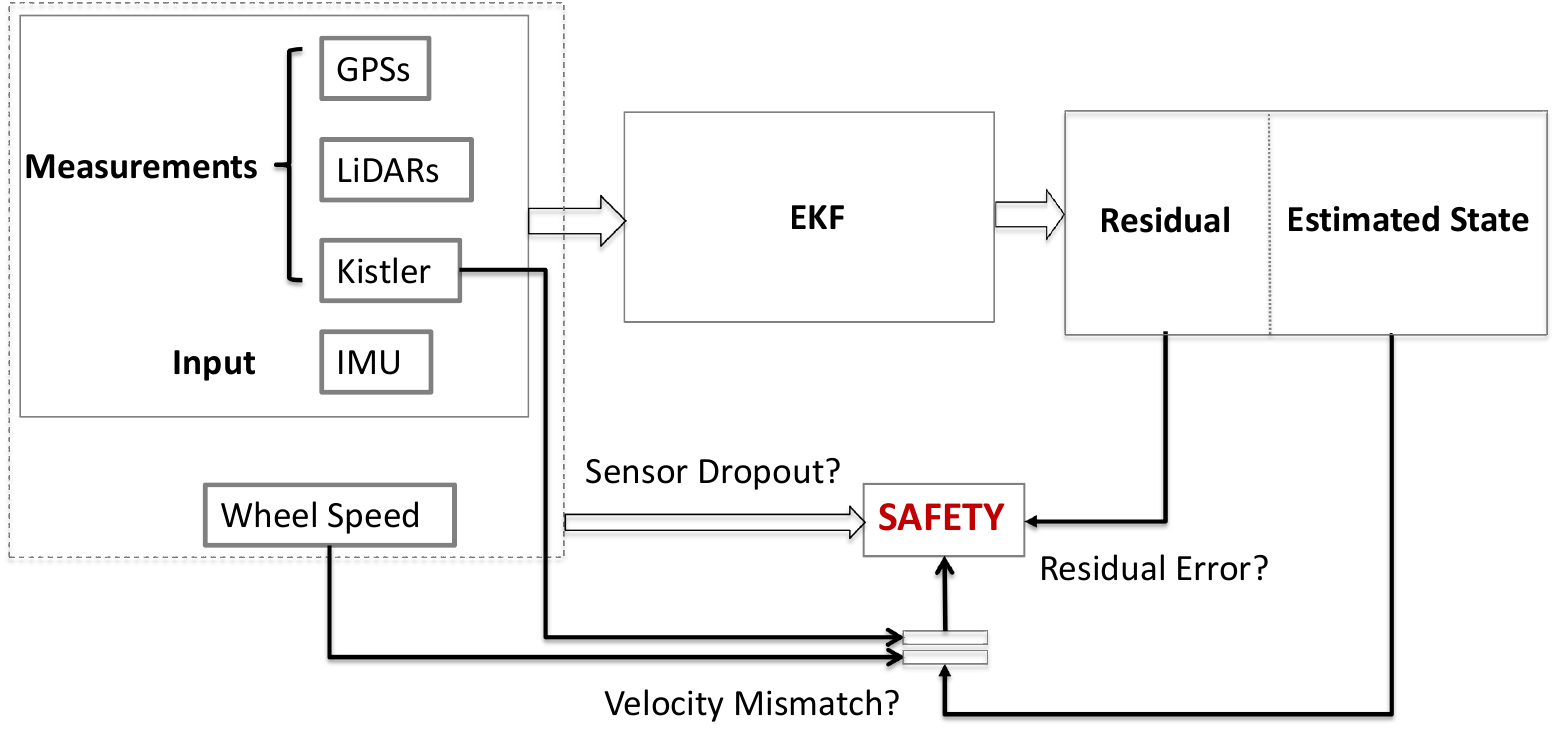}
    \caption{Safety monitoring framework for localization and ego-state estimation.}
    \label{SF17_safety_fw_est}
\end{figure}

The output safety mode is defined as
\begin{equation}
\mathrm{Mode}
\in
\{
\text{\textbf{Normal}},
\text{\textbf{Degradation}},
\text{\textbf{Emergency stop}}
\}.
\end{equation}
In \textbf{Normal} mode, the autonomy stack operates without additional restrictions. In \textbf{Degradation} mode, the ego vehicle switches to a reduced-speed operating mode with conservative planning and control limits. In \textbf{Emergency stop} mode, the ego vehicle applies immediate braking while maintaining the current heading as far as possible. The thresholds and timing parameters are platform-dependent and are therefore configured according to sensor characteristics and deployment conditions. Fig.~\ref{SF17_safety_fw_est} illustrates the overall safety logic.

\subsubsection*{S9.1 Sensor Dropout Logic}

Each sensor stream is associated with a binary availability flag
\begin{equation}
s \in \{0,1\},
\end{equation}
where $s=1$ indicates a valid and up-to-date signal and $s=0$ indicates dropout or timeout. The position-source flag is defined as
\begin{equation}
s_{\mathrm{pos}}
=
s_{\mathrm{GNSS}}
\lor
s_{\mathrm{LiDAR}},
\end{equation}
which indicates whether at least one absolute-position source is available.

The dropout-based safety mode is
\begin{equation}
\mathrm{Mode}
=
\begin{cases}
\text{\textbf{Emergency stop}},
&
\neg s_{\mathrm{IMU}}
\lor
\neg s_{\mathrm{pos}}
\lor
\neg s_{\mathrm{whl}},
\\[3pt]
\text{\textbf{Degradation}},
&
s_{\mathrm{IMU}}
\land
s_{\mathrm{pos}}
\land
\neg s_{\mathrm{Kis}},
\\[3pt]
\text{\textbf{Normal}},
&
s_{\mathrm{IMU}}
\land
s_{\mathrm{pos}}
\land
s_{\mathrm{Kis}}.
\end{cases}
\label{eq:supp_safety_dropout_logic}
\end{equation}
Here $s_{\mathrm{IMU}}$, $s_{\mathrm{GNSS}}$, $s_{\mathrm{LiDAR}}$, $s_{\mathrm{whl}}$, and $s_{\mathrm{Kis}}$ denote the availability of IMU, GNSS, LiDAR localization, wheel-speed, and Kistler velocity signals, respectively. The logic enforces that the ego vehicle cannot continue normal operation without inertial sensing, at least one absolute-position source, and wheel-speed information.

\subsubsection*{S9.2 Velocity Consistency Logic}

Velocity consistency is evaluated using redundant velocity sources. The mismatch terms are defined as
\begin{equation}
e_{kw}
\triangleq
\left|
v_x^{\mathrm{Kis}}
-
v^{\mathrm{whl}}
\right|,
\qquad
e_{ke}
\triangleq
\left|
v_x^{\mathrm{Kis}}
-
v^{\mathrm{est}}
\right|,
\end{equation}
where $v_x^{\mathrm{Kis}}$ is the Kistler longitudinal speed, $v^{\mathrm{whl}}$ is the average front-wheel longitudinal speed, and
\begin{equation}
v^{\mathrm{est}}
=
\sqrt{
(v_x^{\mathrm{est}})^2
+
(v_y^{\mathrm{est}})^2
}
\end{equation}
is the estimator-derived speed magnitude.

A window-based Boolean comparison operator is used to avoid reacting to isolated spikes. For a signal $e$ and threshold $\delta$, the statement
\begin{equation}
e
\overset{\mathcal{W},\rho}{>}
\delta
\end{equation}
is true when the proportion of samples in sliding window $\mathcal{W}$ satisfying $e>\delta$ exceeds the ratio $\rho$.

The velocity-consistency safety mode is
\begin{equation}
\mathrm{Mode}
=
\begin{cases}
\text{\textbf{Emergency stop}},
&
e_{kw}
\overset{\mathcal{W},\rho}{>}
\delta_{kw}
\lor
e_{ke}
\overset{\mathcal{W},\rho}{>}
\delta_{ke},
\\[4pt]
\text{\textbf{Normal}},
&
\text{otherwise}.
\end{cases}
\label{eq:supp_speed_consistency_emergency}
\end{equation}
This check detects large disagreements among independent velocity estimates, which may indicate wheel-speed slip, external velocity sensor failure, or state-estimation inconsistency.

\subsubsection*{S9.3 Residual-Based Consistency Logic}

Residual-based consistency checks compare incoming sensor measurements with predicted observations from the estimator. Let $r_i$ denote the $i$-th measurement residual. Residual indicators are defined as
\begin{equation}
\begin{alignedat}{2}
\mathcal{D}_{\mathrm{res}}
&=
\exists\, i\in\mathcal{I}:
|r_i|
\overset{\mathcal{W},\rho}{>}
\delta_{d,i},
&\qquad
\mathcal{C}_{\mathrm{Kis}}
&=
\exists\, i\in\mathcal{I}_{\mathrm{Kis}}:
|r_i|
\overset{\mathcal{W},\rho}{>}
\delta_{c,i},
\\[2pt]
\mathcal{C}_{\mathrm{GNSS}}
&=
\exists\, i\in\mathcal{I}_{\mathrm{GNSS}}:
|r_i|
\overset{\mathcal{W},\rho}{>}
\delta_{c,i},
&\qquad
\mathcal{C}_{\mathrm{LiDAR}}
&=
\exists\, i\in\mathcal{I}_{\mathrm{LiDAR}}:
|r_i|
\overset{\mathcal{W},\rho}{>}
\delta_{c,i}.
\end{alignedat}
\label{eq:supp_residual_indicators}
\end{equation}
Here $\mathcal{I}$ denotes all residual channels, while $\mathcal{I}_{\mathrm{Kis}}$, $\mathcal{I}_{\mathrm{GNSS}}$, and $\mathcal{I}_{\mathrm{LiDAR}}$ denote modality-specific residual subsets. The thresholds $\delta_{d,i}$ and $\delta_{c,i}$ represent degradation and critical residual thresholds, respectively.

The residual-based safety mode is
\begin{equation}
\mathrm{Mode}
=
\begin{cases}
\text{\textbf{Emergency stop}},
&
\mathcal{C}_{\mathrm{Kis}}
\lor
(
\mathcal{C}_{\mathrm{GNSS}}
\land
\mathcal{C}_{\mathrm{LiDAR}}
),
\\[4pt]
\text{\textbf{Degradation}},
&
\mathcal{D}_{\mathrm{res}},
\\[4pt]
\text{\textbf{Normal}},
&
\text{otherwise}.
\end{cases}
\label{eq:supp_residual_based_mode}
\end{equation}
An emergency stop is triggered when critical residuals indicate Kistler failure, or when both GNSS and LiDAR position residuals violate their critical thresholds simultaneously. The latter condition implements mutual backup between the two absolute-position sources and avoids emergency intervention from a single-source anomaly alone.

\subsubsection*{S9.4 Intervention Priority and Deployment Use}

When multiple safety checks are active, the final safety mode is selected according to the most conservative triggered mode:
\begin{equation}
\mathrm{Mode}_{\mathrm{final}}
=
\max_{\mathrm{priority}}
\left(
\mathrm{Mode}_{\mathrm{dropout}},
\mathrm{Mode}_{\mathrm{velocity}},
\mathrm{Mode}_{\mathrm{residual}}
\right),
\end{equation}
where the priority order is
\begin{equation}
\text{\textbf{Emergency stop}}
>
\text{\textbf{Degradation}}
>
\text{\textbf{Normal}}.
\end{equation}

The selected safety mode is then passed to the planning and control stack. In Normal mode, the system follows the standard world-model-driven perception--prediction--decision--control loop. In Degradation mode, the planner and controller use reduced speed limits and conservative physical-limit envelopes. In Emergency stop mode, the control module overrides the selected racing trajectory and applies a braking maneuver.

This safety monitoring layer provides an independent deployment safeguard for the high-frequency localization and state-estimation subsystem. It helps prevent sensor or estimator failures from propagating into incorrect world-state construction and unsafe high-speed execution.

\subsection*{S10. Additional Validation and Ablation Details}

This section summarizes the validation and ablation settings used to support the main results. The details focus on four parts that are directly evaluated in the main manuscript: world-model-based future-aware reasoning, real-world overtaking case reproduction, physical-limit-aware adaptation in the spin case, and cross-circuit generalization.

\subsubsection*{S10.1 Ablation of World-Model-Based Future-Aware Reasoning}

The proposed method is compared with an ablated variant that removes recursive world-model rollouts during online trajectory evaluation. In the ablated variant, candidate trajectories are generated using the same decision policy and multimodal planner, but the final trajectory is selected using only the current observed world state.
\begin{equation} 
\tau_{t,\mathrm{abl}}^\ast 
= 
\arg\max_{\tau_t^k\in\mathcal{T}_t} 
S_{\mathrm{current}}(\tau_t^k,w_t). 
\end{equation}
where $S_{\mathrm{current}}$ scores candidate trajectories through speed score and the driving safety field.
In contrast, the full method generates multimodal candidate trajectories and selects the final trajectory using predicted future world states, as described in Eq.~(\ref{eq:supp_selected_candidate}). 

This ablation isolates the effect of predictive reasoning. Comparing the full method with this variant shows whether world-model rollouts improve the agent's ability to anticipate opponent motion, avoid unsafe overtaking decisions, and select trajectories that remain executable under downstream physical-limit-constrained control.

\subsubsection*{S10.2 Real-World Overtaking Case Reproduction}

The real-world overtaking case is reproduced to evaluate whether the world model captures short-horizon ego--opponent interaction evolution during wheel-to-wheel racing. The reproduced scenario uses the observed interaction geometry, ego trajectory, opponent trajectory, and track context as reference conditions. The analysis compares realized ego--opponent trajectories with world-model-predicted trajectories, and evaluates whether future-aware reasoning selects a maneuver that remains collision-free, physically feasible, and competitive.

This case supports the qualitative validation of the predictive reasoning process. The key comparison is whether the world-model-predicted interaction remains consistent with the realized wheel-to-wheel racing interaction, and whether the selected trajectory avoids the unsafe decision observed in the no-world-model ablation.

\subsubsection*{S10.3 Physical-limit-aware adaptation and Spin-Case Reproduction}

The spin case observed during real-world racing is reproduced to evaluate physical-limit-aware adaptation and failure recovery. The reproduced scenario includes the high-curvature corner, aggressive acceleration command, high lateral load, and tire-state condition associated with the observed instability.

The comparison includes four settings:
\begin{enumerate}
        \item baseline control without additional limit degradation;
        \item thermal-adaptive limit degradation;
        \item world-model-prediction-deviation-based limit degradation;
    \item synergistic degradation combining thermal and world-model-based signals.
\end{enumerate}

The evaluated quantities include slip angle, yaw response, trajectory tracking behavior, and corner-exit performance. This validation assesses whether the adaptive physical-limit mechanism reduces the effective feasible-motion boundary before instability develops, thereby preventing or mitigating the spin failure mode observed during the race.

\subsubsection*{S10.4 Cross-Circuit Generalization Protocol}

Cross-circuit generalization is conducted by deploying the agent on target circuits without additional training. 
The source circuit is the Yas Marina North Circuit, which has a total length of $3002~\mathrm{m}$ and contains $8$ corner segments. 
The target circuits include the Yas Marina Circuit, Autonodrome, and Suzuka Circuit, with total lengths of $5284~\mathrm{m}$, $6117~\mathrm{m}$, and $5809~\mathrm{m}$, respectively. 
They contain $16$, $16$, and $18$ corner segments, respectively. 
These circuits differ in total length, curvature distribution, corner density, and the arrangement of high-speed straights and consecutive cornering segments, thereby providing diverse track geometries for evaluating cross-circuit transfer.

The same world-model interface, decision policy, multimodal planner, and physical-limit-constrained controller were used across all target circuits.
For each circuit, random interaction scenarios were generated by sampling ego--opponent initial relative positions, speed differences, and opponent behavior conditions from the same predefined ranges.
The evaluation reports the average speed, maximum speed, overtaking frequency, and success rate in random interaction scenarios and collision-related results. This protocol assesses whether the world model and policy preserve interaction capability and limit-aware execution under changes in track geometry and interaction contexts.

\end{appendices}

\end{document}